\documentclass[10pt,journal,compsoc]{IEEEtran}
\usepackage{textcomp}

\ifCLASSOPTIONcompsoc
  \usepackage[nocompress]{cite}
\else
  \usepackage{cite}
\fi

\usepackage{graphicx}
\usepackage{trimclip}
\usepackage{adjustbox} 
\usepackage{subfigure}
\usepackage{float}
\usepackage{epstopdf}

\usepackage{longtable}
\usepackage{lscape}
\usepackage{setspace}
\usepackage{multirow,multicol}
\usepackage{xcolor}
\usepackage{colortbl} 
\usepackage{url}      

\definecolor{bsPink}{HTML}{F9DBDF}
\definecolor{bsGreen}{HTML}{E3F2D9}
\definecolor{bsBlue}{HTML}{DBE3F4}
\definecolor{bsYellow}{HTML}{FEF2CB}

\newcommand{\bsfigurefile}[3]{%
  \IfFileExists{#1.pdf}{\includegraphics[width=#2]{#1.pdf}}{%
  \IfFileExists{#1.png}{\includegraphics[width=#2]{#1.png}}{%
  \IfFileExists{#1.jpg}{\includegraphics[width=#2]{#1.jpg}}{%
  \IfFileExists{#1.jpeg}{\includegraphics[width=#2]{#1.jpeg}}{%
  \IfFileExists{#1.eps}{\includegraphics[width=#2]{#1.eps}}{%
    \fbox{\parbox[c][#3][c]{#2}{\centering\footnotesize
      \texttt{#1}\\[0.45em]\textit{Figure placeholder}}}%
  }}}}}%
}

\usepackage{stfloats}
\usepackage{capt-of} 
\usepackage{placeins} 

\newcommand{\bsblue}[1]{#1}
\usepackage[switch]{lineno}
\usepackage{array}

\newcommand{\bsSub}[1]{#1}
\newcommand{\bsSup}[1]{#1}

\usepackage{indentfirst}
\usepackage{amsmath}
\usepackage{mathtools} 
\usepackage{amsfonts}

\newcounter{bsStressFigureBase}

\makeatletter
\newcommand{\bssection}[1]{%
  \FloatBarrier
  \section{#1}%
  \suppressfloats[t]%
  \global\@dbltopnum\z@
  \global\@dbltoproom\z@
}
\makeatother

\newcommand{\bsleftformulasetup}{%
  \small
}

\newcommand{\bswideformulasetup}{%
  \small
  \setlength{\abovedisplayskip}{0pt}%
  \setlength{\belowdisplayskip}{0pt}%
  \setlength{\abovedisplayshortskip}{0pt}%
  \setlength{\belowdisplayshortskip}{0pt}%
}

\usepackage[linesnumbered,ruled,vlined]{algorithm2e}

\SetKwInOut{Input}{Input}
\SetKwInOut{Output}{Output}

\begin{document}

\raggedbottom

%

\title{
\fontsize{17pt}{24pt}\selectfont
Brain-SAD: A Brain-Inspired Safe Autonomous Driving Control Framework with Dynamic Fear-Oriented Constraint on Dual-Policy
}

\author{Huan Rong, Chao Yin, ANOUAR IMEL, Yijie Xia,Tinghuai Ma* %
 \IEEEcompsocitemizethanks{ 
  
\IEEEcompsocthanksitem * corresponding author:Tinghuai Ma
\IEEEcompsocthanksitem Huan Rong is with the School of Artificial Intelligence, Nanjing University of Information Science \& Technology, Jiangsu, Nanjing, 210-044, China, E-mail: 1227558210@qq.com
\IEEEcompsocthanksitem Chao Yin is with the School of Artificial Intelligence, Nanjing 
University of Information Science \& Technology, Jiangsu, Nanjing, 210-044, China, E-mail: yinchao0127@163.com

\IEEEcompsocthanksitem ANOUAR IMEL is with the School of Information and Communication Engineering, University of Electronic Science and Technology of China, Sichuan, Chengdu, 611731, China, E-mail: 202524010105@std.uestc.edu.cn

\IEEEcompsocthanksitem Yijie Xia is with the School of Artificial Intelligence, Nanjing University of Information Science \& Technology, Jiangsu, Nanjing, 210-044, China, E-mail: xyj03930437@189.cn  

\IEEEcompsocthanksitem Tinghuai MA is with the School of Software, Nanjing University of Information Science \& Technology, Nanjing, 210044, Jiangsu, China.School of Computer Engineering, Jiangsu Ocean University, Lianyungang, 222005, Jiangsu,China. E-mail: thma@nuist.edu.cn

\IEEEcompsocthanksitem This work is supported in part by the National Natural Science Foundation of China (No. 62372243).
}
\\
\thanks{}
}

\markboth{}%
{Chao~Yin \MakeLowercase{\textit{et al.}}: Brain-SAD}

%



\IEEEtitleabstractindextext{%
\begin{abstract}
\begingroup
\rightskip=0pt\relax
\leftskip=0pt\relax
\parfillskip=0pt plus 1fil\relax
\emergencystretch=0.8em
\label{sec:abstract}Constrained Reinforcement Learning (RL) has recently gained increasing attention in the field of Safe Autonomous Driving (AD), where the general mechanism is to maximize the expected reward while keeping the overall action risk bounded. In this way, the safety issues arising in AD can be mitigated through constrained actions. However, existing Constrained RL methods still lack dynamics on the imposed constraints. For instance, the action cost adopted by the existing Primal-Dual/soft-constrained methods is often defined as static state-to-cost mapping, and the safe-action projection in hard-constrained methods relies on the static projection with the fixed feasible region boundary estimated from offline demonstrations. The above drawback tightly couples the imposed constraints to the training scenarios, leaving the AD policy hard to handle different interaction scenarios, due to the improper state-level action-cost and the static projection boundary. Consequently, in this paper, we propose Brain-SAD, a brain-inspired safe autonomous driving control framework with dynamic fear-oriented constraints. By perceiving the current vehicle-interaction scene, Brain-SAD generates dynamic fear signal as fear reaction to online decide long-term policy for regular interaction or short-term policy for urgent-collision defense. In such two policy, the above fear-reaction will be constructed as the dynamic fear constraints, respectively reflecting the overall fear cost directly coupled with action-impact, and the dynamic fear boundary of the feasible region derived from different risky neighbors, both of which will in turn serve for the online policy optimization. Experimental results show that Brain-SAD outperforms existing methods, achieving higher success rate in shorter task-completion and collision-recovery time, and exhibits stronger reliability across continuous intersections of fluctuating complexity.
\par
\endgroup
\end{abstract}

\begin{IEEEkeywords}
Dynamic Constrained RL; Safe Autonomous Driving; Brain-Inspired Computation;
\end{IEEEkeywords}}

\maketitle

\IEEEdisplaynontitleabstractindextext

\IEEEpeerreviewmaketitle
\IEEEraisesectionheading{
\section{Introduction}
\label{sec:introduction}}
\IEEEPARstart{W}{ith} the success of Reinforcement Learning (RL) applied to Autonomous Driving (AD), the AD systems are now expected to operate in increasingly complex scenarios, such as heavy traffic with cross-lane vehicle interaction \cite{01} or continuous intersections with traffic lights and roadblocks \cite{02}. Consequently, the requirement of \textbf{\textit{Safe RL}} has been emphasized, aimed at enhancing the \textbf{\textit{robustness}} (or \textbf{\textit{reliability}}) of RL policy on AD controlling \cite{03}. As the major realization of \textit{Safe RL}, \textbf{\textit{Constrained RL}} incorporates safety constraints into traditional RL policy, where the policy-controlled ego-vehicle is enforced to complete entire driving task while adhering to safety constraints and avoiding collision risk \cite{03}.

\nocite{04,05,06,07,08,09,10,11,12,13}

\begin{figure}[tbp]
\setlength{\abovecaptionskip}{-0.0cm}
\setlength{\belowcaptionskip}{0pt}
\centering
\includegraphics[
    width=\columnwidth,
    keepaspectratio
]{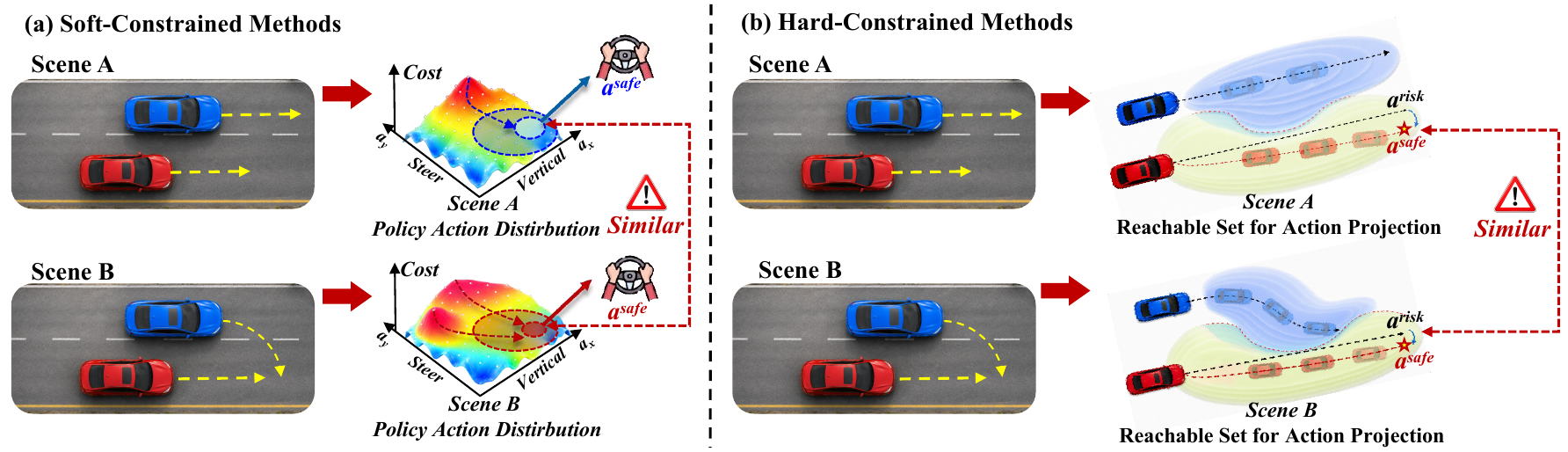}
\caption{Two \textbf{\textit{Constrained RL}} paradigms for Safe AD: Soft-constraint (Primal-dual) \& Hard-constraint.}
\label{Fig:1}
\vspace{-0.15cm}   
\end{figure}

\begin{figure*}[tbp]
\setlength{\abovecaptionskip}{0.05cm}
\setlength{\belowcaptionskip}{0pt}
\centering
\includegraphics[width=0.9\textwidth]{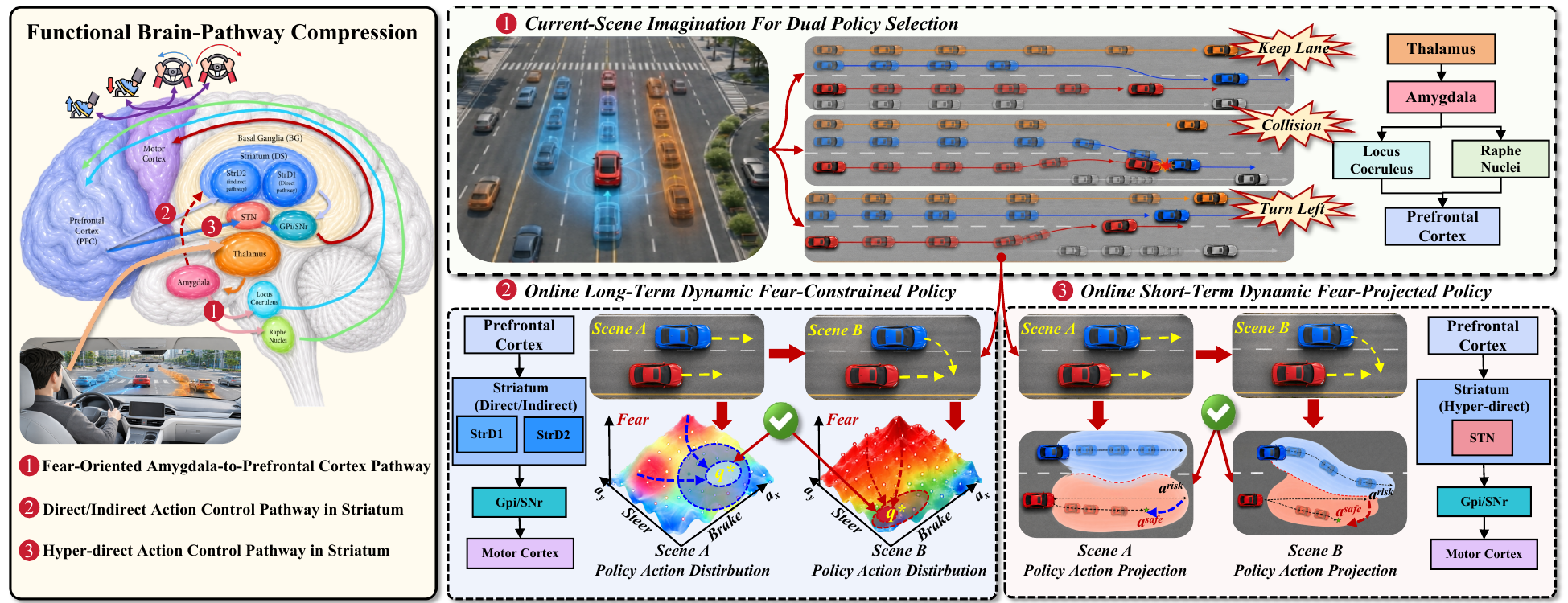}
\caption{Our proposed Brain-SAD safe AD control framework with dynamic fear-oriented constraint (\textbf{NOTE}: Striatal D1/D2 (StrD1/StrD2) neurons belong to Direct/Indirect pathway in Striatum \cite{14}, Subthalamic Nucleus (STN) in Hyper-direct pathway \cite{15}).}
\label{Fig:2}
\vspace{-0.05cm}  
\end{figure*}

Generally, as illustrated in Fig. \ref{Fig:1}, the formulation of Constrained RL can be categorized into two paradigms including Primal-dual/soft-constrained (\textbf{Cumulative}) \cite{04,05,06} \& hard-constrained (\textbf{Instantaneous}) methods \cite{07,08}. Supposing that, at every \textbf{\textit{$t$-th}} time step, the \textbf{\textit{ego-vehicle}} currently in state \textbf{\textit{$s_t$}} selects an action \textbf{\textit{$a_t$}} by policy \textbf{\textit{$\pi$}}, receiving reward \textbf{\textit{$r_t$}} at cost \textbf{\textit{$c_t$}} (e.g., the \textit{risk} incurred by executing \textbf{\textit{$a_t$}}), thereby generating a trajectory \textbf{\textit{$\tau$}}=$\{s_1,a_1,r_1,c_1,s_2,a_2,r_2,c_2,\ldots,s_t,a_t,r_t,c_t,\ldots\}$. In Fig. \ref{Fig:1} (\textit{a}), given the trajectory starting from $(s_1,a_1)$, the soft-constrained (\textbf{Cumulative}) method (see Fig. \ref{Fig:1} (\textit{a})) requires the policy \textbf{\textit{$\pi$}} to maximize the cumulative \textbf{\textit{future}} rewards as $\mathbb{E}_{\tau\sim\pi}[\sum_{t=1}^{T}\gamma^t\cdot r(s_t,a_t)]$ while satisfying \textbf{\textit{cost}} constraints as $[\sum_{t=1}^{T}\gamma^t\cdot cost(s_t,a_t)\leq\delta]$ (where $\delta$ is the cost threshold). In contrast, in Fig. \ref{Fig:1} (\textit{b}), the hard-constrained (\textbf{Instantaneous}) method (see Fig. \ref{Fig:1} (\textit{b})) deals with the original risk action $a_t^{risk}$ output by policy \textbf{\textit{$\pi$}} directly via \textbf{\textit{action projection}} for efficiency \cite{07}. That is, given the current $(s_t,a_t^{risk})$ of the ego-vehicle, its reachable set will be predicted as all the reachable states/actions over the \textbf{\textit{future}} horizon \textbf{\textit{$t$}} steps, and so will its surrounding vehicle's \cite{09}. Then, the intersection of the reachable sets between the ego and surrounding vehicles will be considered as the \textbf{\textit{unsafe region}}, or the \textbf{\textit{safety constraint}} used to project the original risk action $a_t^{risk}$ of the ego-vehicle onto the closest \textbf{\textit{safe action}} $a_t^{safe}$ in the feasible region. Particularly, the above \textbf{\textit{cumulative}} action-cost constraint for action distribution and \textbf{\textit{instantaneous}} feasible-region constraint for action projection can be mixed to satisfy more stringent \textit{Safe RL} requirements \cite{10}.

Unfortunately, it has been found that the constraints imposed by the current soft-(\textbf{Cumulative}) \& hard-(\textbf{Instantaneous}) methods lacks \textbf{\textit{dynamics}} across different interaction scenarios \cite{11}. Specifically in Fig. \ref{Fig:1}, the \textbf{\textit{red}} ego-car always keeps straight in Scene A, while its \textbf{\textit{blue}} neighbor has direction turning intent in Scene B. Obviously, in \textbf{Scene B}, the \textbf{\textit{ego-vehicle}} is expected to avoid collision, however, similar actions have been selected in both the Scene A \& B. The above drawback can be ascribed to the reason that, the action-cost constraint in soft-(\textbf{Cumulative}) methods often relies on static-rules \cite{04,05,06}, while the feasible-region constraint in hard-(\textbf{Instantaneous}) methods are often realized by static projection operators \cite{07,08}. Consequently, the \textbf{\textit{cumulative}} action-cost (see Fig. \ref{Fig:1}(\textit{a})) and the \textbf{\textit{instantaneous}} action-projection (see Fig. \ref{Fig:1}(\textit{b})) tend to induce similar action distributions and feasible regions across scenes with different safety requirements, preventing the ego-vehicle from reacting appropriately\cite{12}.

In our view, the key challenge in overcoming this drawback is how to construct a \textbf{\textit{Dynamic Constraint}} adapted to different interaction scenarios \cite{13}. For one thing, the \textbf{\textit{action-cost}} constraint (i.e., $[\sum_{t=1}^{T}\gamma^t\cdot cost(s_t,a_t)\leq\delta]$) in soft-(\textbf{Cumulative}) methods should be dynamic action-cost estimation, instead of static-rules for $cost(s_t,a_t)$. For another, the \textbf{\textit{feasible-region}} constraint in hard-(\textbf{Instantaneous}) methods should consider dynamic boundary, instead of static projection operators. If possible, we could further integrate these two constraints into a unified controlling framework via such a \textbf{\textit{dynamic constraint}}, so as to deal with the regular scene-evolving and urgent-collision more flexibly and effectively.

\begingroup
\setlength{\emergencystretch}{1em}
Fortunately, the bottom-up-to-top-down action-control pathway of the human brain, which follows a \mbox{\textbf{\textit{Perception}}$\rightarrow$}\allowbreak\mbox{\textbf{\textit{Determination}}$\rightarrow$}\allowbreak\mbox{\textbf{\textit{Action}}} routine, offers a better inspiration. As the functional compression shown in Fig. \ref{Fig:2} (\textit{left}), when our brain observes the current driving scene (entailing \textbf{\textit{implicit}} neighboring interaction intent), Amygdala (\textbf{AMY}) will release the fear-oriented neurotransmitter (i.e., Norepinephrine (NE) for nervousness, Serotonin (5-HT) for calmness) as the production of perception. Such the fear-oriented neurotransmitter will be integrated into a unified signal conveyed to Prefrontal Cortex (\textbf{PFC}), determining the action (\textit{policy}) more appropriate with the current interaction scene, or correspondingly invoking two Striatum action pathways in Basal Ganglia (\textbf{BG}). Here, the Direct/Indirect pathway in Striatum generates action in regular interaction scene, while Hyper-direction pathway generates emergent defensive action\footnote{See more details on the brain-mechanism of action-control in Appendix \textbf{A}.}.
\par
\endgroup

Based on the above, in this paper, we propose Brain-SAD, a novel brain-inspired safe autonomous driving control framework with dynamic fear-oriented constraint. In Fig. \ref{Fig:2} (\textit{right}), Brain-SAD perceives and quantifies a \textbf{\textit{dynamic fear-reaction}} from the current interaction scenario and its predicted evolution for policy selection. This fear-reaction penetrates into the long/short-term policies, respectively constructed as \textbf{\textit{dynamic}} action fear-cost and feasible-region fear-boundary constraints, to handle different scenarios. To the best of our knowledge, Brain-SAD is the first to integrate two types of dynamic constraints via brain-inspiration as a \textbf{\textit{dynamic Constrained RL}} for higher reliability in Safe AD. The overall contributions are listed as follows:

1) Inspired by the fear-oriented Amygdala-to-Prefrontal Cortex Pathway (AMY$\rightarrow$PFC, see Fig. \ref{Fig:2} \textit{left}-\textcircled{1}), we perceive a \textit{dynamic fear-reaction} of \textbf{AMY} from the ego-neighbor-vehicle interaction by rolling out future interaction trajectories (see Fig. \ref{Fig:2} \textit{right}-\textcircled{1}). The dynamic fear-reaction of \textbf{AMY} is further quantified into the \textit{learnable} policy-selection signal of \textbf{PFC} to select long-/short-term policy, correspondingly generating ego-vehicle action under different safety requirements.

2) Inspired by the Direct/Indirect action-control pathway in Striatum (PFC$\rightarrow$StrD1/StrD2, see Fig. \ref{Fig:2} \textit{left}-\textcircled{2}), we adopt an \textit{online} long-term dynamic fear-constrained policy when the above fear-reaction is mild, particularly dealing with regular vehicle-interaction (see Fig. \ref{Fig:2} \textit{right}-\textcircled{2}). The long-term policy constructs a \textit{dynamic action reward-cost} (\textbf{StrD1-StrD2}) constraint under the above fear-reaction to \textit{post-correct} the \textit{prior} action distribution, \textit{online} adapted with different scenarios.

3) Inspired by the Hyper-direct action control Pathway in Striatum (PFC$\rightarrow$STN, see Fig. \ref{Fig:2} \textit{left}-\textcircled{3}), we adopt a \textit{online} short-term dynamic fear-projected policy when facing a fierce fear-reaction, to handle urgent braking in emergency (see Fig. \ref{Fig:2} \textit{right}-\textcircled{3}). The short-term policy constructs a \textit{dynamic fear boundary} from risky neighboring vehicles to \textit{online} tighten or loosen the projected feasible-region in different scenarios, selecting a safe action directly for efficiency(\textbf{STN}).

In this way, our proposed Brain-SAD can handle continuous vehicle-interaction scenes with flexible policy selection driven by the dynamic fear-reaction. The fear-reaction is constructed as a specific \textbf{\textit{dynamic}} policy constraint that is \textbf{\textit{online- adapted}} across different scenarios, with the strength to deal with both regular scene evolving and urgent braking in emergency.

According to the experimental results, first, by tracking the fear-reaction derived policy selection signal, Brain-SAD can adopt long-/short-term policy matched to the current vehicle-interaction scenario. Based on this dual-path policy collaboration, Brain-SAD outperforms existing works, achieving a higher success rate in shorter task-completion and collision-recovery time. More importantly, the dynamic fear-oriented constraint enables Brain-SAD to generate a more stable action distribution even across continuous intersections of fluctuating complexity, revealing stronger reliability in handling different scenarios.

\bssection{Related Works}

\subsection{Constrained Behaviour Learning in Safe AD}

To address the safety issues involved in AD, constrained RL has gained increasing attention to keep action risk below the \textit{predefined} thresholds while optimizing expected reward. From an optimization standpoint, existing constrained RL works can be categorized into Primal-Dual/soft-constrained (\textbf{Cumulative}) and hard-constrained (\textbf{Instantaneous}) methods \cite{03}. The soft-constrained ones enable an agent to select actions that balance reward maximization and constraint satisfaction, for instance, W. U. Mondal et al. \cite{04} formulate a soft-constrained policy gradient algorithm based on a Markov Decision Process that guarantees an $\boldsymbol{\varepsilon}$ global optimality gap and $\boldsymbol{\varepsilon}$ constraint violation on sample complexity for general parameterized policies. Similarly, J. Dai et al. \cite{05} and C. Zhang et al. \cite{06} use \textbf{\textit{static projection}} to estimate the cost of a \textit{state as} caused by an action (e.g., action$\rightarrow$speed/distance$\rightarrow$cost), keeping the overall cost below a predefined threshold (i.e., $\sum_t^T cost\leq b$), thereby deriving a cost-constrained gradient online to optimize the ego-vehicle policy. However, the above \textbf{\textit{state-mapped cost}} incurs tight coupling between cost estimation and policy optimization, further causing \textit{policy instability} during online exploration \cite{03}. Consequently, lots of trajectory samples are needed in soft-constrained methods.

To overcome this instability, the hard-constrained methods, especially the projection-based ones, guarantee a state-wise constraint during execution, where a sub-level of the constraint function is defined as the safe set/region, ensuring each action remaining within it \cite{03}. Specifically, S. Lin et al. \cite{07} first constrain the current policy to remain close to a reference base policy under a divergence bound, and then derive the feasible action region from this constraint.Y. Zheng et al. \cite{08} use Hamilton-Jacobi (HJ) to estimate the violation value of each state, where the optimal upper boundary of the safe state is estimated by \textbf{\textit{offline}} \textbf{(as }\textbf{\textit{static}}\textbf{) }regression, projecting an implicit feasible region that maximizes the action's long-term reward. Consequently, an improperly estimated feasible-region boundary (e.g., an overly restrictive one) will reduce achievable rewards.

Beyond the soft-constrained and hard-constrained methods above, we further group the constrained-behaviour-learning works by how the constraint is imposed, into \textbf{\textit{ego-vehicle Policy-Oriented Constrained}} and \textbf{\textit{Environment-Oriented Prior-Constrained}} methods.

For \textbf{\textit{ego-vehicle policy-oriented constrained}} ones, Zhao Y. et al \cite{16} propose \textbf{CADRE}, a cascade cumulative reward-constrained model-free RL policy formulated via proximal optimization form to penalize risky actions \textbf{\textit{online}} through a specifically designed reward for safe behavior learning. Similarly, Coelho D et al. \cite{17} construct \textbf{RLfOLD} to maximize the probability on the explored action by online-generated expert demonstration as supervisory constraint. Pan M. et al \cite{18} propose \textbf{Iso-Dream++}, which \textbf{\textit{online-}}disentangles controllable and non-controllable dynamics during vehicle interactions, where by multi-step imagination, the controllable state-prediction at each step flows into the non-controllable state-transition as the corrected \textbf{\textit{future}} trajectory, altogether fed into the ego-vehicle policy to decide the current safe action. Moreover, through \textit{offline} trajectory (behavior) imitation, \textbf{MARL-CCE}\cite{19} aggregates the $\boldsymbol{K}$ action distributions corresponding to the interactions between the ego-vehicle and its $\boldsymbol{K}$ neighbors, aligning them with the offline \textit{true} actions as the constrained action space.

For \textbf{\textit{environment-oriented prior-constrained}} ones, on the one hand, some works first pre-train a recurrent world-model to generate feasible state-transition trajectory in the simulated environment, as a prior-constraint on situational understanding. For instance, \textbf{LS-Imagination}, proposed by Li J et al. \cite{20}, decodes the quantified signal from the current interaction scene to adopt either short-term imagination over a single future step or long-term imagination directly across multiple future steps, jumping closer to the task-goal. \textbf{CarPlanner} proposed by Zhang D et al. \cite{21}, imagines multiple future state-trajectories for vehicle-interactions, and selects the one with the highest driving-mode consistency and safety-rule compliance.The above imagined state-trajectories are used to train a Critic network \textit{offline}. In turn, the Critic is treated as an indirect supervisory reward constraint updating the Actor network to select actions critical to task-completion. On the other hand, LM-based models such as \textbf{SimLingo} \cite{22}, \textbf{Hybrid-Driving} \cite{23}, and \textbf{SafeAuto} \cite{24}, enhance reliable understanding of the current interaction scenario respectively by offline hazard exploration to constrain action space, a scenario-evolving knowledge graph that constrains reasoning hallucinations, and first-order predicate-logic embedding of traffic rules that corrects the initial action.

\subsection{Brain-Inspired RL on Agent Action Control}

Within the NeuroAI direction, brain-inspired Reinforcement Learning (RL) agents mainly adopt brain mechanisms to improve the policy-learning process, so as to improve the plausibility and performance of the artificial agent.

Specifically, Yang Z et al. \cite{25} propose \textbf{SVPG}, a synaptic-plasticity-inspired spiking RL policy, which adopts Leaky Integrate-and-Fire (LIF) spiking neurons to build a Winner-Take-All (WTA) circuit as the policy network, trained by reward-modulated spike-timing-dependent plasticity (R-STDP). He X et al. \cite{26} construct \textbf{FNI-RL}, an Amygdala fear-reaction inspired Safe-AD RL framework based on the interaction between a protagonist and an adversary agent: the protagonist maximizes long-term reward subject to a bound on its cumulative \textbf{\textit{state-level}} fear reaction, while the adversary maximizes that same fear reaction to perturb the protagonist.The action distribution of the two agents are aggregated for action selection at each step. Notably, in \textbf{FNI-RL}, although the protagonist agent incorporates fear-reaction as a cost to train the policy (i.e., $Q(s,a)-fear(s,a)$), the fear-reaction $fear(s,a)$ is a static mapping from the current state, rather than an estimation of the future fear influence after taking an action. That is, the similar state in different interaction scenarios may incur similar fear-reaction, no direct causality with action itself. Therefore, \textbf{FNI-RL} should still be attributed as \textbf{\textit{static}} fear-constraint at \textbf{\textit{state-level}}.

Other representative brain-inspired RL works on action control include: \textbf{FoG-RL} \cite{27}, a hippocampus-inspired RL policy with memory forgetting and growth along temporal exploration; \textbf{EFT-RL} \cite{28}, a Theory-of-Mind like RL policy that makes its action by predicting other agents' behavior; \textbf{Goal-Reducer} \cite{29}, a goal-decomposition RL policy that completes a  task via sub-goal actions, similar to human problem-solving; BRYANT \cite{30}, a brain-inspired RL policy that transforms sequential latent representations, reward, and action signals into the frequency domain to capture temporal causality and establish the neural dynamics; and the spiking neural network (SNN)-based RL policy \cite{31}, which models the joint action-control function of multiple brain regions, including the PFC, motor cortex, hippocampus, and cerebellum.

\subsection{Discussion on Existing Works}

Based on the above analysis, it is evident that the current Primal-Dual/soft-constrained (\textbf{Cumulative}) and hard-constrained (\textbf{Instantaneous}) methods still lack dynamics on their constraints, where the action-cost (including the fear-cost imposed by the brain-inspired \textbf{FNI-RL} \cite{26}) is, in effect, a static-state-mapping rule such as $\boldsymbol{Cost}(s,a)$ (i.e., $action\rightarrow state\rightarrow cost$), and the projection boundary of the feasible-state/action-region is often estimated from offline sampled trajectories and remains fixed during training. Although online expert demonstration and other prior environmental explorations have been attempted, to the best of our knowledge, \textbf{\textit{none}} has considered \textbf{\textit{Dynamic Constrained RL}} that enforces the constraint to be directly \textbf{\textit{online}} coupled with (adapted to) the action impact, or the subsequent influence after executing a certain action. Consequently, existing works are prone to action instability or reward inaccessibility, which limits the policy's ability to learn how to handle collisions.
\bssection{Details on Our Proposed Brain-SAD}
\label{sec:3}

In sum, the key motivation for innovating on existing Constrained RL works on Safe AD is that constraints should be imposed more \textit{dynamically} across different interaction scenarios, that is, by constructing a \textbf{\textit{dynamic constraint}} that more explicitly reveals the difference in safety requirements, especially as caused by the adopted action. For this purpose, \textit{policy selection}, \textit{policy execution} and \textit{policy optimization} should be considered as a whole.

\subsection{Current-Scene Imagination for Dual Policy Selection}
\label{sec:3.1}

\begin{figure*}[tbp]
\setlength{\abovecaptionskip}{0.05cm}
\setlength{\belowcaptionskip}{0pt}
\centering
\includegraphics[width=0.9\textwidth]{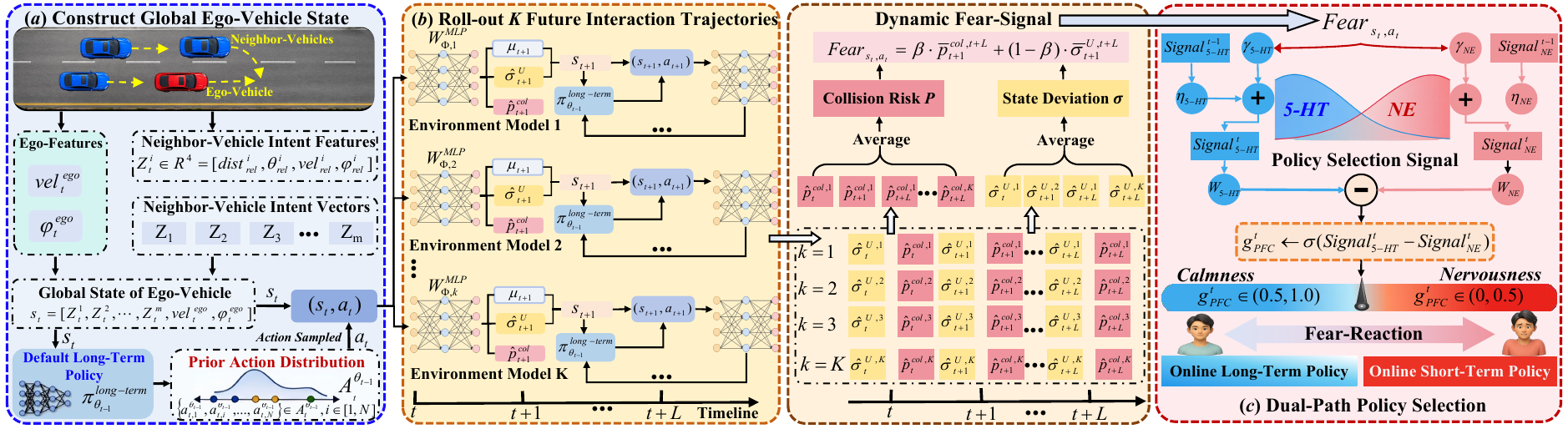}
\caption{Current-Scene Imagination for Dual Policy Selection (\textbf{NOTE}: We use $\boldsymbol{K}$ parameter-independent yet structurally identical multi-layer MLPs as the Ensemble Environment Model; a multi-layer MLP as the \textbf{default} long-term policy (\textbf{Actor}) network).}
\label{Fig:3}
\vspace{-0.05cm} 
\end{figure*}

Inspired by the brain-cognition mechanism, the main controlled \textit{ego-vehicle} should perceive interaction risk from the current scene, before it determines or selects the appropriate action policy. In Fig. \ref{Fig:3}, we incorporate the intent vectors of neighboring cars into ego-state as \textit{perception} based on which $L$-step future \textit{ego-neighbor} interaction trajectory will be rolled out or imagined from the current scene to derive \textit{dynamic fear-signal}, then activating \textit{Norepinephrine} (\textbf{NE})-signal \& \textit{Serotonin} (\textbf{5-HT})-signal as interaction \textit{risk} \& \textit{safety} intensity (i.e., inspired by Amygdala neurotransmitter release \cite{32}), so as to aggregate the final policy selection signal (i.e., inspired by \textit{fear-oriented} Amygdala$\rightarrow$Prefrontal Cortex pathway \cite{33}).

\textbf{First, we construct the global ego-state of the controlled \textit{ego-vehicle}} (see Fig. \ref{Fig:3} (\textit{a})). At the current $t$ step, in order to directly incorporate the \textit{interaction intent} of neighbouring vehicles into the ego-state $s_t$, in eq (1) we extract relative distance $dist_{rel}^{i,t}$, bearing angle $\theta_{rel}^{i,t}$, velocity $vel_{rel}^{i,t}$ and heading angle $\varphi_{rel}^{i,t}$ from \textit{each} neighboring \textit{vehicle} (i.e., $i\in[1,m]$), denoted as the $i$-th intent vector $Z_t^i\in\mathrm{R}^4$. Consequently, the \textbf{global} ego-state $s_t$ of the \textit{ego-vehicle} consists of $m$ neighboring intent vectors in all, along with its own velocity $vel_t^{ego}$ and heading angle $\varphi_t^{ego}$.

\begingroup
\bsleftformulasetup
\begin{equation}
\left\{
\begin{aligned}
&s_t=[Z_t^1,\ldots,Z_t^i,\ldots,Z_t^m,vel_t^{ego},\varphi_t^{ego}],\,i\in[1,m]
\\
&Z_t^i\in\mathrm{R}^4
=[dist_{rel}^{i,t},\theta_{rel}^{i,t},vel_{rel}^{i,t},\varphi_{rel}^{i,t}]
\end{aligned}
\right.
\label{bs:eq:1}
\end{equation}
\endgroup

\textbf{Second, we roll out future interaction trajectories from the current scene} (see Fig. \ref{Fig:3} (\textit{b})). Based on eq (1), since the current ego-state $s_t$ has intent vectors on $\boldsymbol{m}$ neighbors, thus in eq (2) we roll out future interaction trajectories (scene imagination) to \textbf{\textit{reveal}} the entailed interaction risk. At the current $t$ step, we invoke the \textbf{\textit{default}} long-term policy $\pi_{\theta_{t-1}}^{long-term}$ (see Section 3.2) to generate the \textbf{\textit{prior}} action distribution $A_t^{\theta_{t-1}}$ of the ego-vehicle in eq (2) (\textbf{NOTE}: the \textbf{\textit{default}} long-term policy $\pi_{\theta_{t-1}}^{long-term}$ is still in old-parameter $\theta_{t-1}$, not updated yet), from which \textbf{\textit{one}} and \textbf{\textit{only one}} \textbf{\textit{prior action}} $a_t^{\theta_{t-1}}\in A_t^{\theta_{t-1}}$ is sampled. Then, $\boldsymbol{K}$ structurally identical yet parameter-independent multi-layer MLPs or Ensemble Environment Model ($\pi_{\Phi_{k:1\rightarrow K}}^{MLP}$) \cite{26}, take in the sampled action $a_t^{\theta_{t-1}}$ and roll-out (or imagine) $\boldsymbol{K}$ trajectories over future $L$-step from $(s_t,a_t^{\theta_{t-1}})$.

\begingroup
\bsleftformulasetup
\begin{equation}
\left\{
\begin{aligned}
&a_{t+1}^{\theta_{t-1}}\sim\hat A_{t+1}^{\theta_{t-1}}\in\mathrm{R}^{dim}
\leftarrow\pi_{\theta_{t-1}}^{long-term}(\hat s_{t+1})\\
&\hat s_{t+1}\sim N(\hat\mu_{t+1},(\hat\sigma_{t+1}^{U})^2)\\
&(\hat\mu_{t+1},\hat\sigma_{t+1}^{U},\hat p_{t+1}^{col})
\leftarrow W_{\Phi,k}^{MLP}(s_t,a_t^{\theta_{t-1}}),k\in[1,K]\\
&a_t^{\theta_{t-1}}\sim A_t^{\theta_{t-1}}\in\mathrm{R}^{dim}
\leftarrow\pi_{\theta_{t-1}}^{long-term}(s_t)
\end{aligned}
\right.
\label{bs:eq:2}
\end{equation}
\endgroup

Specifically, in eq. (\ref{bs:eq:2}), the $k$-th environment model takes in 
the current intent-entailed ego-state $s_t$ and the sampled action 
$a_t^{\theta_{t-1}}$ of the ego-vehicle, then predicts the next 
\textbf{\textit{potential state space}} of the ego-vehicle, consisting of 
the mean state shift $\hat\mu_{t+1}$, the standard state deviation 
$\hat\sigma_{t+1}^{U}$ and the collision risk $\hat p_{t+1}^{col}$, from 
which the next most possible ego-vehicle state is estimated as 
$\hat s_{t+1}\sim N(\hat\mu_{t+1},(\hat\sigma_{t+1}^{U})^2)$ via a Gaussian 
distribution. In this way, standing at the predicted state $\hat s_{t+1}$, 
the scene imagination continues up to $L$ steps, outputting the $k$-th 
imagination-trajectory like 
$a_{t+1}^{\theta_{t-1}}\in\hat A_{t+1}^{\theta_{t-1}}\in\mathrm{R}^{dim}
\leftarrow\pi_{\theta_{t-1}}^{long-term}(\hat s_{t+1})$, then
{\((\hat s_{t+1},a_{t+1}^{\theta_{t-1}})\xrightarrow{W_{\Phi,k}^{MLP}}\allowbreak
(\hat\mu_{t+2},\hat\sigma_{t+2}^{U},\hat p_{t+2}^{col})\rightarrow
\allowbreak(\hat s_{t+2},a_{t+2}^{\theta_{t-1}})\rightarrow\allowbreak\cdots\allowbreak \xrightarrow{W_{\Phi,k}^{MLP}}\allowbreak(\hat\mu_{t+L},\hat\sigma_{t+L}^{U},\hat p_{t+L}^{col})
\rightarrow\allowbreak(\hat s_{t+L},a_{t+L}^{\theta_{t-1}})\).}
\begingroup
\bsleftformulasetup

\begin{equation}
\left\{
\begin{aligned}
&Fear_{s_t,a_t^{\theta_{t-1}}}\in\mathrm{R}
=\beta\cdot\bar p_{t+1}^{col,t+L}
+(1-\beta)\cdot\bar\sigma_{t+1}^{U,t+L}\\
&(\bar p_{t+1}^{col,t+L},\bar\sigma_{t+1}^{U,t+L})
\leftarrow\frac{1}{K}\sum_{k=1}^{K}\sum_{l=t+1}^{t+L}W_{\Phi,k}^{MLP}(s_l,a_l)
\end{aligned}
\right.
\label{bs:eq:3}
\end{equation}
\endgroup

All $\boldsymbol{K}$ of the $L$-step imagination-trajectories output by the
$\boldsymbol{K}$ environment models are collected and averaged into
$\bigl(\bar p_{\bsSub{t+1}}^{\bsSup{col,t+L}},
\bar\sigma_{\bsSub{t+1}}^{\bsSup{U,t+L}}\bigr)$
(see eq. (3)), with
$\bar p_{\bsSub{t+1}}^{\bsSup{col,t+L}}$
as overall collision risk and
$\bar\sigma_{\bsSub{t+1}}^{\bsSup{U,t+L}}$
as overall state deviation (from $t+1$ to $t+L$).
Then, in eq. (3), by aggregating the overall collision risk
$\bar p_{\bsSub{t+1}}^{\bsSup{col,t+L}}$
and overall state deviation
$\bar\sigma_{\bsSub{t+1}}^{\bsSup{U,t+L}}$
via $\beta\in[0,1]$, we derive the
\textbf{\textit{dynamic fear-signal}}
$Fear_{\bsSub{s_t,a_t^{\theta_{t-1}}}}$
as the overall ego-vehicle \textit{fear}-reaction at the current
$(s_t,a_t^{\theta_{t-1}})$.
This \textbf{\textit{dynamic fear-signal}}
$Fear_{\bsSub{s_t,a_t^{\theta_{t-1}}}}$
thus varies across different $L$-step scene-imaginations when given different
$(s_t,a_t^{\theta_{t-1}})$s.

\textbf{Third, we competitively select long/short-term policy online} (see Fig. \ref{Fig:3} (\textit{c})). Based on the above \textbf{\textit{dynamic fear-signal}} $Fear_{s_t,a_t^{\theta_{t-1}}}$ at the current \textbf{\textit{$t$-th}} time step, in eq (4), we induce the \textbf{\textit{increment}} on the \textit{Norepinephrine} (\textbf{NE})-signal as $\gamma_{NE}\cdot Fear_{s_t,a_t^{\theta_{t-1}}}$ for \textbf{\textit{nervousness}} alteration, and the opposite on the \textit{Serotonin} (\textbf{5-HT})-signal as $\gamma_{5-HT}\cdot(1-Fear_{s_t,a_t^{\theta_{t-1}}})$ for \textbf{\textit{calmness}} alteration. These two signal-increments $\gamma_{NE}\cdot Fear_{s_t,a_t^{\theta_{t-1}}}$ \& $\gamma_{5-HT}\cdot(1-Fear_{s_t,a_t^{\theta_{t-1}}})$ are aggregated with the former $Signal_{NE}^{t-1}$ \textbf{\&} $Signal_{5-HT}^{t-1}$ at \textbf{\textit{t-1}} time step, resulting in the latest $Signal_{NE}^{t}$ \textbf{\&} $Signal_{5-HT}^{t}$ at the current $t$-th time step.

\begingroup
\bsleftformulasetup
\begin{equation}
\left\{
\begin{aligned}
&Signal_{NE}^{t}\in\mathrm{R}
\leftarrow
\eta_{NE}\cdot Signal_{NE}^{t-1}
+\gamma_{NE}\cdot Fear_{s_t,a_t^{\theta_{t-1}}}
\\
&Signal_{5-HT}^{t}\in\mathrm{R}
\leftarrow
\eta_{5-HT}\cdot Signal_{5-HT}^{t-1}
\\
&\hspace{3.2cm}
+\gamma_{5-HT}\cdot
\bigl(1-Fear_{s_t,a_t^{\theta_{t-1}}}\bigr)
\\
&g_{PFC}^{t}\in[0,1]
\leftarrow
\sigma\Big(
W_{5-HT}\cdot Signal_{5-HT}^{t}
\\
&\hspace{3.2cm}
-W_{NE}\cdot Signal_{NE}^{t}
+bias
\Big)
\end{aligned}
\right.
\label{bs:eq:4}
\end{equation}
\endgroup

In eq (4), the latest $Signal_{NE}^{t}$ represents the interaction-risk intensity between the ego- and neighbor-vehicles induced by $(s_t,a_t^{\theta_{t-1}})$, whereas $Signal_{5-HT}^{t}$ represents the interaction-safety intensity. These two signals are fused via subtraction as $\sigma(Signal_{5-HT}^{t}-Signal_{NE}^{t})$ ($\sigma$ for \textit{sigmoid}), depicting the competition between interaction \textit{risk} \& \textit{safety}, and deriving the \textit{online} dual-path policy selection signal $g_{PFC}^{t}\in(0,1)$. The parameters $\{\eta_{NE},\gamma_{NE},\eta_{5-HT},\gamma_{5-HT}\}$ in eq (4) are set as \textbf{\textit{trainable}} variables, rendering policy-selection signal $g_{PFC}^{t}$ \textbf{\textit{learnable}} according to exploration feedback (see eq (17) in Section 3.4).

Obviously, by scene imagination from $(s_t,a_t^{\theta_{t-1}})$ via eq (2--4), when having \textbf{\textit{lower-risk}} interaction between ego- and neighbor-vehicles, it will bring about \textbf{\textit{lower}} dynamic \textit{fear-signal} $Fear_{s_t,a_t^{\theta_{t-1}}}\downarrow$, \textbf{\textit{lower}} risk intensity $Signal_{NE}^{t}\downarrow$, \textbf{\textit{higher}} safety intensity $Signal_{5-HT}^{t}\uparrow$, with $Signal_{5-HT}^{t}-Signal_{NE}^{t}>0$ (i.e., \textbf{\textit{calmness $>$ nervousness}}) and $g_{PFC}^{t}\rightarrow(0.5,1)$, where \textbf{Online Long-Term Dynamic Fear-Constrained Policy} (see Section 3.2) for regular scene-evolving will be selected. Otherwise, the \textbf{Online Short-Term Dynamic Fear-Projected Policy} (see Section 3.3) for urgent-collision defense will be selected with $Signal_{5-HT}^{t}-Signal_{NE}^{t}<0$ and $g_{PFC}^{t}\rightarrow(0,0.5)$. It should be noted that during policy selection in Fig. \ref{Fig:3}, we actually use the \textbf{\textit{default}} \textit{long-term} policy in \textbf{\textit{old}}-parameter $\theta_{t-1}$ (\textbf{\textit{not $\theta_t$}}). The \textbf{\textit{prior}} action $a_t^{\theta_{t-1}}\leftarrow\pi_{\theta_{t-1}}^{long-term}(s_t)$ sampled for scene imagination in eq (2) is \textbf{\textit{not}} the final action $a_t^{\theta_t}$ to be adopted before executing and updating the corresponding policy from $\theta_{t-1}\rightarrow\theta_t$.

\subsection{Online Long-Term Dynamic Fear-Constrained Policy}
\label{sec:3.2}
\vspace{-4pt}

\begin{figure*}[tbp]
\setlength{\abovecaptionskip}{0.05cm}
\setlength{\belowcaptionskip}{0pt}
\centering
\includegraphics[width=0.9\textwidth]{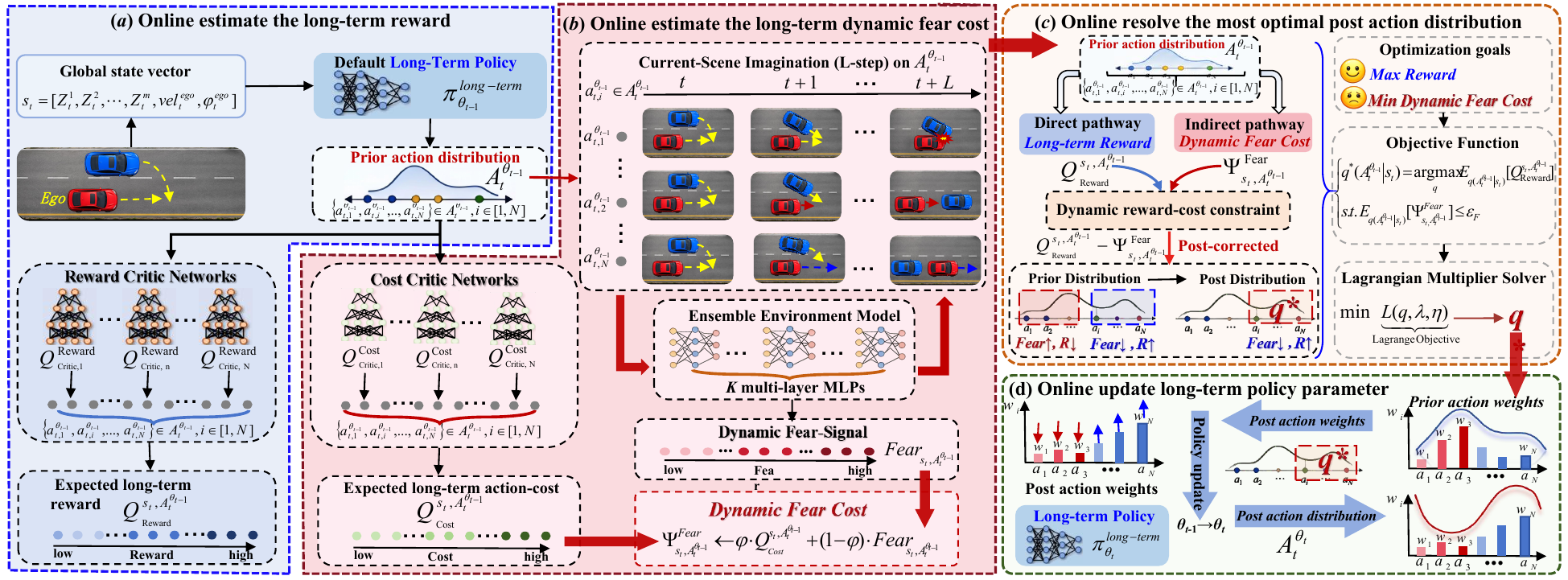}
\caption{Online Long-Term Dynamic Fear-Constrained Policy (\textbf{NOTE}: We reuse the same $\boldsymbol{K}$ multi-layer MLPs mentioned in Fig. \ref{Fig:3} as the Ensemble Environment Model in Fig. \ref{Fig:4}; a multi-layer MLP as long-term policy (\textbf{Actor}) network).}
\label{Fig:4}
\vspace{-0.05cm} 
\end{figure*}

In terms of the controlled \textit{ego-vehicle} in state $s_t$ at the current $t$ step, it executes \textbf{Online Long-Term Dynamic Fear-Constrained Policy} when $g_{PFC}^{t}\in(0.5,1.0)$ (\bsblue{see Fig. \ref{Fig:4}}) to deal with regular scene-evolving during vehicle interaction. In Particularly, the \textbf{\textit{prior action distribution}} $A_t^{\theta_{t-1}}\sim\pi_{\theta_{t-1}}^{long-term}(s_t)$ output by the above current-scene imagination (\bsblue{see Fig. \ref{Fig:3}}) is reused and estimated into the overall expected reward
\raisebox{0.05ex}{\scalebox{0.96}{$Q_{Reward}^{s_t,A_t^{\theta_{t-1}}}$}},
along with the corresponding \textbf{\textit{dynamic fear cost}}
\raisebox{0.05ex}{\scalebox{0.90}{$\Psi_{s_t,A_t^{\theta_{t-1}}}^{Fear}$}},
still via $L$-step scene imagination yet oriented to the whole \textbf{\textit{prior}} distribution $A_t^{\theta_{t-1}}$, \textbf{not} a single sampled action $a_t^{\theta_{t-1}}\in A_t^{\theta_{t-1}}$. Then, inspired by the Direct/Indirect action control pathway in Striatum \cite{14}, the \textbf{\textit{prior}} distribution $A_t^{\theta_{t-1}}$ is \textbf{\textit{post-corrected}} via the \textbf{\textit{dynamic constraint}} constructed from the reward
\raisebox{0.05ex}{\scalebox{0.96}{$Q_{Reward}^{s_t,A_t^{\theta_{t-1}}}$}}
and \textbf{\textit{dynamic fear cost}}
\raisebox{0.05ex}{\scalebox{0.96}{$\Psi_{s_t,A_t^{\theta_{t-1}}}^{Fear}$}},
and further resolved into the most optimal \textbf{\textit{posterior}} action distribution $q^*\sim\pi_{\theta_{t-1}}^{long-term}(A_t^{\theta_{t-1}}\mid s_t)$. This $q^*$ is finally projected onto a set of \textbf{\textit{post}} action weights that constrain the policy's action distribution, updating the policy parameter ($\theta$) \textbf{\textit{online}} as $A_t^{\theta_{t-1}}\sim\pi_{\theta_{t-1}}^{long-term}(s_t)\rightarrow A_t^{\theta_t}\sim\pi_{\theta_t}^{long-term}(s_t)$.

\begingroup
\bsleftformulasetup
\begin{equation}
\left\{
\begin{aligned}
&Q_{Reward}^{s_t,A_t^{\theta_{t-1}}}
=\frac{1}{N}\sum_{i=1}^{N}Q_{Critic,i}^{Reward}(s_t,A_t^{\theta_{t-1}})\\
&A_t^{\theta_{t-1}}
=\{a_{t,1}^{\theta_{t-1}},a_{t,2}^{\theta_{t-1}},\ldots,a_{t,N}^{\theta_{t-1}}\}
\leftarrow\pi_{\theta_{t-1}}^{long-term}(s_t)
\end{aligned}
\right.
\label{bs:eq:5}
\end{equation}
\endgroup

\textbf{First, we }\textbf{\textit{online }}\textbf{estimate the long-term reward of }\textbf{\textit{ego-vehicle}}\textbf{ in the current scene} (see Fig. \ref{Fig:4} (\textit{a})). When it comes to the ego-vehicle with state $s_t$ in the current interaction scene, in \bsblue{eq (5)}, we reuse the \textbf{\textit{prior}} action distribution $A_t^{\theta_{t-1}}=\{a_{t,1}^{\theta_{t-1}},a_{t,2}^{\theta_{t-1}},\ldots,a_{t,N}^{\theta_{t-1}}\}$, $a_{t,i}^{\theta_{t-1}}\in A_t^{\theta_{t-1}}$, $i\in[1,N]$ of the ego-vehicle, output by the \textbf{\textit{default}} long-term policy $\pi_{\theta_{t-1}}^{long-term}(s_t)$ in the aforementioned scene imagination (\bsblue{see Fig. \ref{Fig:3} \& eq (2)}, \textbf{NOTE:} in old-parameter $\boldsymbol{\theta}_{\boldsymbol{t-1}}$). Then, $N$ parameter-independent yet structurally identical \textit{Reward Critic} networks (i.e., each is a multi-layer MLP) are adopted to \textbf{\textit{online}} estimate the expected \textit{long-term} reward of $A_t^{\theta_{t-1}}$, denoted as $\pi_{\theta_{t-1}}^{long-term}(s_t)\rightarrow A_t^{\theta_{t-1}}\rightarrow Q_{Reward}^{s_t,A_t^{\theta_{t-1}}}(s_t,A_t^{\theta_{t-1}})$. Obviously, starting from different $(s_t,A_t^{\theta_{t-1}})$s, this long-term reward $Q_{Reward}^{s_t,A_t^{\theta_{t-1}}}$ estimated in \bsblue{eq (5)}, varies with the interaction dynamics.

\begin{figure*}[tp]
\begingroup
\bswideformulasetup
\begin{equation}
\left\{
\begin{alignedat}{1}
&\Psi_{s_t,A_t^{\theta_{t-1}}}^{Fear}
\leftarrow\varphi\cdot Q_{Cost}^{s_t,A_t^{\theta_{t-1}}}
 +(1-\varphi)\cdot Fear_{s_t,A_t^{\theta_{t-1}}}\\
&Q_{Cost}^{s_t,A_t^{\theta_{t-1}}}
=\frac{1}{N}\sum_{i=1}^{N}Q_{critic,i}^{Cost}
(s_t,A_t^{\theta_{t-1}})\\&Fear_{s_t,A_t^{\theta_{t-1}}}
\leftarrow\left\{a_{t,i}^{\theta_{t-1}}\in A_t^{\theta_{t-1}}\ \middle|\beta\cdot p_{t+1}^{col,t+L}
 +(1-\beta)\cdot\sigma_{t+1}^{U,t+L}=\frac{1}{K}\sum_{k=1}^{K}\sum_{t+1}^{t+L}
 W_{\Phi,k}^{MLP}(s_t,a_{t,i}^{\theta_{t-1}})\right\}\\
&s_t=[Z_t^1,\ldots,Z_t^i,\ldots,Z_t^m,vel_t^{ego},\varphi_t^{ego}],i\in[1,m]
\end{alignedat}
\right.
\label{bs:eq:6}
\end{equation}
\endgroup
\end{figure*}

\textbf{Second, we }\textbf{\textit{online }}\textbf{estimate the long-term }\textbf{\textit{dynamic fear cost }}\textbf{of }\textbf{\textit{ego-vehicle}}\textbf{ in the current scene} (see Fig. \ref{Fig:4} (\textit{b})). For the prior action distribution $A_t^{\theta_{t-1}}$, in \bsblue{eq (6)}, we further roll out $N$ future interaction trajectories over an $L$-step horizon for the ego-vehicle. Particularly, each trajectory rolled-out by $K$ multi-layer MLPs corresponds to one candidate action $a_{t,i}^{\theta_{t-1}}\in A_t^{\theta_{t-1}}$, $i\in[1,N]$, resulting in the overall collision risk \& state deviation $(\bar p_{t+1}^{col,t+L},\bar\sigma_{t+1}^{U,t+L})$ of each action, which are used to derive the \textbf{\textit{dynamic fear-signal}} $Fear_{s_t,A_t^{\theta_{t-1}}}=\{Fear_{s_t,a_{t,1}^{\theta_{t-1}}},\ldots,Fear_{s_t,a_{t,N}^{\theta_{t-1}}}\}$ over the entire prior distribution $A_t^{\theta_{t-1}}$. Then, $N$ \textit{Cost Critic} networks (i.e., each is a multi-layer MLP) \cite{05,06} are adopted to estimate the long-term action-cost $Q_{Cost}^{s_t,A_t^{\theta_{t-1}}}$ of $A_t^{\theta_{t-1}}$, which is further fused with the above \textbf{\textit{dynamic fear-signal}} $Fear_{s_t,A_t^{\theta_{t-1}}}$ to \textbf{\textit{online}} estimate the \textit{long-term} \textbf{\textit{dynamic fear cost}} $\Psi_{s_t,A_t^{\theta_{t-1}}}^{Fear}$ of the prior action distribution $A_t^{\theta_{t-1}}$. Obviously, the \textit{long-term} \textbf{\textit{dynamic fear cost}} $\Psi_{s_t,A_t^{\theta_{t-1}}}^{Fear}$ in \bsblue{eq (6)} can reflect different future cumulative \textbf{\textit{action-costs}} coupled with the corresponding \textbf{\textit{fear-signals}}, as the overall fear reaction of the ego-vehicle when given different $(s_t,A_t^{\theta_{t-1}})$s.

\begin{figure*}[tp]
\begingroup
\bswideformulasetup
\begin{equation}
\left\{
\begin{alignedat}{1}
&q^*(A_t^{\theta_{t-1}}\mid s_t)
=\operatorname*{argmax}_{q}
E_{q(A_t^{\theta_{t-1}}\mid s_t)}
[Q_{Reward}^{s_t,A_t^{\theta_{t-1}}}]\\
&\text{s.t.}\; E_{q(A_t^{\theta_{t-1}}\mid s_t)}
[\Psi_{s_t,A_t^{\theta_{t-1}}}^{Fear}]\leq\varepsilon_F,\; D_{KL}\!\left[q(A_t^{\theta_{t-1}}\mid s_t)\middle\|\pi_{\theta_{t-1}}^{long-term}
(A_t^{\theta_{t-1}}\mid s_t)\right]\leq\varepsilon_{KL}\\
&\int q(A_t^{\theta_{t-1}}\mid s_t)\,dA_t^{\theta_{t-1}}=1
\end{alignedat}
\right.
\label{bs:eq:7}
\end{equation}
\endgroup

\begingroup
\bswideformulasetup
\begin{equation}
\min\,
\underbrace{
L(q,\lambda,\eta)
}_{\smash{\text{Lagrange Objective}}}
\!=\!
-E_{A_t^{\theta_{t-1}}\!\sim\!q(\cdot\!\mid\!s_t)}
\!\left[
\smash[b]{%
\underbrace{
Q_{\text{Reward}}^{s_t,A_t^{\theta_{t-1}}}
-\lambda\!\cdot\!
\overbrace{
\left(
\Psi_{s_t,A_t^{\theta_{t-1}}}^{\text{Fear}}
-\varepsilon_F
\right)
}^{\smash{\text{Dynamic Fear Cost}}}
}_{\smash{\text{Brain-Inspired Dynamic Constraint}}}
}
\right]
\!+\!
\eta\!\left\{
D_{KL}\!\left[
q(A_t^{\theta_{t-1}}\!\mid\!s_t)
\!\middle\|\!
\pi_{\theta_{t-1}}^{\text{long-term}}\!
(A_t^{\theta_{t-1}}\!\mid\!s_t)
\right]
-\varepsilon_{KL}
\right\}
\label{bs:eq:8}
\end{equation}
\endgroup

\end{figure*}

\textbf{Third, we }\textbf{\textit{online }}\textbf{resolve the most optimal }\textbf{\textit{posterior }}\textbf{action distribution} $q^*(A_t^{\theta_{t-1}}\mid s_t)\sim\pi_{\theta_{t-1}}^{long-term}(A_t^{\theta_{t-1}}\mid s_t)$ (see Fig. \ref{Fig:4}(\textit{c})). Obviously, it is expected that the \textbf{\textit{prior action distribution}} $A_t^{\theta_{t-1}}\sim\pi_{\theta_{t-1}}^{long-term}(s_t)$ output by the long-term policy, should maximize long-term reward $Q_{Reward}^{s_t,A_t^{\theta_{t-1}}}$ in \bsblue{eq (5)} while minimizing the long-term \textbf{\textit{dynamic fear cost}} $\Psi_{s_t,A_t^{\theta_{t-1}}}^{Fear}\leq\varepsilon_F$ in \bsblue{eq (6)}. Therefore, the most optimal \textbf{\textit{posterior action distribution}} $q^*(A_t^{\theta_{t-1}}\mid s_t)$ can be derived from the \textbf{\textit{prior}} distribution $\pi_{\theta_{t-1}}^{long-term}(A_t^{\theta_{t-1}}\mid s_t)$ as the Constrained Optimization Problem defined in \bsblue{eq (7)} \cite{05}, where the KL-divergence trust region $D_{KL}[\cdot]\leq\varepsilon_{KL}$ prevents a dramatic policy shift from the \textbf{\textit{prior}} $\pi_{\theta_{t-1}}^{long-term}(A_t^{\theta_{t-1}}\mid s_t)$ to the \textbf{\textit{posterior}} $q(A_t^{\theta_{t-1}}\mid s_t)$, with the constant $\varepsilon_F$ as the upper limit on the \textbf{\textit{dynamic fear cost}} $\Psi_{s_t,A_t^{\theta_{t-1}}}^{Fear}$.

Next, by introducing a \textit{dynamic fear cost factor} $\lambda\geq0$ and KL trust region \textit{factor} $\eta>0$, we convert the constrained optimization in \bsblue{eq (7)} into the equivalent \textit{Lagrange Objective} in \bsblue{eq (8)} \cite{06}. Here, in \bsblue{eq (8)}, inspired by the action-control mechanism of the Striatum in our human brain that the Direct pathway selects optimal action with higher reward, while the Indirect pathway avoids action with higher risk (or cost) \cite{14}, we construct a brain-inspired \textbf{\textit{dynamic reward-cost constraint}} at action-level as $Q_{Reward}^{s_t,A_t^{\theta_{t-1}}}-\lambda\cdot(\Psi_{s_t,A_t^{\theta_{t-1}}}^{Fear}-\varepsilon_F)$, depicting the trade-off between long-term reward and long-term dynamic fear cost of action. In \bsblue{eq (8)}, since the dynamic fear cost $\lambda\cdot(\Psi_{s_t,A_t^{\theta_{t-1}}}^{Fear}-\varepsilon_F)$ alters across different ego-neighbor vehicle interactions, the reward-cost constraint $Q_{Reward}^{s_t,A_t^{\theta_{t-1}}}-\lambda\cdot(\Psi_{s_t,A_t^{\theta_{t-1}}}^{Fear}-\varepsilon_F)$ is constructed dynamically adapted to different scenarios.

\begin{figure*}[tp]
\begingroup
\bswideformulasetup
\begin{equation}
\min\,
\underbrace{
L(q,\lambda,\eta)
}_{\smash{\text{Lagrange Objective}}}
\rightarrow
q^*(A_t^{\theta_{t-1}}\mid s_t)
=
\pi_{\theta_{t-1}}^{long-term}
(A_t^{\theta_{t-1}}\mid s_t)
\cdot
\exp\!\left(
\frac{
Q_{\text{Reward}}^{s_t,A_t^{\theta_{t-1}}}
-\lambda\cdot
\Psi_{s_t,A_t^{\theta_{t-1}}}^{\text{Fear}}
}{
\eta
}
\right)
\cdot
\frac{1}{Z(s_t)}
\label{bs:eq:9}
\end{equation}
\endgroup

\begingroup
\bswideformulasetup
\begin{equation}
\left\{
w_{t,1}^{\theta_{t-1}}\propto a_{t,1}^{\theta_{t-1}},\;
w_{t,2}^{\theta_{t-1}}\propto a_{t,2}^{\theta_{t-1}},\ldots,
w_{t,N}^{\theta_{t-1}}\propto a_{t,N}^{\theta_{t-1}}
\right\}
\leftarrow
\operatorname{softmax}\!\left(
q^*\!\left(A_t^{\theta_{t-1}}\mid s_t\right)
\right),\;
w_{t,i}^{\theta_{t-1}}
\propto
a_{t,i\in[1,N]}^{\theta_{t-1}}
\in A_t^{\theta_{t-1}}
\label{bs:eq:10}
\end{equation}
\endgroup
\end{figure*}

Furthermore, we use the Lagrange Multiplier to \textbf{\textit{online}} resolve the Stationary Point of the \textit{Lagrange Objective} in \bsblue{eq (8)}, obtaining the most optimal \textbf{\textit{posterior action distribution}} $q^*(A_t^{\theta_{t-1}}\mid s_t)$ defined in \bsblue{eq (9)}, with $Z(s_t)$ as the State-Dependent Partition Function for normalization \cite{34}. In this way, the \textbf{\textit{prior}} action distribution $A_t^{\theta_{t-1}}$ is \textbf{\textit{online}} corrected into the most optimal \textbf{\textit{posterior}} one $q^*(A_t^{\theta_{t-1}}\mid s_t)$. This resolved \textbf{\textit{posterior}} distribution $q^*(A_t^{\theta_{t-1}}\mid s_t)$ in \bsblue{eq (9)} is then projected onto $N$ \textbf{\textit{post action weights}} $w_{t,i}^{\theta_{t-1}}\propto a_{t,i}^{\theta_{t-1}}$, $i\in[1,N]$, for each candidate action $\{a_{t,1}^{\theta_{t-1}},a_{t,2}^{\theta_{t-1}},\ldots,a_{t,N}^{\theta_{t-1}}\}$ in $A_t^{\theta_{t-1}}$ (\bsblue{see eq (10)})\footnote{See full derivation on eq (7--10) in Appendix \textbf{B.1}.}.

\begin{figure*}[tp]
\begingroup
\bswideformulasetup
\begin{equation}
\mathit{Loss}_{\pi_{\theta}}^{\text{long-term}}(\theta)=-E_{s_t\sim \mathit{Buff}}\Bigg[\underbrace{\sum_{i=1}^{N}w_{t,i}^{\theta_{t-1}}\cdot \log\pi_{\theta}^{\text{long-term}}\!(a_{t,i}^{\theta_t}\mid s_t)}_{{\text{online posterior supervision from }\,q^*(A_t^{\theta_{t-1}}\mid s_t)}}+\beta\cdot KL_{s_t\sim \mathit{Buff}}\!\left[\pi_{\theta}^{\text{long-term}}\!(A_t^{\theta_t}\mid s_t)\middle\|\pi_{\theta_{t-1}}^{\text{long-term}}\!(A_t^{\theta_{t-1}}\mid s_t)
\right]\Bigg]
\label{bs:eq:11}
\end{equation}
\endgroup
\end{figure*}

\textbf{Finally, we }\textbf{\textit{online }}\textbf{update long-term policy parameter }$\theta$\textbf{ via the }\textbf{\textit{posterior }}$q^*(A_t^{\theta_{t-1}}\mid s_t)$ (see Fig. \ref{Fig:4} (\textit{d})). Obviously, the above $N$ \textbf{\textit{post action weights}} $w_{t,i}^{\theta_{t-1}}\propto a_{t,i}^{\theta_{t-1}}$, $i\in[1,N]$, in {eq (11)}, derived from the most optimal \textbf{\textit{posterior action distribution}} $q^*(A_t^{\theta_{t-1}}\mid s_t)$ in {eq (9)}, entailed the expected evolve direction of the long-term policy $\pi_{\theta_{t-1}}^{long-term}$. Therefore, the $Loss_{\pi_{\theta}}^{long-term}(\theta)$ in {eq (11)} takes the $N$ post action weights in {eq (10)} as the \textbf{\textit{online posterior}} supervision from $q^*$, so as to \textbf{\textit{post-constrain}} the action distribution of the long-term policy $\pi_{\theta}^{long-term}$, such as $w_{t,i}^{\theta_{t-1}}\cdot\log\pi_{\theta}^{long-term}(a_{t,i}^{\theta_t}\mid s_t)$, updating the parameter $\theta$ in $\pi_{\theta}^{long-term}(a_{t,i}^{\theta_t}\mid s_t)$: $\theta_{t-1}\rightarrow\theta_t$ within KL a trust region, to prevent a dramatic policy shift.

\begingroup
\setstretch{1.24}
\lineskiplimit=-1000pt

In sum, when given \textbf{\textit{prior action distribution}} 
$A_t^{\theta_{t-1}}\sim\pi_{\theta_{t-1}}^{long-term}(s_t)$, 
the long-term policy $\pi_{\theta}^{long-term}$ expects a higher reward 
$Q_{Reward}^{s_t,A_t^{\theta_{t-1}}}\uparrow$ in {eq (7)} 
alongside a lower \textbf{\textit{dynamic fear cost}} 
$\Psi_{s_t,A_t^{\theta_{t-1}}}^{Fear}\downarrow$ in {eq (8)}, 
altogether constructing the \textbf{\textit{dynamic reward-cost constraint}} 
$(Q_{Reward}^{s_t,A_t^{\theta_{t-1}}}
-\Psi_{s_t,A_t^{\theta_{t-1}}}^{Fear})$ 
to correct the \textbf{\textit{prior action distribution}} 
$A_t^{\theta_{t-1}}$ into the most optimal 
\textbf{\textit{posterior action distribution}} 
$q^*(A_t^{\theta_{t-1}}\mid s_t)$ in {eq (9)}, 
further projected onto \textbf{\textit{post action weights}} 
$w_{t,i}^{\theta_{t-1}}\propto 
a_{t,i\in[1,N]}^{\theta_{t-1}}\in A_t^{\theta_{t-1}}$ 
in {eq (10)} as the online supervision constraining 
the policy parameter update as 
$\pi_{\theta}^{long-term}:\theta_{t-1}\rightarrow\theta_t$ 
via the $Loss_{\pi_{\theta}}^{long-term}(\theta)$ in {eq (11)}.

\par
\endgroup

\subsection{Online Short-Term Dynamic Fear-Projected Policy}
\label{sec:3.3}

Since the long-term policy constrains action selection via \textbf{\textit{online}} policy optimization with a dynamic reward-cost constraint, it lacks the efficiency to deal with \textit{urgent-collision defense} (e.g., a sudden side-lane cut-in) \cite{07}. Consequently, to handle the \textbf{\textit{urgent braking}} under high risk more efficiently, we adopt the \textbf{Online Short-Term Dynamic Fear-Projected Policy} when $g_{PFC}^{t}\in(0,0.5)$. Inspired by the Hyper-direct pathway in Striatum \cite{15}, the original \textbf{\textit{risk action}} $a_t^{risk}\sim\pi_{\varphi}^{short-term}(s_t)$ output by the short-term policy\footnote{We use a multi-layer MLP as the short-term policy (Actor) network $\pi_{\varphi}^{short-term}$.} is directly projected as a reachable action set $A_{safe}(s_t)$, from which a safe action $a_t^{safe}$ with minimum action shift towards the original $a_t^{risk}$ is selected as substitute. Furthermore, without policy optimization, the above action projection is \textbf{\textit{online}} constrained by the \textbf{\textit{dynamic fear boundary}} $\Psi_{s_t,a_t}^{Fear}$, which tightens or loosens the reachable action set $A_{safe}(s_t)$ according to the \textit{risk intent} of the neighboring vehicles surrounding the ego-vehicle.

\par\begingroup
\small
\begin{equation}
\left\{
\begin{aligned}
&Neighbor(s_t)=\{i\mid dist_{rel}^{i,t}<dist_r,\\[-0.3ex]
&vel_{rel}^{i,t}>0,\;TTC^{i,t}<TTC_r,\;brake_{req}^{i,t}\},i\in[1,m]\\
&brake_{req}^{i,t}<0\leftarrow
\left(
-\frac{\max(vel_{rel}^{i,t},0)^2}
{2\max(dist_{rel}^{i,t}-dist_{min},\varepsilon)}
\right)
,\;\varepsilon>0\\
&TTC^{i,t}=
\frac{dist_{rel}^{i,t}}
{\max(vel_{rel}^{i,t},\varepsilon)},
\;\varepsilon>0\\[-0.3ex]
&(TTC^{i,t},vel_{rel}^{i,t},brake_{req}^{i,t})
\leftarrow Z_t^i\in s_t
\end{aligned}
\right.
\label{bs:eq:12}
\end{equation}
\endgroup

\textbf{First, we identify risk neighboring vehicles.} In \bsblue{eq (12)}, given the ego-vehicle state $s_t$ containing \textbf{\textit{m}} neighboring intent vectors $\{Z_t^1,Z_t^2,\ldots,Z_t^m\}$, $Z_t^i\in\mathrm{R}^4=[dist_{rel}^{i,t},\theta_{rel}^{i,t},vel_{rel}^{i,t},\varphi_{rel}^{i,t}]$ (\bsblue{see eq (1)}), we extract risk neighbor set $Neighbor(s_t)$ of the ego-vehicle: the $i$-th neighbor is included when its intent vector $Z_t^i$ has with relative distance $dist_{rel}^{i,t}<dist_r$, relative velocity $vel_{rel}^{i,t}>0$, and Time-To-Collision $TTC^{i,t}<TTC_r$. Moreover, in \bsblue{eq (12)}, each risk-neighbor vehicle in $Neighbor(s_t)$ is also been assigned a \textbf{\textit{negative}} braking demand towards ego-vehicle (i.e., $brake_{req}^{i,t}<0$). The \textbf{\textit{more}} negative $brake_{req}^{i,t}$, the \textbf{\textit{higher}} braking intensity.

\begin{figure*}[tp]
\begingroup
\bswideformulasetup
\begin{equation}
\left\{\begin{alignedat}{1}
&A_{safe}(s_t)=G_{proj}^{set}\otimes(a_t^{risk})^{T}=\{A_{proj},e_{brake}\}\otimes
\begin{pmatrix}
a_t^{steer}\\
a_t^{vertical}
\end{pmatrix}
\\[0.4ex]
&G_{proj}^{set}=\{A_{proj},e_{brake}\},\; A_{proj}=\begin{pmatrix}1&0\\0&1\end{pmatrix},\; e_{brake}=\begin{pmatrix}0&1\end{pmatrix}\\[0.4ex]
&a_t^{risk}=\bigl(a_t^{steer}\in\mathrm{R},a_t^{vertical}\in\mathrm{R}\bigr)\leftarrow
\pi_{\varphi}^{short-term}(s_t)
\end{alignedat}
\right.
\label{bs:eq:13}
\end{equation}
\endgroup

\begingroup
\bswideformulasetup
\begin{equation}
A_{safe}(s_t)
=G_{proj}^{set}\otimes(a_t^{risk})^T
\leftarrow
\Bigg\{
\overbrace{
\begin{pmatrix}
1 & 0\\
0 & 1
\end{pmatrix}
}^{A_{proj}}
\,
\overbrace{
\begin{pmatrix}
a_t^{steer}\\
a_t^{vertical}
\end{pmatrix}
}^{(a_t^{risk})^T}
=
\begin{pmatrix}
\Gamma_{t,steer}^{risk}\\
\Gamma_{t,vertical}^{risk}
\end{pmatrix},
\;
\overbrace{
\begin{pmatrix}
0 & 1
\end{pmatrix}
}^{e_{brake}}
\,
\overbrace{
\begin{pmatrix}
a_t^{steer}\\
a_t^{vertical}
\end{pmatrix}
}^{(a_t^{risk})^T}
=
\left(a_t^{vertical}\right)
\Bigg\}
\label{bs:eq:14}
\end{equation}
\endgroup
\end{figure*}

\textbf{Second, we }\textbf{\textit{online }}\textbf{project original risk action as reachable action set.} The short-term policy outputs the original two-dimensional \textbf{\textit{risk action}} $a_t^{risk}=(a_t^{steer},a_t^{vertical})\sim\pi_{\varphi}^{short-term}(s_t)$, where $a_t^{steer}\in\mathrm{R}$ is the \textbf{\textit{steering angle}} (negative for turning left, positive for right), and $a_t^{vertical}\in\mathrm{R}$ is the \textbf{\textit{vertical acceleration}} (negative for braking, positive for speeding). This original \textbf{\textit{risk action}} $a_t^{risk}=(a_t^{steer},a_t^{vertical})$ is not to be adopted immediately, instead it is projected as a reachable action set $A_{safe}(s_t)=G_{proj}^{set}\otimes(a_t^{risk})^T$ (\bsblue{see eq (13--14)}).

Specifically, in \bsblue{eq (13)}, we incorporate complex projection operator $G_{proj}^{set}=\{A_{proj},e_{brake}\}$, consisting of the action-dimension operator $A_{proj}$ and the vertical-braking operator $e_{brake}$, both applied element-wise to $(a_t^{risk})^T=(a_t^{steer},a_t^{vertical})^T$ as $A_{proj}\otimes(a_t^{risk})^T$ and $e_{brake}\otimes(a_t^{risk})^T$. These two terms are further computed via the dot-product in \bsblue{eq (14)}. In this way, by $A_{proj}\otimes(a_t^{risk})^T$, we project the entire $(a_t^{risk})^T$ restrictively onto $(a_t^{risk})^T\rightarrow\Gamma_{t,steer}^{risk}$ and $(a_t^{risk})^T\rightarrow\Gamma_{t,vertical}^{risk}$ from steering angle $a_t^{steer}$ and vertical acceleration $a_t^{vertical}$ aspects, while $e_{brake}\otimes(a_t^{risk})^T$ merely extracts the vertical acceleration $(a_t^{vertical})$.

\begin{figure*}[tp]
\begingroup
\bswideformulasetup
\begin{equation}
\left\{\!
\begin{alignedat}{1}
&\operatorname*{argmin}_{a_t^{safe}\in A_{safe}(s_t)}
\left\|a_t^{safe}-a_t^{risk}\right\|,
\; \text{s.t.}\;
\overbrace{
A_{safe}(s_t)
=
G_{proj}^{set}\otimes(a_t^{risk})^T
}^{\text{reachable action set projection}}
\leq
\underbrace{
\Psi_{s_t,a_t}^{\text{Fear}}
=
\left\{
\tau_{safe}=(-1,1)^T,\;
b_t^{req}<0
\right\}
}_{\text{Brain-Inspired Dynamic Constraint}}
\\
&b_t^{req}\in\mathrm{R}
=
\min\!\left(brake_{req}^{i,t}\right),
\;
brake_{req}^{i,t}<0
\propto Neighbor(s_t)
\end{alignedat}
\right.
\label{bs:eq:15}
\end{equation}
\endgroup
\end{figure*}

\textbf{Third, we }\textbf{\textit{online }}\textbf{construct dynamic fear boundary for safe action selection.} In order to ensure that the projected reachable action set $A_{safe}(s_t)=G_{proj}^{set}\cdot(a_t^{risk})^T$ reflect the different braking demands towards the ego-vehicle when facing interactions with different risk neighbor sets $Neighbor(s_t)$s across time steps (\bsblue{see eq (12)}), we \textbf{\textit{online}} construct a \textbf{\textit{dynamic fear boundary}} $\Psi_{s_t,a_t}^{Fear}$ for $A_{safe}(s_t)$ based on the current $Neighbor(s_t)$ in \bsblue{eq (15)}. This dynamic fear boundary $\Psi_{s_t,a_t}^{Fear}$ consists of the constant vector $\tau_{safe}=(-1,1)^T$ and $b_t^{req}<0$, where $\tau_{safe}$ constrains the action-dimension projection $(\Gamma_{t,steer}^{risk},\Gamma_{t,vertical}^{risk})$ in \bsblue{eq (14)} within value scope $(-1,1)^T$, and negative $b_t^{req}<0$ constrains the vertical-braking projection $e_{brake}\otimes(a_t^{risk})^T=(a_t^{vertical})$ in \bsblue{eq (14)} not to exceed the \textbf{\textit{most negative}} braking demand (or the lowest bound as $\min(brake_{req}^{i,t})$ in \bsblue{eq (15)}) among \textbf{\textit{all}} the vehicles across the entire \textit{risk neighbor set} $Neighbor(s_t)$, altogether constraining $A_{safe}(s_t)\leq\Psi_{s_t,a_t}^{Fear}=\{\tau_{safe},b_t^{req}\}$.

In this way, \bsblue{eq (15)} formally couples the \textbf{\textit{online}} reachable action set projection $A_{safe}(s_t)=G_{proj}^{set}\otimes(a_t^{risk})^T$ with the dynamic fear boundary constraint ($\Psi_{s_t,a_t}^{Fear}$). As the ego- and neighbor-vehicles interact over time, the \textit{risk-neighbor} set $Neighbor(s_t)$ derived from the ego-vehicle state $s_t$ alters correspondingly, leading to different lowest braking demand $b_t^{req}<0$ in $\Psi_{s_t,a_t}^{Fear}$, in turn, dynamically tightening or loosening the action set $A_{safe}(s_t)$. Finally, the most optimal safe action $a_t^{safe}$ with minimum shift from the original $a_t^{risk}$ (i.e., $\arg\min\|a_t^{safe}-a_t^{risk}\|$) is selected from $A_{safe}(s_t)\leq\Psi_{s_t,a_t}^{Fear}$ as the short-term policy's output, like $a_t^{risk}\rightarrow A_{safe}(s_t)=G_{proj}^{set}\cdot(a_t^{risk})^T\leq\Psi_{s_t,a_t}^{Fear}\rightarrow\arg\min(a_t^{safe})$, without policy optimization, for efficiency\footnote{See derivation on convergence of eq (15) in Appendix \textbf{B.2}.}.

\subsection{Online-to-Offline Optimization Loop}
\label{sec:3.4}

In fact, for the long-term policy, we construct the \textit{dynamic fear cost} $\Psi_{s_t,A_t^{\theta_{t-1}}}^{Fear}$ of the action in \bsblue{eq (6)}, then the derive dynamic action reward-cost constraint in \bsblue{eq (8)} (i.e., $Q_{Reward}^{s_t,A_t^{\theta_{t-1}}}-\lambda\cdot(\Psi_{s_t,A_t^{\theta_{t-1}}}^{Fear}-\varepsilon_F)$). For the short-term policy, we directly take the \textit{dynamic fear boundary} $\Psi_{s_t,a_t}^{Fear}$ in \bsblue{eq (15)} as the constraint. Based on these two \textbf{\textit{dynamic constraints}}, the long-term policy $\pi_{\theta}^{long-term}$ is \textbf{\textit{online}} optimized by the \textit{Lagrange Objective} in \bsblue{eq (9)} \& $Loss_{\pi_{\theta}}^{long-term}(\theta)$ in \bsblue{eq (11)}, and the action-projection of the short-term policy $\pi_{\varphi}^{short-term}$ is \textbf{\textit{online}} optimized by the minimum action shift ($\arg\min\|a_t^{safe}-a_t^{risk}\|$) in \bsblue{eq (15)}. In this way, the ego-vehicle can constantly interact with neighboring ones to generate sate-action-reward-signal trajectories such as $\{s_t,a_t,s_{t+1},r_t,Fear_{s_t,a_t},policy\_flag_t=long/short,y_t=0/1\}$s, all stored into \textbf{\textit{offline-buffer}} for replay (i.e., $Fear_{s_t,a_t}$ as the dynamic fear-signal in \bsblue{eq (3)}, $policy\_flag_t$ as policy-selection label, and $y_t$ as whether the trajectory ended with a collision (\textbf{1}) or not (\textbf{0})). All the above \textit{online} optimizations provide the explored experiences for \textit{offline} optimization.

\begin{figure*}[tp]
\begingroup
\bswideformulasetup
\begin{equation}
MLE\_Loss_{\Phi_k}
=-E_{(s_t,a_t,s_{t+1},r_t,y_t)\sim Buff}\Big[
\log N(s_{t+1};\hat\mu_{t+1},(\hat\sigma_{t+1}^{U})^2)
+\lambda\Big(
y_t\log\hat p_{t+1}^{col}
+(1-y_t)\log(1-\hat p_{t+1}^{col})
\Big)\Big]
\label{bs:eq:16}
\end{equation}
\endgroup
\end{figure*}

Based on the offline-buffer, the \textbf{Ensemble Environment Model} for scene imagination, involved in both the policy selection (\bsblue{see Fig. \ref{Fig:3}}) and the long-term policy (\bsblue{see Fig. \ref{Fig:4}}) is \textbf{\textit{first}} trained using $MLE\_Loss_{\Phi_k}$ defined in \bsblue{eq (16)}. Here, we expect the imagined trajectories to end with the same \textbf{\textit{true}} collision label ($y_t$) as in the buffer, where each imagination-step in the trajectory is enforced to fit the mean state shift $\hat\mu_{t+1}$ and standard state deviation $\hat\sigma_{t+1}^{U}$ via a Gaussian distribution, supervised by the \textbf{\textit{true}} next state $s_{t+1}$ in the buffer (i.e., $\log N(s_{t+1};\hat\mu_{t+1},(\hat\sigma_{t+1}^{U})^2)$, with $N$ denoting the Gaussian distribution).

\begin{figure*}[tp]
\begingroup
\bswideformulasetup
\begin{equation}
\begin{alignedat}{1}
L_{\substack{\Phi=\{\eta_{NE},\gamma_{NE},\\\eta_{5-HT},\gamma_{5-HT}\}}}^{\text{Selection}}
&=-E_{\mathit{Buff}}\Bigl[
\underbrace{\lambda_{safe}\|g_{PFC}^{\Phi,t}-(1-y_t)\|_{L_2}^{2}
+\lambda_{\text{Reward}}\|g_{PFC}^{\Phi,t}-(0.5\pm\tanh(r_t))_{policy\_flag_t}\|_{L_2}^{2}}_{{\text{Policy Selection}}}\\
&{}+\underbrace{\lambda_{NE}\|\mathit{Signal}_{NE}^{\Phi,t}-\min(1,1.5\!\cdot\!Fear_{s_t,a_t})\|_{L_2}^{2}
+\lambda_{5-HT}\|\mathit{Signal}_{5-HT}^{\Phi,t}-\max(0,1-1.5\!\cdot\!Fear_{s_t,a_t})\|_{L_2}^{2}}_{{\text{Neurotransmitter Alignment}}}
\Bigr]
\end{alignedat}
\label{bs:eq:17}
\end{equation}
\endgroup
\end{figure*}

Next, after the Ensemble Environment Model becomes reliable via \bsblue{eq (16)}, we \textbf{\textit{offline}} train the policy selection in \bsblue{eq (4)} using $L_{\Phi}^{Selection}$ in \bsblue{eq (17)}. In \bsblue{eq (17)}, after loading $\{r_t,Fear_{s_t,a_t},policy\_flag_t,y_t\}$ from the buffer, we initialize $Signal_{NE}^{\Phi,0}=Signal_{5-HT}^{\Phi,0}=0.5$ in \bsblue{eq (4)}, then recompute policy-selection process in \bsblue{eq (4)} to regenerated the signals $\{Signal_{NE}^{\Phi,t},Signal_{5-HT}^{\Phi,t},g_{PFC}^{\Phi,t}\}$ from $1\rightarrow t$ via 4 trainable parameters $\Phi=\{\eta_{NE},\gamma_{NE},\eta_{5-HT},\gamma_{5-HT}\}$ (\bsblue{see eq (4)}). Based on the above regenerate signals, we design 4 objective functions in \bsblue{eq (17)} including: $g_{PFC}^{\Phi,t}-(1-y_t)$ drives policy selection toward the \textbf{\textit{true}} collision label $y_t$; $g_{PFC}^{\Phi,t}-(0.5\pm\tanh(r_t))$ drives policy selection toward the \textbf{\textit{true}} long/short policy label $policy\_flag_t$ (reward $r_t$ as value projection factor), $Signal_{NE}^{\Phi,t}-\min(1,1.5\cdot Fear_{s_t,a_t})$ and $Signal_{5-HT}^{\Phi,t}-\max(0,1-1.5\cdot Fear_{s_t,a_t})$ collaboratively constrain the opposite polarity between $\langle Signal_{NE}^{\Phi,t},Signal_{5-HT}^{\Phi,t}\rangle$ based on the \textbf{\textit{true}} fear-signal $Fear_{s_t,a_t}$. The above 4 functions are back-propagated to optimize the policy-selection parameters $\Phi=\{\eta_{NE},\gamma_{NE},\eta_{5-HT},\gamma_{5-HT}\}$ \textbf{\textit{offline}}\footnote{See details on \textbf{4} signal-guidance objective functions in Appendix \textbf{B.4}.}.

\begin{figure*}[tp]
\begingroup
\bswideformulasetup
\begin{equation}
\left\{
\begin{alignedat}{1}
&Loss_{R,Critic}^{long-term}
=-E_{Buff}\Big[\big(Q_{Reward,Critic}^{long-term}(s_t,a_t)-\big(r_t+\gamma \cdot E_{a\sim\pi_{\theta_{old}}^{long-term}}[Q_{Reward,Critic}^{long-term,Target}(s_{t+1},a_{t+1})]
\big)\big)^2\Big]\\
&Loss_{C,Critic}^{long-term}
=-E_{Buff}\Big[\big(Q_{Cost,Critic}^{long-term}(s_t,a_t)-\big(c_t+\gamma \cdot E_{a\sim\pi_{\theta_{old}}^{long-term}}[Q_{Cost,Critic}^{long-term,Target}(s_{t+1},a_{t+1})]
\big)\big)^2\Big]
\end{alignedat}
\right.
\label{bs:eq:18}
\end{equation}
\endgroup

\begingroup
\bswideformulasetup
\begin{equation}
\left\{
\begin{alignedat}{1}
&Loss_{Critic}^{short-term}
=-E_{Buff}\Big[\big(Q_{Reward,Critic}^{short-term}(s_t,a_t^{safe})-\big(r_t+\gamma \cdot Q_{Reward,Critic}^{short-term,Target}
(s_{t+1},a_{t+1}^{safe})\big)\big)^2\Big]\\
&Loss_{Actor}^{short-term}
=-E_{Buff}\Bigg[(1-\beta)\cdot\underbrace{Q_{Reward,Critic}^{short-term}\big(s_t,\pi_{\varphi}^{short-term}(\hat a_t^{safe}\mid s_t)\big)}_{\text{Critic-to-Actor}}+\beta\cdot\underbrace{\|\pi_{\varphi}^{short-term}(\hat a_t^{safe}\mid s_t)-a_t^{safe}\|_{L_2}^{2}}_{\text{safety projection imitation}}\Bigg]
\end{alignedat}
\right.
\label{bs:eq:19}
\end{equation}
\endgroup
\end{figure*}

When policy selection becomes reliable, based on the trajectories with $policy\_flag=long$ in the buffer, we \textbf{\textit{offline}} train the \textbf{Reward \& Cost Critic Networks} of the long-term policy using $Loss_{R/C,Critic}^{Long-term}$ in \bsblue{eq (18)}. Here, following standard TD-error minimization \cite{30} via offline-buffer replay, the (true) trajectories are reloaded to train Reward \& Cost Critic Networks ($N$ multi-layer MLPs in Fig. \ref{Fig:4}), providing more estimates of the ego-vehicle's long-term reward in \bsblue{eq (5)} and the corresponding dynamic fear cost in \bsblue{eq (6)}.

Following TD-error minimization, we \textbf{\textit{offline}} train the Actor/Critic network in the short-term policy using $Loss_{Actor/Critic}^{short-term}$ in \bsblue{eq (19)}, based on the trajectories with $policy\_flag=short$ in the buffer. In \bsblue{eq (19)}, the safe action $a_t^{safe}$ is the action $a_t$ stored in buffer. After loading $\{s_t,a_t^{safe}\}$ from the buffer, $Loss_{Critic}^{short-term}$ trains \textbf{Reward Critic Networks} (\textbf{\textit{N}} multi-layer MLPs) to evaluate long-term reward of the safe action $a_t^{safe}$ in buffer. Then, based on Critic, $Loss_{Actor}^{short-term}$ trains the short-term policy (\textbf{Actor}) Network (a multi-layer MLP as $\pi_{\varphi}^{short-term}$) to directly generate a safe action by \textbf{\textit{offline imitation}} like $\pi_{\varphi}^{short-term}(\hat a_t^{safe}\mid s_t)\rightarrow a_t^{safe}$, skipping the action projection $A_{safe}(s_t)=G_{proj}^{set}\otimes(a_t^{risk})^T$ in \bsblue{eq (15)}, where the Critic is used to evaluate the long-term reward of the generated imitated action $\hat a_t^{safe}$.

\begin{figure*}[tp]
\begingroup
\bswideformulasetup
\begin{equation}
\min Loss_{\text{Brain-SAD}}^{\text{Offline}}
=
\lambda_1\cdot MLE\_Loss_{\Phi_k}
+\lambda_2\cdot L_{\Phi}^{\text{Selection}}
+\lambda_3\cdot Loss_{R/C,\text{Critic}}^{\text{Long-term}}
+\lambda_4\cdot Loss_{\text{Critic\&Actor}}^{\text{Short-term}}
\label{bs:eq:20}
\end{equation}
\endgroup
\end{figure*}

Finally, all the \textbf{\textit{offline}} optimizations in \bsblue{eq (16--19)} will be aggregated into \bsblue{eq (20)} with $\lambda_{1-4}\in[0,1]$, to update all the components across Brain-SAD, which in turn serves the next round of \textbf{\textit{online}} exploration and optimization (\bsblue{see eq (4, 11, 15)}) to generate new trajectories into the offline-buffer. In this way, an \textbf{\textit{Online-to-Offline}} optimization loop is formally enclosed (see full algorithm of Brain-SAD in Appendix \textbf{B.3}).

\section{Analysis on Experimental Results}
\label{sec:experimental_results}

\subsection{Configuration on Simulation Environment}
\label{sec:simulation_configuration}

\begin{figure}[tbp]
\setlength{\abovecaptionskip}{-0.05cm}
\setlength{\belowcaptionskip}{0pt}
\centering

\bsfigurefile{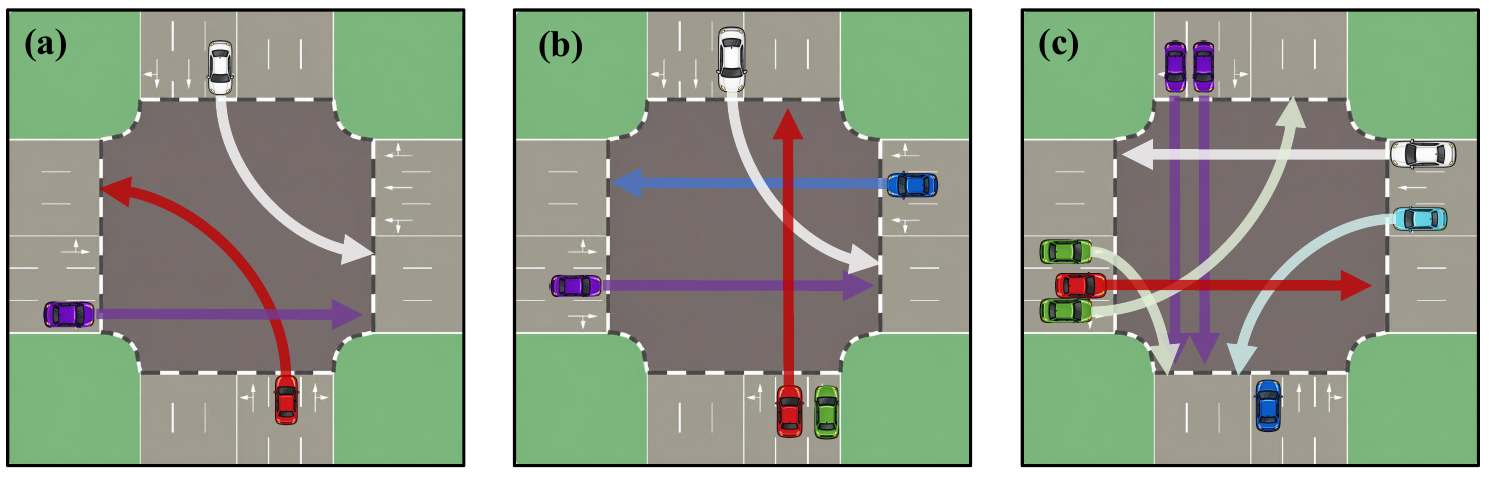}{\columnwidth}{3.4cm}

\caption{No traffic light intersection simulation with different complexity
(NOTE: \textbf{\textit{red}} as ego-vehicle;
\textbf{Scene A}: Turn left at intersection \textbf{\textit{only}} with
\textbf{\textit{multiple ahead}} interaction;
\textbf{Scene B}: Straight across at intersection with
\textbf{\textit{multiple ahead}} and \textbf{\textit{single aside}} interaction;
\textbf{Scene C}: Straight across at intersection with
\textbf{\textit{multiple ahead}} and \textbf{\textit{multiple aside}} interaction).}
\label{Fig:5}
\vspace{-0.05cm}

\end{figure}

For experimental analysis, we adopt Simulation of Urban Mobility (SUMO) \cite{35} to manage vehicle interaction scenes with dynamic traffic flows in different complexity. As the intersections shown in Fig. \ref{Fig:5}, in addition to the \textbf{Brain-SAD} controlled \textbf{\textit{ego-vehicle}} in red, \textbf{Scene A} has other two \textit{background} traffic flows including \textbf{1 \textit{ahead straight forward}} and \textbf{1 \textit{ahead turning left}}. The probabilities on vehicles departing per second are all configured as 0.7. \textbf{Scene B} has three background traffic flows including \textbf{2 \textit{ahead straight forward}} and \textbf{1 \textit{ahead turning left}} with departing probabilities all as 0.8. In addition, in \textbf{Scene B}, we add \textbf{1 \textit{sudden aside turning}} (in green), appearing at the 39-\textit{th} second with initial speed 9 \textit{m}/s. \textbf{Scene C} has 5 background traffic flows with the highest complexity including \textbf{3 \textit{ahead straight forward}} (departing probability=0.5 in purple and white), \textbf{2 \textit{ahead turning}} (departing probability=0.7 in cyan, 0.3 in blue) and \textbf{2 \textit{sudden aside turning}} in green, appearing at the 43-\textit{rd} \& 45-\textit{th} second with initial speed 10 \textit{m}/s \& 8 \textit{m/s} turning \textit{right} \& \textit{left} respectively.

Taking the above SUMO simulation as environment, our proposed Brain-SAD is implemented in Python 3.7.1, PyTorch 1.9.1, where by adopting Traffic Control Interface (TraCI), we convey the two-dimensional action (i.e., steering angle \& vertical acceleration) to the \textbf{\textit{red}} ego-vehicle in SUMO playing with other neighbor-vehicles, in turn, the SUMO environmental feedback will be back-propagated still via TraCI for training.

When training, each Scene A/B/C in Fig. \ref{Fig:5} will be run by \textbf{\textit{one}} episode (about 60 second) consisting of \textbf{\textit{at most}} 600 steps with early stopping by \textit{success} or \textit{collision}, and in all, we use 150 episodes as \textbf{\textit{one round}} to train Brain-SAD. During each round, all the explored trajectories ($\{s_t,a_t,s_{t+1},r_t,Fear_{s_t,a_t},policy\_flag_t=long/short,y_t=0/1\}$) will be stored into buffer with adequate size as $|10^6|$. Specifically, in \textbf{\textit{one}} episode, each step (600 at most) will bring about \textbf{\textit{once online}} training via eq (9-11) \& eq (15), then at intervals, 5 steps will result in \textbf{\textit{once offline}} training via eq (20) based on the trajectories in buffer. When one round of 150 episodes completed, evaluation metrics of the current round will be computed, then averaged by 3 rounds for comparison. The above online-offline training has been conducted on NVIDIA GeForce RTX 2080 Ti, 32 GB memory (local workstation) and Intel Xeon Gold 6459C, 64 GB memory (online server), with source codes on detailed configuration at \url{https://github.com/cy-research-lab/Brain-SAD}.

Finally, we compute \textbf{5} \textbf{\textit{offline metrics}} based on trajectories in buffer, including average Task Complete Time or Travel Time (\textbf{TCT}) \cite{36} of one-episode running on the given scene, Success Rate (\textbf{SR}) \cite{36} as non-collision ending at episode-level, average cumulative \textbf{Reward} \cite{36} across episode, Speed Limit Compliance (\textbf{SLC}) \cite{37} as the non-overspeed rate in one-episode, \textbf{Comfortable} \cite{37} as the average speed \textbf{\textit{stability}} rate between adjacent steps across episode. We also compute \textbf{2} \textbf{\textit{online metrics}} that after ego-vehicle makes action at each step in one-episode, we compute Time-to-Collision (\textbf{TTC}) \cite{38} by the latest relative distance from ego- to neighbor-vehicles, further derive Recovery Time (\textbf{RECT}) as the average duration of ego-vehicle to react when \textbf{TTC}$<\tau$, and \textbf{Median TTC} as the overall \textit{safe distance} maintenance after ranking across steps in episode\footnote{See equations on \textbf{7} evaluation metrics in Appendix \textbf{C.1}.}. Obviously, except \textbf{TCT} \& \textbf{RECT}, all the other \textbf{5} metrics are the higher, the better.

\subsection{Ablation Study}
\label{sec:ablation_study}

Based on the above experimental setting, we \textbf{\textit{first}} turn to the ablation study of Brain-SAD shown in \textbf{Table 1} for \textbf{Scene A} and \textbf{Table 2} for \textbf{Scene B}, where 10 variants or combos have been derived from Brain-SAD to ascertain the effectiveness of policy selection, long-term policy and short-term policy.

Similar tendency has been detected in both Table 1 and Table 2 that, with the increase on interaction complexity from Scene A to Scene B, the \textbf{FULL-}version of Brain-SAD (Combo 1, see Fig. 2) has outperformed the \textbf{single long}-term (Combo 5) and \textbf{single short}-term policy (Combo 8) across \textbf{\textit{all}} the metrics. Based on such the observation, it can be further detected that, the \textbf{SR}, \textbf{Reward} and \textbf{SLC} deteriorate from FULL-version, followed by single long-term, ending at single short-term in order (see \textbf{Table 1 \& 2}). However, fluctuations on \textbf{TCT} and \textbf{Comfortable} have occurred across Table 1 \& 2, where the long-term policy with higher \textbf{TCT} (in Table 1) needs more time to complete episode than short-term policy, and in Table 2, long-term policy has lower \textbf{Comfortable} speed \textbf{\textit{stability}} than that of short-term policy, proving that merely long-term policy is inadequate to handle vehicle interaction with higher complexity (e.g., Scene B). Consequently, the collaboration between long-term and short-term policy (FULL-version in Table 1 \& 2) by policy selection in eq (4), has ensured Brain-SAD to complete AD task most successfully with best efficiency and speed steadily when facing different interaction complexity.

\begin{table*}[tbp]
\setlength{\abovecaptionskip}{-0cm}
\setlength{\belowcaptionskip}{4pt}
\caption{The Ablation Study of Brain-SAD on \textbf{Scene A} (\textbf{NOTE:} \textbf{\textit{all}} the performances are averaged by three rounds).}
\label{tab:ablation_scene_a}
\centering
\renewcommand{\arraystretch}{1.1}
\resizebox{\textwidth}{!}{%
\begin{tabular}{clccccc}
\hline
\textbf{Combo} & \textbf{Module Combination} & \textbf{TCT $\downarrow$(s)} & \textbf{SR$\uparrow$(\%)} & \textbf{Reward$\uparrow$} & \textbf{SLC $\uparrow$(\%)} & \textbf{Comfortable $\uparrow$(\%)} \\
\hline
\cellcolor{bsBlue}\textbf{1} & \cellcolor{bsBlue}\textbf{Brain-SAD-FULL} & \cellcolor{bsPink}\textbf{28.50 $\pm$ 0.10} & \cellcolor{bsGreen}\textbf{98.67 $\pm$ 0.67} & \cellcolor{bsGreen}\textbf{105.77 $\pm$ 2.57} & \cellcolor{bsGreen}\textbf{92.05 $\pm$ 0.04} & \cellcolor{bsGreen}\textbf{92.70 $\pm$ 0.28} \\
2 & \textbf{\textit{only}} State Deviation in Scene Imagination & 29.20 $\pm$ 0.80 & 96.22 $\pm$ 0.38 & 95.11 $\pm$ 3.01 & 87.29 $\pm$ 0.67 & 88.90 $\pm$ 0.96 \\
3 & \textbf{\textit{only}} Collision Rate in Scene Imagination & 28.40 $\pm$ 0.60 & 98.44 $\pm$ 0.77 & 94.43 $\pm$ 3.63 & 89.10 $\pm$ 0.28 & 90.13 $\pm$ 0.03 \\
4 & \textbf{\textit{w/o}} Scene Imagination & 29.40 $\pm$ 1.50 & 95.33 $\pm$ 1.15 & 88.43 $\pm$ 3.57 & 83.27 $\pm$ 2.18 & 86.70 $\pm$ 0.55 \\
\cellcolor{bsBlue}\textbf{5} & \cellcolor{bsBlue}\textbf{Brain-SAD \textit{only} Long Term Policy} & \cellcolor{bsPink}\textbf{30.00 $\pm$ 2.20} & \cellcolor{bsGreen}\textbf{97.78 $\pm$ 0.38} & \cellcolor{bsGreen}\textbf{99.32 $\pm$ 0.63} & \cellcolor{bsGreen}\textbf{84.55 $\pm$ 2.43} & \cellcolor{bsGreen}\textbf{89.63 $\pm$ 0.08} \\
6 & \textbf{\textit{w/o}} Dynamic Fear Cost & 31.60 $\pm$ 1.30 & 96.67 $\pm$ 1.33 & 88.39 $\pm$ 0.17 & 81.35 $\pm$ 2.09 & 85.40 $\pm$ 0.34 \\
7 & \textbf{\textit{w/o}} Dynamic Fear Cost \& KL Divergence & 32.30 $\pm$ 0.10 & 94.00 $\pm$ 0.67 & 85.63 $\pm$ 0.04 & 79.80 $\pm$ 0.14 & 83.18 $\pm$ 0.14 \\
\cellcolor{bsBlue}\textbf{8} & \cellcolor{bsBlue}\textbf{Brain-SAD \textit{only} Short Term Policy} & \cellcolor{bsPink}\textbf{29.50 $\pm$ 0.20} & \cellcolor{bsGreen}\textbf{95.33 $\pm$ 0.00} & \cellcolor{bsGreen}\textbf{85.43 $\pm$ 0.64} & \cellcolor{bsGreen}\textbf{81.83 $\pm$ 1.78} & \cellcolor{bsGreen}\textbf{82.10 $\pm$ 0.40} \\
9 & \textbf{\textit{w/o}} $b_{t}^{req}$ in Dynamic Fear Boundary & 29.60 $\pm$ 0.50 & 93.11 $\pm$ 1.02 & 80.70 $\pm$ 0.96 & 76.56 $\pm$ 0.08 & 77.10 $\pm$ 0.20 \\
10 & \textbf{\textit{w/o}} $(\tau_{safe},b_t^{req})$ / \textbf{\textit{no}} Dynamic Fear Boundary & 30.60 $\pm$ 0.60 & 86.67 $\pm$ 0.00 & 69.88 $\pm$ 0.68 & 67.53 $\pm$ 4.99 & 75.30 $\pm$ 0.25 \\
\hline
\end{tabular}
}

\end{table*}

\begin{table*}[tbp]
\setlength{\abovecaptionskip}{-0cm}
\setlength{\belowcaptionskip}{4pt}
\caption{The Ablation Study of Brain-SAD on \textbf{Scene B} (\textbf{NOTE:} \textbf{\textit{all}} the performances are averaged by three rounds).}
\label{tab:ablation_scene_b}
\centering
\renewcommand{\arraystretch}{1.1}
\resizebox{\textwidth}{!}{%
\begin{tabular}{clccccc}
\hline
\textbf{Combo} & \textbf{Module Combination} & \textbf{TCT $\downarrow$(s)} & \textbf{SR$\uparrow$(\%)} & \textbf{Reward$\uparrow$} & \textbf{SLC $\uparrow$(\%)} & \textbf{Comfortable $\uparrow$(\%)} \\
\hline
\cellcolor{bsBlue}\textbf{1} & \cellcolor{bsBlue}\textbf{Brain-SAD-FULL} & \cellcolor{bsPink}\textbf{22.30 $\pm$ 0.10} & \cellcolor{bsGreen}\textbf{97.56 $\pm$ 0.38} & \cellcolor{bsGreen}\textbf{94.35 $\pm$ 0.05} & \cellcolor{bsGreen}\textbf{88.91 $\pm$ 0.17} & \cellcolor{bsPink}\textbf{89.97 $\pm$ 1.82} \\
2 & \textbf{\textit{only}} State Deviation in Scene Imagination & 27.10 $\pm$ 0.10 & 97.33 $\pm$ 0.67 & 89.91 $\pm$ 0.24 & 86.40 $\pm$ 0.49 & 87.00 $\pm$ 0.40 \\
3 & \textbf{\textit{only}} Collision Rate in Scene Imagination & 28.20 $\pm$ 0.00 & 96.67 $\pm$ 1.15 & 86.33 $\pm$ 0.21 & 82.80 $\pm$ 0.31 & 86.30 $\pm$ 1.70 \\
4 & \textbf{\textit{w/o}} Scene Imagination & 28.60 $\pm$ 0.60 & 96.00 $\pm$ 0.67 & 80.56 $\pm$ 0.78 & 80.60 $\pm$ 1.09 & 83.00 $\pm$ 0.40 \\
\cellcolor{bsBlue}\textbf{5} & \cellcolor{bsBlue}\textbf{Brain-SAD \textit{only} Long Term Policy} & \cellcolor{bsPink}\textbf{29.80 $\pm$ 0.10} & \cellcolor{bsGreen}\textbf{97.33 $\pm$ 1.15} & \cellcolor{bsGreen}\textbf{87.78 $\pm$ 0.28} & \cellcolor{bsGreen}\textbf{86.50 $\pm$ 1.80} & \cellcolor{bsPink}\textbf{85.60 $\pm$ 0.50} \\
6 & \textbf{\textit{w/o}} Dynamic Fear Cost & 29.80 $\pm$ 0.60 & 93.56 $\pm$ 0.77 & 82.20 $\pm$ 1.70 & 76.20 $\pm$ 1.83 & 80.80 $\pm$ 0.69 \\
7 & \textbf{\textit{w/o}} Dynamic Fear Cost \& KL Divergence & 30.60 $\pm$ 0.10 & 94.89 $\pm$ 1.02 & 78.57 $\pm$ 1.60 & 72.20 $\pm$ 0.93 & 77.20 $\pm$ 0.22 \\
\cellcolor{bsBlue}\textbf{8} & \cellcolor{bsBlue}\textbf{Brain-SAD \textit{only} Short Term Policy} & \cellcolor{bsPink}\textbf{23.30 $\pm$ 0.30} & \cellcolor{bsGreen}\textbf{93.33 $\pm$ 0.00} & \cellcolor{bsGreen}\textbf{82.55 $\pm$ 0.71} & \cellcolor{bsGreen}\textbf{78.70 $\pm$ 0.79} & \cellcolor{bsPink}\textbf{86.50 $\pm$ 5.20} \\
9 & \textbf{\textit{w/o}} $b_{t}^{req}$ in Dynamic Fear Boundary & 24.20 $\pm$ 0.20 & 92.00 $\pm$ 0.67 & 75.91 $\pm$ 0.20 & 66.70 $\pm$ 1.31 & 72.60 $\pm$ 0.58 \\
10 & \textbf{\textit{w/o}} $(\tau_{safe},b_t^{req})$ / \textbf{\textit{no}} Dynamic Fear Boundary & 26.40 $\pm$ 0.20 & 82.00 $\pm$ 1.15 & 63.32 $\pm$ 0.56 & 62.00 $\pm$ 1.38 & 70.60 $\pm$ 0.28 \\
\hline
\end{tabular}
}

\end{table*}

Next, diving into each branch in Table 1 \& Table 2 that, for \textbf{Combo 1-4} on policy selection, we retain dual-path policy while gradually deducting State Deviation (Combo 2), Collision Rate (Combo 3) from scene imagination (see $(\bar p_{t+1}^{col,t+L},\bar\sigma_{t+1}^{U,t+L})$ in eq (2-3) used to perceive fear-signal $Fear_{s_t,a_t^{\theta_{t-1}}}$), finally discarding whole scene imagination (Combo 4). In this way, the policy selection via 5-HT \& NE (Calmness \& Nervousness) competition in eq (4) will degrade as random selection, or the \textbf{\textit{worst}} in Combo 4. Following the above deduction, we focus on the \textit{dynamic constraint} of long-term policy in \textbf{Combo 5-7}, keeping long-term reward $Q_{\text{reward}}^{s_t,A_t^{\theta_{t-1}}}$, gradually deleting dynamic fear cost ($\Psi_{s_t,A_t^{\theta_{t-1}}}^{\text{Fear}}$) and KL-divergence trust region $D_{KL}[\cdot]\leq\varepsilon_{KL}$ in eq (8-11), leading to the \textbf{\textit{worst}} Combo 7 where long-term policy will \textbf{\textit{only}} be online optimized by long-term reward without dynamic fear-oriented constraint. Similarly, we degrade the short-term policy by gradually removing dynamic fear boundary $\Psi_{s_t,a_t}^{\text{Fear}}=\{\tau_{safe},b_t^{req}\}$ in eq (15), with Combo 10 as the \textbf{\textit{worst}}. Consequently, we take the \textbf{\textit{FULL-version}} of Brain-SAD (see Fig. 1) as the most optimal one to be compared with the existing works.

\subsection{Comparative Study}
\label{sec:comparative_study}

Based on ablation study, we select \textbf{\textit{FULL}}-version of Brain-SAD (see Fig. 2) in Table 1 \& 2 to be compared with the current constrained and brain-inspired action control works directly or indirectly concerning Safe AD listed as follows (see \textbf{Section II} and \textbf{Appendix C.2.} for detailed explanations on the compared methods, with our reproduced codes here\footnote{See codes on the compared methods at \url{https://github.com/cy-research-lab/Compared-methods}.}). Moreover, \textbf{Scene B} and \textbf{Scene C} with increasing complexity (see Fig. 5) will be adopted for simulations.

For ego-vehicle Policy-Oriented Constrained methods, we select \textbf{CADRE} \cite{16}, \textbf{Iso-Dream++} \cite{18}, \textbf{RLfOLD} \cite{17} and \textbf{MARL-CCE} \cite{19}. All the above will act as ego-vehicle policy to be trained in SUMO, so as to feedback trajectories, rewards and others via TraCI. For Environment-Oriented Prior Constrained methods, we select \textbf{LS-Imagination} \cite{20}, \textbf{CarPlanner} \cite{21}, \textbf{Hybrid-Driving} (LLM-based) \cite{23}, \textbf{SafeAuto} (MLLM-based) \cite{24} and \textbf{SimLingo} (VLM-based) \cite{22}. Here, in terms of \textbf{LS-Imagination \& CarPlanner}, the state-transition imagination will be pre-trained by SUMO simulation then frozen as the foundation to train Actor/Critic of ego-vehicle. When it comes to the other three LM-based models (Hybrid-Driving, SafeAuto, SimLingo), we use SUMO simulation to construct environment-related resources and take multi-modal scene features along with agent states in SUMO as input. For Brain-Inspired Action Control methods, we select \textbf{SVPG} \cite{25}, \textbf{FoG-RL} \cite{27}, \textbf{EFT-RL} \cite{28} and \textbf{Goal-Reducer} \cite{29}. All the above are trained in SUMO to as the ego-policy.

%
\begin{table*}[!t]
\setlength{\abovecaptionskip}{-0cm}
\setlength{\belowcaptionskip}{4pt}
\caption{The Comparative Study of Brain-SAD on \textbf{Scene B} (\textbf{NOTE:} the \textbf{\textit{bold}} font represents the most optimal performance in each column, with the colored as the sub-optimal performance in each category. \textbf{ALL} the performance are averaged by \textbf{three} rounds).}
\label{tab:comparison_scene_b}
\centering
\renewcommand{\arraystretch}{1.1}
\resizebox{\textwidth}{!}{%
\begin{tabular}{cclccccccc}
\hline
\textbf{NO} & \textbf{Category} & \textbf{Methods} & \textbf{TCT $\downarrow$(s)} & \textbf{RECT $\downarrow$(s)} & \textbf{SR$\uparrow$(\%)} & \textbf{Reward$\uparrow$} & \textbf{SLC $\uparrow$(\%)} & \textbf{Comfortable $\uparrow$(\%)} & \textbf{Median-TTC$\uparrow$(s)} \\
\hline
1 & \multirow{4}{*}{\shortstack{Policy-Oriented\\Constrained}} & CADRE & 36.30 $\pm$ 0.60 & 0.7181 $\pm$ 0.0032 & 93.56 $\pm$ 0.83 & 85.45 $\pm$ 0.44 & 82.35 $\pm$ 0.51 & 83.34 $\pm$ 0.72 & 0.7162 $\pm$ 0.0045 \\
2 &  & Iso-Dream++ & \cellcolor{bsGreen}30.20 $\pm$ 0.20 & 0.6245 $\pm$ 0.0057 & \cellcolor{bsGreen}97.29 $\pm$ 0.13 & \cellcolor{bsGreen}86.23 $\pm$ 0.82 & \cellcolor{bsGreen}86.12 $\pm$ 0.10 & \cellcolor{bsGreen}87.52 $\pm$ 0.23 & \cellcolor{bsGreen}1.1387 $\pm$ 0.0102 \\
3 &  & RLfOLD & 33.20 $\pm$ 0.80 & \cellcolor{bsGreen}0.5661 $\pm$ 0.0310 & 92.00 $\pm$ 0.94 & 82.88 $\pm$ 1.80 & 83.52 $\pm$ 0.37 & 84.75 $\pm$ 2.11 & 1.0266 $\pm$ 0.0053 \\
4 &  & MARL-CCE & 31.60 $\pm$ 0.40 & 0.8417 $\pm$ 0.0335 & 92.67 $\pm$ 1.10 & 81.05 $\pm$ 0.02 & 81.72 $\pm$ 2.58 & 81.16 $\pm$ 0.04 & 0.7340 $\pm$ 0.0057 \\
\hline
5 & \multirow{5}{*}{\shortstack{Environment-Oriented\\Prior Constrained}} & LS-Imagination & 31.20 $\pm$ 0.60 & 0.6547 $\pm$ 0.0125 & 96.03 $\pm$ 0.00 & 84.57 $\pm$ 0.02 & 85.47 $\pm$ 1.11 & 85.77 $\pm$ 0.06 & 1.0934 $\pm$ 0.0013 \\
6 &  & CarPlanner & 33.10 $\pm$ 0.10 & 0.8333 $\pm$ 0.0471 & 95.57 $\pm$ 0.57 & 84.37 $\pm$ 0.14 & 85.30 $\pm$ 1.79 & 82.20 $\pm$ 0.01 & 0.8956 $\pm$ 0.0060 \\
7 &  & SimLingo & 26.00 $\pm$ 0.10 & 0.5127 $\pm$ 0.0034 & 95.78 $\pm$ 1.02 & 93.79 $\pm$ 0.29 & \cellcolor{bsPink}\textbf{92.12 $\pm$ 2.17} & \cellcolor{bsPink}\textbf{92.56 $\pm$ 0.19} & 1.6939 $\pm$ 0.0090 \\
8 &  & Hybrid-Driving & 25.30 $\pm$ 0.10 & 0.5273 $\pm$ 0.0012 & \cellcolor{bsPink}97.11 $\pm$ 0.38 & \cellcolor{bsPink}94.26 $\pm$ 0.33 & 92.01 $\pm$ 0.22 & 91.00 $\pm$ 0.00 & \cellcolor{bsPink}\textbf{1.7107 $\pm$ 0.0021} \\
9 &  & SafeAuto & \cellcolor{bsPink}22.50 $\pm$ 0.20 & \cellcolor{bsPink}0.4601 $\pm$ 0.0054 & 96.67 $\pm$ 0.00 & 91.83 $\pm$ 1.03 & 91.02 $\pm$ 0.00 & 91.02 $\pm$ 0.00 & 1.6470 $\pm$ 0.0068 \\
\hline
10 & \multirow{6}{*}{Brain-Inspired} & SVPG & 24.60 $\pm$ 0.10 & 0.6106 $\pm$ 0.0201 & \cellcolor{bsBlue}96.89 $\pm$ 0.38 & 85.59 $\pm$ 0.27 & 85.98 $\pm$ 0.10 & 87.04 $\pm$ 0.00 & \cellcolor{bsBlue}1.2653 $\pm$ 0.0006 \\
11 &  & FNI-RL & 24.90 $\pm$ 0.40 & \cellcolor{bsBlue}0.5466 $\pm$ 0.0252 & 96.67 $\pm$ 1.15 & \cellcolor{bsBlue}89.15 $\pm$ 0.44 & 86.78 $\pm$ 0.64 & 87.01 $\pm$ 1.59 & 0.8469 $\pm$ 0.0107 \\
12 &  & FoG-RL & \cellcolor{bsBlue}24.10 $\pm$ 0.70 & 0.6429 $\pm$ 0.0027 & 96.44 $\pm$ 0.38 & 85.08 $\pm$ 0.35 & 85.44 $\pm$ 0.68 & \cellcolor{bsBlue}89.60 $\pm$ 0.62 & 1.1533 $\pm$ 0.0118 \\
13 &  & EFT-RL & 29.00 $\pm$ 1.00 & 0.7276 $\pm$ 0.0527 & 90.86 $\pm$ 0.21 & 87.62 $\pm$ 1.21 & \cellcolor{bsBlue}89.56 $\pm$ 1.45 & 84.05 $\pm$ 0.14 & 1.0935 $\pm$ 0.0193 \\
14 &  & Goal-Reducer & 26.40 $\pm$ 0.20 & 0.6520 $\pm$ 0.0025 & 96.00 $\pm$ 0.67 & 86.38 $\pm$ 0.93 & 83.72 $\pm$ 0.03 & 84.39 $\pm$ 0.11 & 0.9887 $\pm$ 0.0020 \\
15 &  & \cellcolor{bsYellow}\textbf{Brain-SAD (Ours)} & \cellcolor{bsYellow}\textbf{22.30 $\pm$ 0.10} & \cellcolor{bsYellow}\textbf{0.4586 $\pm$ 0.0039} & \cellcolor{bsYellow}\textbf{97.56 $\pm$ 0.38} & \cellcolor{bsYellow}\textbf{94.35 $\pm$ 0.05} & \cellcolor{bsYellow}88.91 $\pm$ 0.17 & \cellcolor{bsYellow}89.97 $\pm$ 1.82 & \cellcolor{bsYellow}1.0014 $\pm$ 0.0012 \\
\hline
\end{tabular}
}

\par\vspace{5pt}
\setlength{\abovecaptionskip}{-0cm}
\setlength{\belowcaptionskip}{4pt}
\caption{The Comparative Study of Brain-SAD on \textbf{Scene C} (\textbf{NOTE:} the \textbf{\textit{bold}} font represents the most optimal performance in each column, with the colored as the sub-optimal performance in each category. \textbf{ALL} the performance are averaged by \textbf{three} rounds).}
\label{tab:comparison_scene_c}
\centering
\renewcommand{\arraystretch}{1.1}
\resizebox{\textwidth}{!}{%
\begin{tabular}{cclccccccc}
\hline
\textbf{NO} & \textbf{Category} & \textbf{Methods} & \textbf{TCT $\downarrow$(s)} & \textbf{RECT $\downarrow$(s)} & \textbf{SR$\uparrow$(\%)} & \textbf{Reward$\uparrow$} & \textbf{SLC $\uparrow$(\%)} & \textbf{Comfortable $\uparrow$(\%)} & \textbf{Median-TTC$\uparrow$(s)} \\
\hline
1 & \multirow{4}{*}{\shortstack{Policy-Oriented\\Constrained}} & CADRE & \cellcolor{bsGreen}28.80 $\pm$ 0.30 & 0.7261 $\pm$ 0.0223 & 89.56 $\pm$ 2.52 & 90.45 $\pm$ 0.90 & 76.20 $\pm$ 0.12 & 79.91 $\pm$ 0.54 & 1.0712 $\pm$ 0.0050 \\
2 &  & Iso-Dream++ & 47.30 $\pm$ 0.90 & \cellcolor{bsGreen}0.6475 $\pm$ 0.0380 & \cellcolor{bsGreen}94.44 $\pm$ 0.77 & \cellcolor{bsGreen}93.98 $\pm$ 0.44 & \cellcolor{bsGreen}84.27 $\pm$ 1.34 & \cellcolor{bsGreen}85.95 $\pm$ 0.02 & \cellcolor{bsGreen}\textbf{1.5657 $\pm$ 0.0159} \\
3 &  & RLfOLD & 41.70 $\pm$ 0.10 & 0.7477 $\pm$ 0.0252 & 89.56 $\pm$ 1.02 & 88.15 $\pm$ 0.34 & 78.08 $\pm$ 0.14 & 78.84 $\pm$ 0.14 & 1.2017 $\pm$ 0.0032 \\
4 &  & MARL-CCE & 42.00 $\pm$ 0.50 & 0.6672 $\pm$ 0.0065 & 90.00 $\pm$ 0.00 & 91.22 $\pm$ 0.09 & 79.96 $\pm$ 0.15 & 81.47 $\pm$ 0.03 & 1.4341 $\pm$ 0.0022 \\
\hline
5 & \multirow{5}{*}{\shortstack{Environment-Oriented\\Prior Constrained}} & LS-Imagination & \cellcolor{bsPink}31.20 $\pm$ 1.40 & 0.6955 $\pm$ 0.0243 & 91.33 $\pm$ 0.67 & 91.98 $\pm$ 0.45 & 80.89 $\pm$ 1.72 & 83.64 $\pm$ 1.57 & 1.1695 $\pm$ 0.0362 \\
6 &  & CarPlanner & 37.10 $\pm$ 0.50 & 0.6464 $\pm$ 0.0199 & 95.35 $\pm$ 1.15 & 93.26 $\pm$ 0.68 & 82.47 $\pm$ 0.67 & 81.00 $\pm$ 0.13 & \cellcolor{bsPink}1.4150 $\pm$ 0.0053 \\
7 &  & SimLingo & 32.50 $\pm$ 0.20 & 0.5950 $\pm$ 0.0342 & 96.33 $\pm$ 0.47 & 89.31 $\pm$ 0.24 & \cellcolor{bsPink}\textbf{86.65 $\pm$ 0.12} & 87.69 $\pm$ 1.13 & 1.2021 $\pm$ 0.0014 \\
8 &  & Hybrid-Driving & 39.10 $\pm$ 0.20 & 0.5441 $\pm$ 0.0523 & 95.78 $\pm$ 1.39 & \cellcolor{bsPink}93.55 $\pm$ 0.47 & 84.11 $\pm$ 0.43 & 86.02 $\pm$ 0.08 & 1.4007 $\pm$ 0.0096 \\
9 &  & SafeAuto & 31.60 $\pm$ 0.10 & \cellcolor{bsPink}0.5336 $\pm$ 0.0088 & \cellcolor{bsPink}96.67 $\pm$ 0.67 & 93.07 $\pm$ 0.23 & 86.25 $\pm$ 0.00 & \cellcolor{bsPink}\textbf{88.93 $\pm$ 0.60} & 1.3079 $\pm$ 0.0021 \\
\hline
10 & \multirow{6}{*}{Brain-Inspired} & SVPG & 32.40 $\pm$ 0.20 & 0.6200 $\pm$ 0.0340 & \cellcolor{bsBlue}94.00 $\pm$ 1.33 & \cellcolor{bsBlue}93.60 $\pm$ 0.47 & 82.26 $\pm$ 0.72 & 84.96 $\pm$ 0.07 & 1.2370 $\pm$ 0.0030 \\
11 &  & FNI-RL & \cellcolor{bsBlue}30.90 $\pm$ 1.50 & \cellcolor{bsBlue}0.6040 $\pm$ 0.0090 & 93.48 $\pm$ 1.72 & 92.91 $\pm$ 0.59 & \cellcolor{bsBlue}84.43 $\pm$ 0.86 & \cellcolor{bsBlue}86.19 $\pm$ 2.07 & 1.1360 $\pm$ 0.0300 \\
12 &  & FoG-RL & 32.90 $\pm$ 2.00 & 0.6559 $\pm$ 0.0302 & 93.11 $\pm$ 0.38 & 83.33 $\pm$ 0.11 & 80.42 $\pm$ 0.97 & 83.96 $\pm$ 0.78 & 1.2190 $\pm$ 0.0170 \\
13 &  & EFT-RL & 51.70 $\pm$ 0.70 & 0.9270 $\pm$ 0.0180 & 76.89 $\pm$ 2.52 & 90.00 $\pm$ 0.49 & 75.35 $\pm$ 0.20 & 81.38 $\pm$ 0.04 & \cellcolor{bsBlue}1.4560 $\pm$ 0.0060 \\
14 &  & Goal-Reducer & 32.80 $\pm$ 0.30 & 0.7181 $\pm$ 0.0058 & 92.00 $\pm$ 2.00 & 88.61 $\pm$ 0.73 & 74.93 $\pm$ 0.17 & 82.65 $\pm$ 0.11 & 0.9774 $\pm$ 0.0026 \\
15 &  & \cellcolor{bsYellow}\textbf{Brain-SAD (Ours)} & \cellcolor{bsYellow}\textbf{28.10 $\pm$ 0.00} & \cellcolor{bsYellow}\textbf{0.5263 $\pm$ 0.0194} & \cellcolor{bsYellow}\textbf{97.11 $\pm$ 0.77} & \cellcolor{bsYellow}\textbf{94.15 $\pm$ 0.28} & \cellcolor{bsYellow}86.12 $\pm$ 0.07 & \cellcolor{bsYellow}87.93 $\pm$ 0.70 & \cellcolor{bsYellow}1.1011 $\pm$ 0.0043 \\
\hline
\end{tabular}
}

\end{table*}

In terms of the effectiveness on the AD task completion across \textbf{Scene B} and \textbf{Scene C} (see Table 3 \& Table 4), our proposed Brain-SAD has achieved the most optimal performance with the \textbf{\textit{shortest}} Task Completion Time (\textbf{TCT}), the \textbf{\textit{shortest}} Recovery Time (\textbf{RECT}) handling collision risk, the \textbf{\textit{highest}} Success Rate (\textbf{SR}) reaching the end, along with the \textbf{\textit{highest}} averaged cumulative \textbf{Reward} compared with the others. Such the advantage has been presented more explicitly in \textbf{Scene C} (see Table 4) when dealing with traffic flows with more ahead and aside interactions. Moreover, we have selected TOP-2 methods in each category by \textbf{Reward} from Table 3 \& Table 4 whose training traces on \textbf{TCT}, \textbf{SR}, Collision Rate (\textbf{CR}) and \textbf{Reward} of the \textbf{\textit{best}} training round (with the highest average cumulative reward consisting of 150 episodes) have been demonstrated in Fig. 6 \& Fig. 7 (\textit{a})--(\textit{d}), from which it can be detected that, our proposed Brain-SAD can smoothly converge towards the most optimal level even when facing more complex interaction in \textbf{Scene C}.

More specifically, for \textbf{SR}, \textbf{Reward} in Table 3 \& Table 4, it can be detected that environment-oriented constrained and brain-inspired methods have performed overall better than policy-oriented ones although with fluctuations. Furthermore, when turning to the most complex \textbf{Scene C} in Table 4, those environment-oriented constrained methods achieve better \textbf{SR} and \textbf{Reward} than brain-inspired ones, yet still all below Brain-SAD. Such the tendency can be ascribed to the reason that, compared with the current brain-inspired methods, \textbf{\textit{only}} our Brain-SAD has been modeled by the complete \textit{perception-determination-action} functional mapping in human brain, from fear-oriented policy selection (i.e., AMY$\rightarrow$PFC in eq (3-4)) to selective policy execution (i.e., PFC$\rightarrow$Striatum) between long-term policy for action distribution optimization in regular evolving scene (see eq (9-11)) and short-term policy for direct action-projection on urgent-braking (see eq (13-15)). That is, the dual-path policy collaboration has brought about higher performance than the single policy of the current methods.

Particularly, for FNI-RL, Iso-Dream++, Hybrid-Driving in Table 3 \& Table 4, although they have incorporated counterpart fear-constraint like Brain-SAD, \textbf{\textit{only}} our Brain-SAD has coupled the \textbf{\textit{dynamic fear constraint}} altering across scenarios with policy \textbf{Value Estimation} to \textbf{\textit{online}} guide policy optimization. That is, in Brain-SAD, the overall impact after adopting action \textit{$a_t$} has been estimated as \textbf{\textit{dynamic fear constraint}} consisting of long-term reward \& long-term fear-cost (see $Q_{Reward}^{s_t,A_t^{\theta_{t-1}}}-\Psi_{s_t,A_t^{\theta_{t-1}}}^{Fear}$ in eq (5-6)) , further resolved into the most optimal posterior action weights as the supervisory constraint to directly update the current policy action distribution \textbf{\textit{online}} (see eq (11)). In this way, different actions will lead to different long-term fear-reaction ($Q_{Reward}^{s_t,A_t^{\theta_{t-1}}}-\Psi_{s_t,A_t^{\theta_{t-1}}}^{Fear}$), resulting in different policy update directions. Consequently, in Brain-SAD, the causality between action and its corresponding fear-reaction has been maintained and incorporated into online policy update, which is different from the \textbf{\textit{state-level}} fear-cost scoring in brain-inspired FNI-RL (fear-cost is statically projected from state, weak connection with action impact), uncontrollable dynamics correction as policy input in Iso-Dream++ (no direct guiding on action distribution optimization), hazard action exploration in Hybrid-Driving (still as offline trajectory for replay).

\enlargethispage{\baselineskip}
\vspace{0.40\baselineskip}
It is worthwhile to be noted when observing \textbf{TCT}, \textbf{RECT}, \textbf{SLC}, \textbf{Comfortable} and \textbf{Median-TTC} in Table 3 (Scene B) \& Table 4 (Scene C), another phenomenon has been detected on Brain-SAD. That is, our Brain-SAD can perform \textbf{\textit{faster}} and \textbf{\textit{better}} (i.e., \textbf{\textit{lower}} TCT, RECT with \textbf{\textit{higher}} SR, Reward), yet somewhat aggressive on speeding in confidence while still within acceptable scope (i.e., \textbf{\textit{lower}} speed compliance in \textbf{SLC}, \textbf{\textit{higher}} adjacent speeding to be not \textbf{Comfortable}, resulting in \textbf{\textit{shorter}} median Time-to-Collision (\textbf{Median-TTC}) or \textbf{\textit{shorter}} relative distance towards neighbors). Such the phenomenon can be explained as in terms of the long-term policy for regular scene-evolving in Brain-SAD, once the \textbf{\textit{dynamic fear constraint}} has provided the right policy direction, the $D_{KL}[\cdot]\leq\varepsilon_{KL}$ KL-divergence in Lagrange Objective (see eq (8) \& (11)) will refine policy update within trust region preventing dramatic policy shift, so as to be formed into stable behaviour pattern. That is, in Fig. 6 \& Fig. 7 (\textit{e})--(\textit{h}), we select TOP-1 methods in each category from Table 3 \& 4, scatter their adopted actions across the \textbf{\textit{best}} episode (selected from the \textbf{\textit{best}} training round in Fig. 6 \& Fig. 7 (\textit{a})--(\textit{d})), fully-connect all the (steer, accelerate) actions as distribution, where in Fig. 6 \& Fig. 7 (\textit{e})--(\textit{h}), Brain-SAD has finally converged to \textbf{\textit{positive}} acceleration as fixed pattern, behaving confident speeding in the shortest TCT\clearpage

\begin{figure*}[!t]
\noindent \& RECT with experience on higher SR and Reward.\par
\vspace{0.06cm}
\setlength{\abovecaptionskip}{0.235cm}
\setlength{\belowcaptionskip}{0pt}
\centering

%
\begingroup
\newlength{\FigSixPanelHeight}
\setlength{\FigSixPanelHeight}{0.175\textwidth}
\renewcommand*{\subcapsize}{\scriptsize}
\setlength{\subfigcapskip}{1pt}

\subfigure[Task Completion Time]{%
  \begin{minipage}[b]{0.238\textwidth}
    \centering
    \parbox[c][\FigSixPanelHeight][c]{\linewidth}{%
      \centering
      \includegraphics[
        width=\linewidth,
        height=\FigSixPanelHeight,
        keepaspectratio
      ]{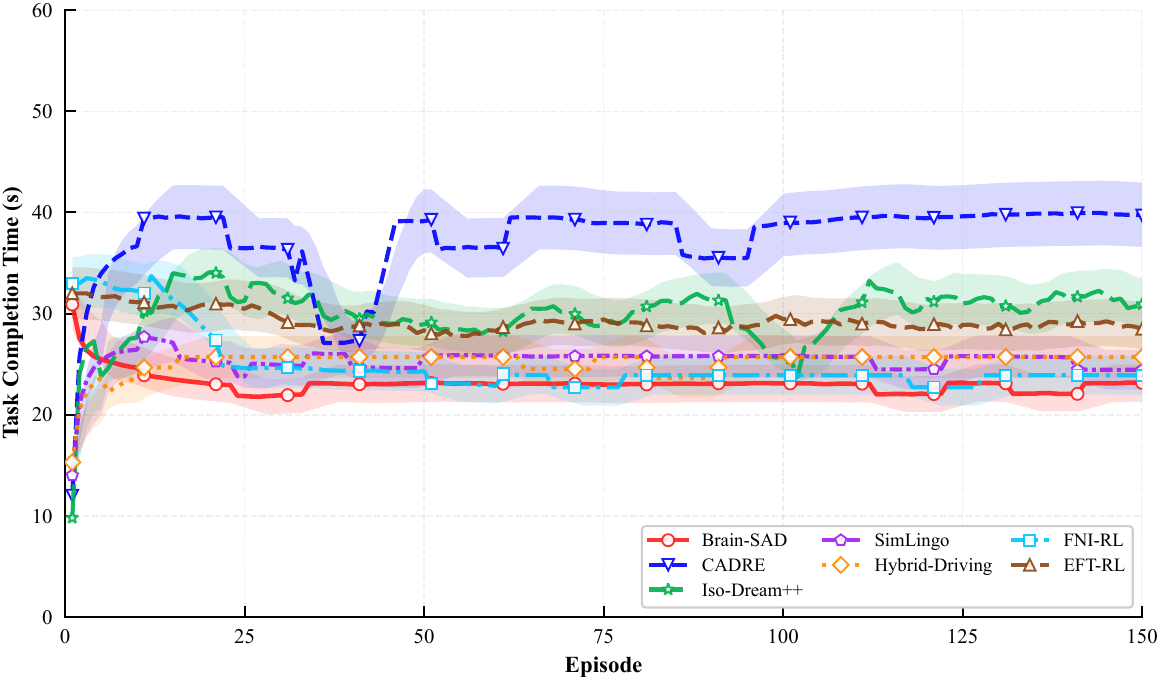}%
    }%
  \end{minipage}%
}%
\hspace{0.006\textwidth}%
\subfigure[Success Rate]{%
  \begin{minipage}[b]{0.238\textwidth}
    \centering
    \parbox[c][\FigSixPanelHeight][c]{\linewidth}{%
      \centering
      \includegraphics[
        trim=0 0 413.7455 22,
        clip,
        width=\linewidth,
        height=\FigSixPanelHeight,
        keepaspectratio
      ]{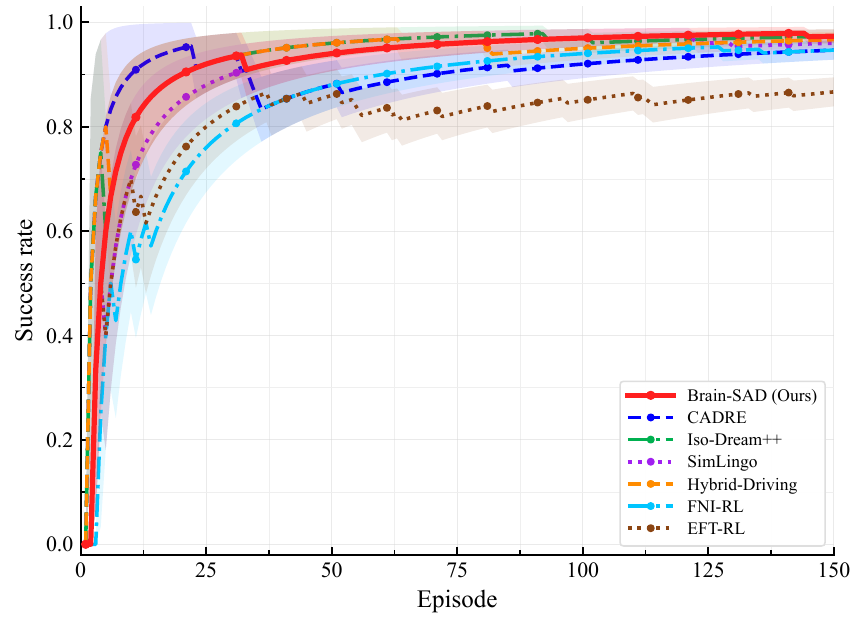}%
    }%
  \end{minipage}%
}%
\hspace{0.006\textwidth}%
\subfigure[Collision Rate]{%
  \begin{minipage}[b]{0.238\textwidth}
    \centering
    \parbox[c][\FigSixPanelHeight][c]{\linewidth}{%
      \centering
      \includegraphics[
        trim=413.7455 0 0 22,
        clip,
        width=\linewidth,
        height=\FigSixPanelHeight,
        keepaspectratio
      ]{Figs/Fig6-2.pdf}%
    }%
  \end{minipage}%
}%
\hspace{0.006\textwidth}%
\subfigure[Cumulative Average Reward]{%
  \begin{minipage}[b]{0.238\textwidth}
    \centering
    \parbox[c][\FigSixPanelHeight][c]{\linewidth}{%
      \centering
      \includegraphics[
        width=\linewidth,
        height=\FigSixPanelHeight,
        keepaspectratio
      ]{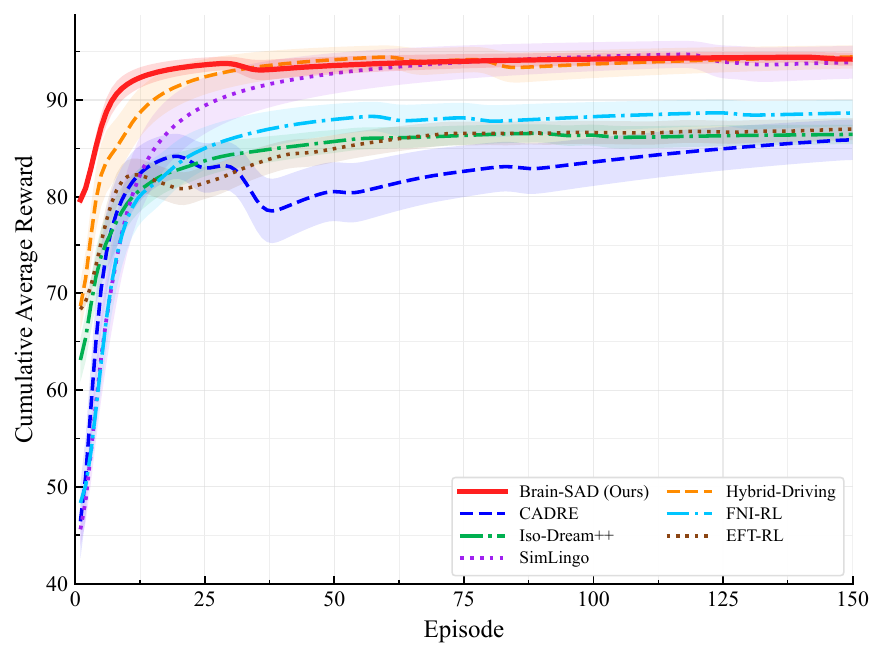}%
    }%
  \end{minipage}%
}%

\par\vspace{-0.5cm}

\subfigure[Iso-Dream++ action distribution]{%
  \begin{minipage}[b]{0.238\textwidth}
    \centering
    \parbox[c][\FigSixPanelHeight][c]{\linewidth}{%
      \centering
      \includegraphics[
        trim=0 30 0 45,
        clip,
        width=\linewidth,
        height=\FigSixPanelHeight,
        keepaspectratio
      ]{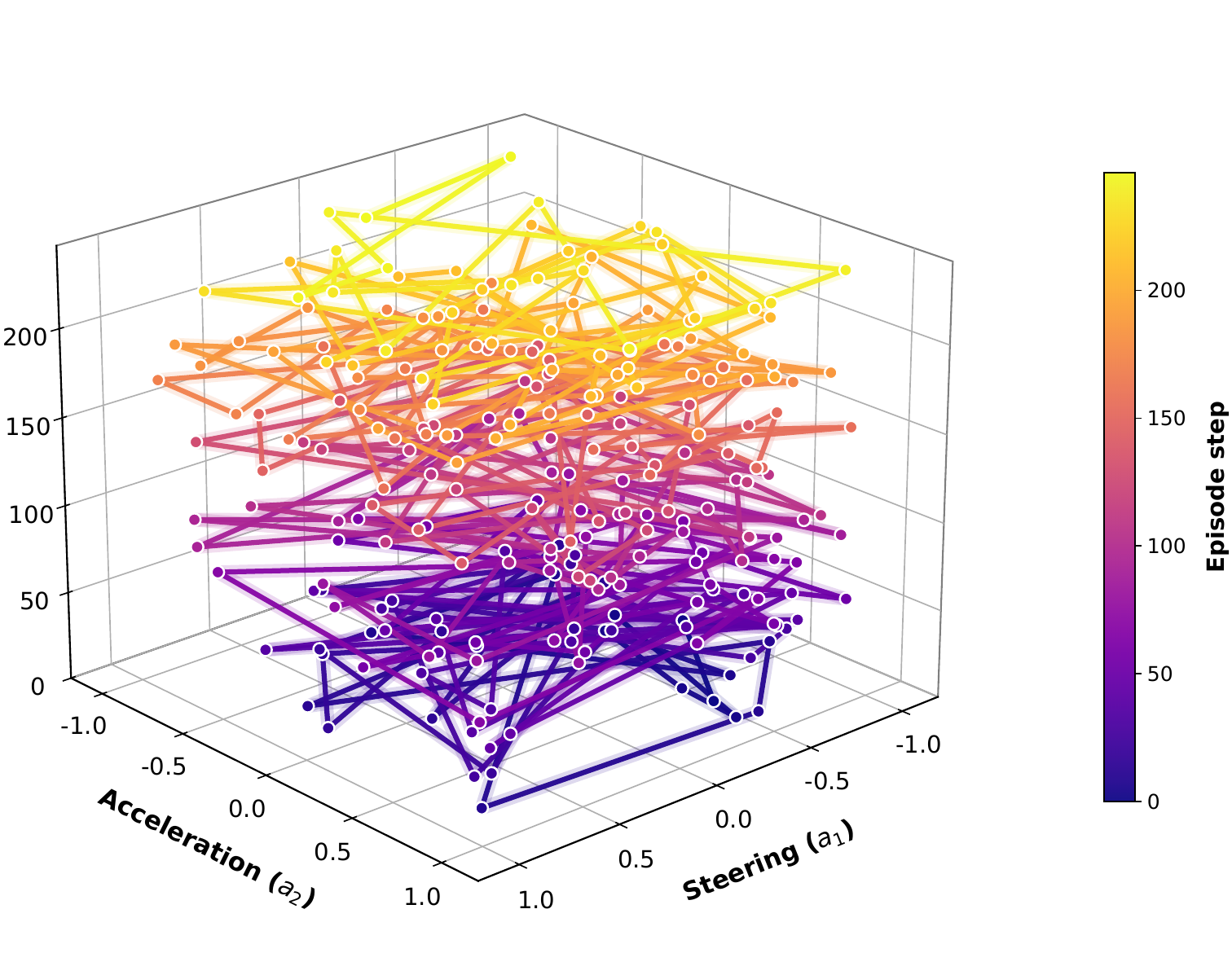}%
    }%
  \end{minipage}%
}%
\hspace{0.006\textwidth}%
\subfigure[Hybrid-Driving action distribution]{%
  \begin{minipage}[b]{0.238\textwidth}
    \centering
    \parbox[c][\FigSixPanelHeight][c]{\linewidth}{%
      \centering
      \includegraphics[
        trim=0 30 0 45,
        clip,
        width=\linewidth,
        height=\FigSixPanelHeight,
        keepaspectratio
      ]{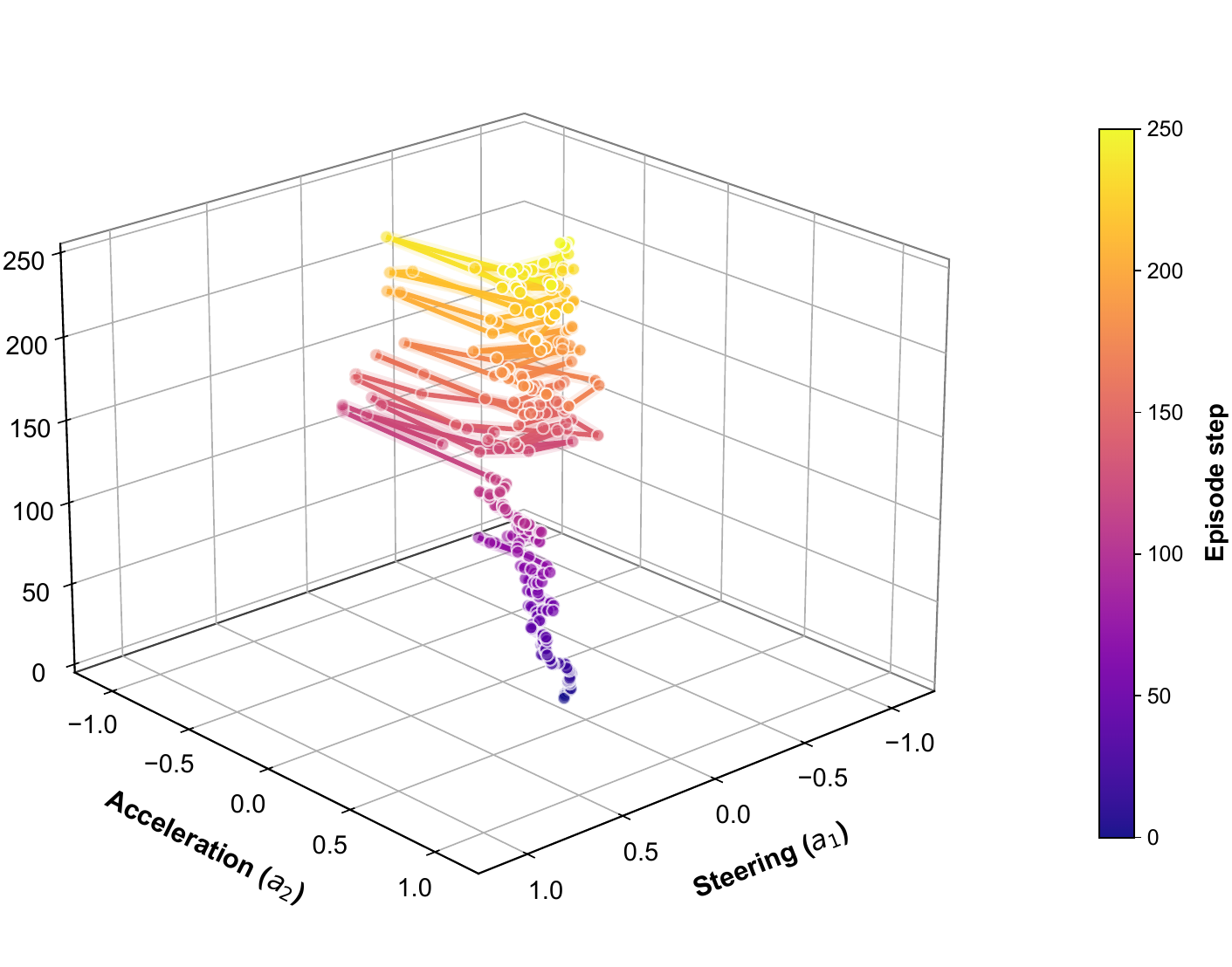}%
    }%
  \end{minipage}%
}%
\hspace{0.006\textwidth}%
\subfigure[FNI-RL action distribution]{%
  \begin{minipage}[b]{0.238\textwidth}
    \centering
    \parbox[c][\FigSixPanelHeight][c]{\linewidth}{%
      \centering
      \includegraphics[
        trim=0 30 0 45,
        clip,
        width=\linewidth,
        height=\FigSixPanelHeight,
        keepaspectratio
      ]{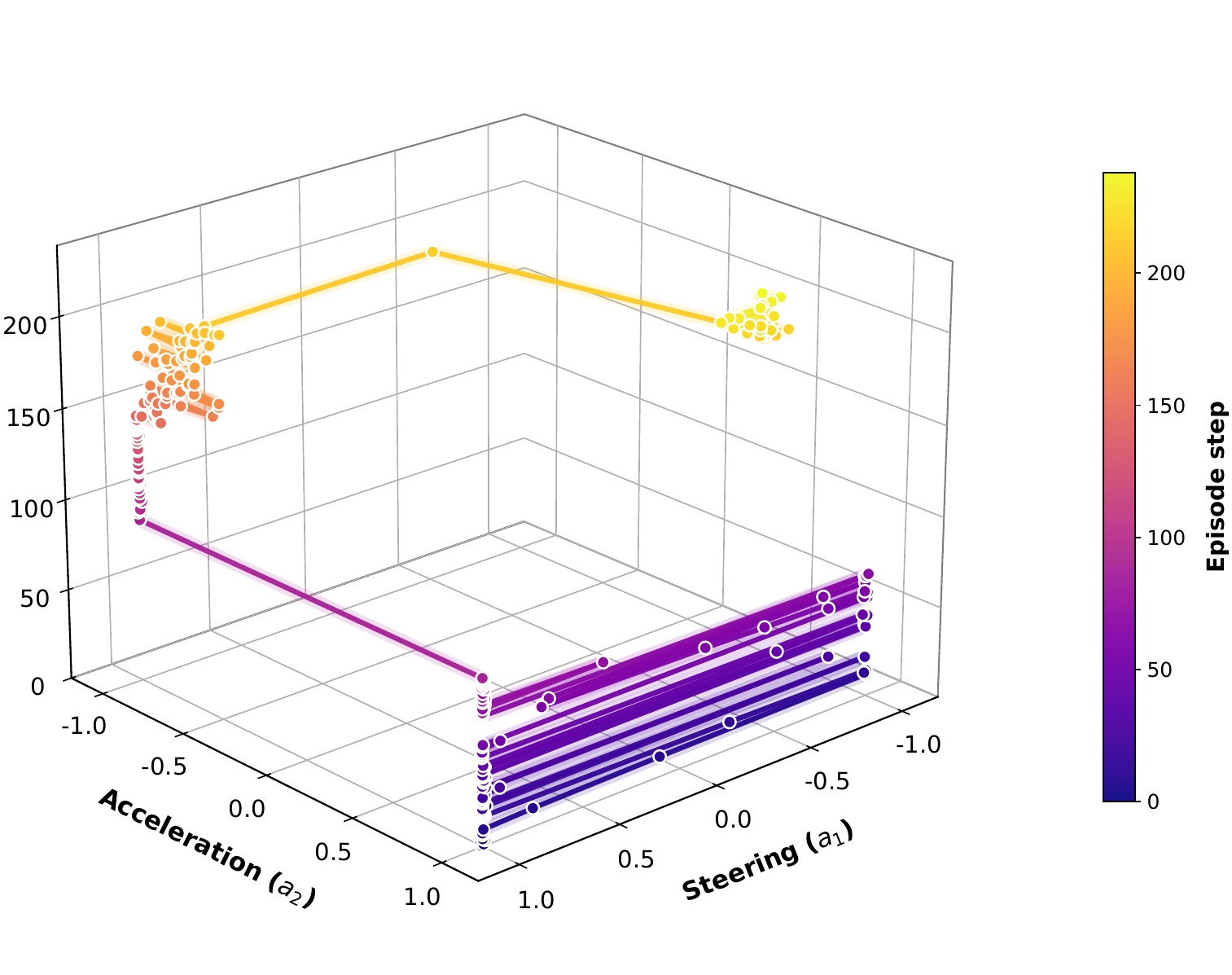}%
    }%
  \end{minipage}%
}%
\hspace{0.006\textwidth}%
\subfigure[Brain-SAD action distribution]{%
  \begin{minipage}[b]{0.238\textwidth}
    \centering
    \parbox[c][\FigSixPanelHeight][c]{\linewidth}{%
      \centering
      \includegraphics[
        trim=0 30 0 45,
        clip,
        width=\linewidth,
        height=\FigSixPanelHeight,
        keepaspectratio
      ]{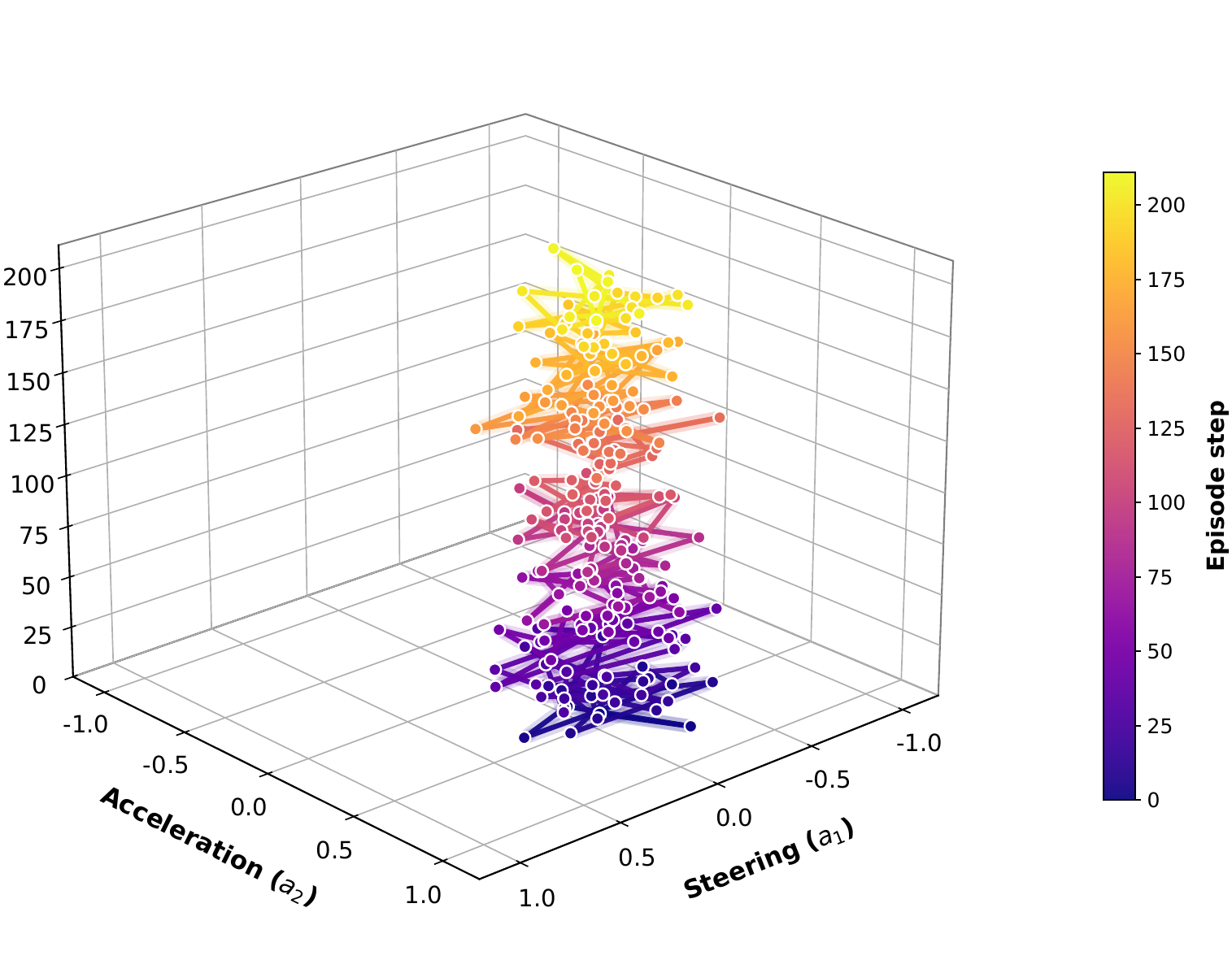}%
    }%
  \end{minipage}%
}%

\endgroup
\vspace{-0.3cm} 
\caption{The training traces and action distributions on representative methods selected from \textbf{Table 3}, simulated in \textbf{Scene B}. (\textbf{NOTE}: In (\textit{a})--(\textit{d}), we present training traces on methods with TOP 2 \textbf{Reward} reported in \textbf{Table 3}; Here, the \textbf{Reward} are the average cumulative reward on the \textbf{\textit{best}} training round (150 episodes, 3 rounds in all); In (\textit{e})--(\textit{h}), we present action distributions across the \textbf{\textit{best episode}} (selected in the above best training round) on methods with TOP 1 \textbf{Reward} reported in \textbf{Table 3}, X-axis for acceleration positive as speeding, Y-axis for steering angle, Z-axis for temporal steps in the best episode.)}
\label{Fig:6}

\setcounter{subfigure}{0}
\setlength{\abovecaptionskip}{0.235cm}
\setlength{\belowcaptionskip}{0pt}
\centering

%
\begingroup

\newlength{\FigSevenPanelHeight}
\setlength{\FigSevenPanelHeight}{0.175\textwidth}
\renewcommand*{\subcapsize}{\scriptsize}
\setlength{\subfigcapskip}{2pt}

\subfigure[Task Completion Time]{%
  \begin{minipage}[b]{0.238\textwidth}
    \centering
    \parbox[c][\FigSevenPanelHeight][c]{\linewidth}{%
      \centering
      \includegraphics[
        width=\linewidth,
        height=\FigSevenPanelHeight,
        keepaspectratio
      ]{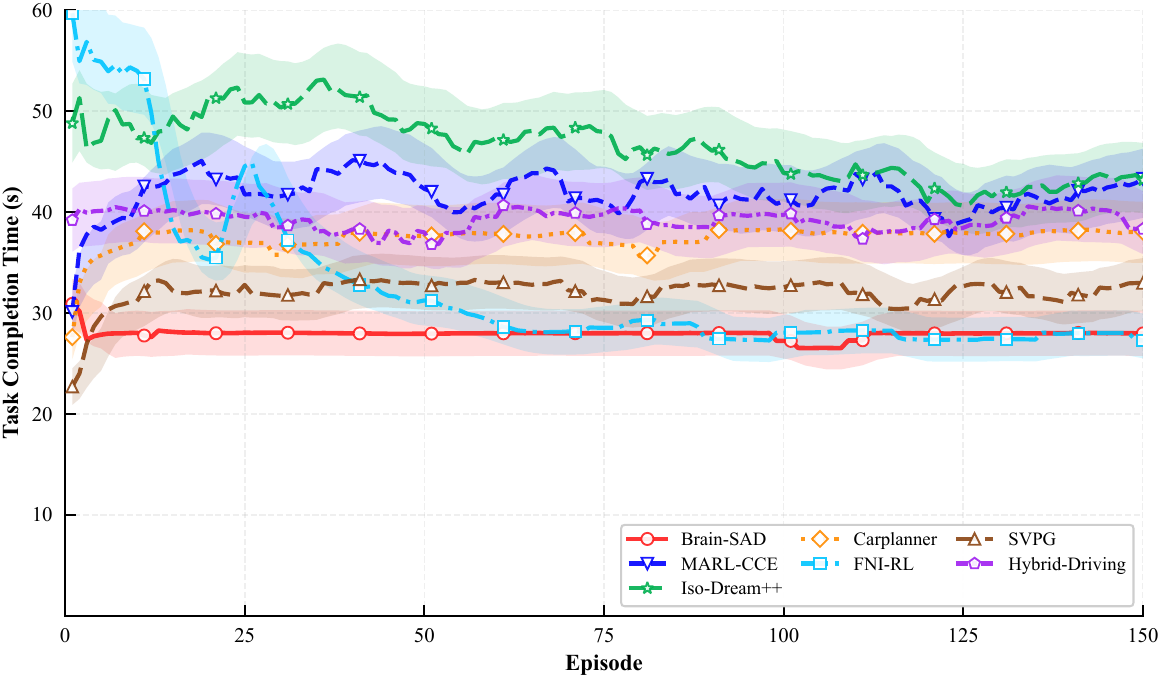}%
    }%
  \end{minipage}%
}%
\hspace{0.006\textwidth}%
\subfigure[Success Rate]{%
  \begin{minipage}[b]{0.238\textwidth}
    \centering
    \parbox[c][\FigSevenPanelHeight][c]{\linewidth}{%
      \centering
      \includegraphics[
        trim=0 0 413.7455 22,
        clip,
        width=\linewidth,
        height=\FigSevenPanelHeight,
        keepaspectratio
      ]{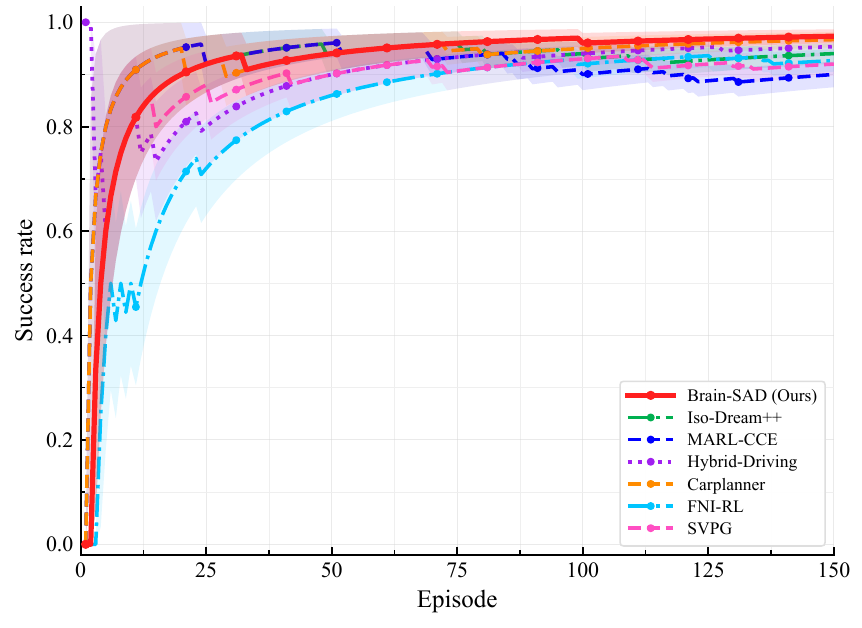}%
    }%
  \end{minipage}%
}%
\hspace{0.006\textwidth}%
\subfigure[Collision Rate]{%
  \begin{minipage}[b]{0.238\textwidth}
    \centering
    \parbox[c][\FigSevenPanelHeight][c]{\linewidth}{%
      \centering
      \includegraphics[
        trim=413.7455 0 0 22,
        clip,
        width=\linewidth,
        height=\FigSevenPanelHeight,
        keepaspectratio
      ]{Figs/Fig7-2.pdf}%
    }%
  \end{minipage}%
}%
\hspace{0.006\textwidth}%
\subfigure[Cumulative Average Reward]{%
  \begin{minipage}[b]{0.238\textwidth}
    \centering
    \parbox[c][\FigSevenPanelHeight][c]{\linewidth}{%
      \centering
      \includegraphics[
        width=\linewidth,
        height=\FigSevenPanelHeight,
        keepaspectratio
      ]{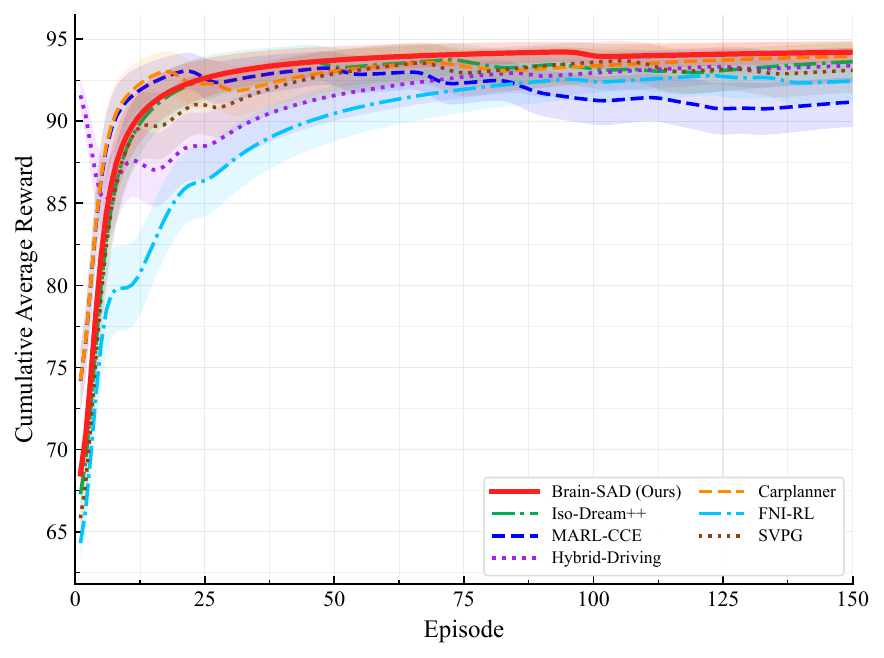}%
    }%
  \end{minipage}%
}%

\par\vspace{-0.5cm}

\subfigure[Iso-Dream++ action distribution]{%
  \begin{minipage}[b]{0.238\textwidth}
    \centering
    \parbox[c][\FigSevenPanelHeight][c]{\linewidth}{%
      \centering
      \includegraphics[
        trim=0 30 0 45,
        clip,
        width=\linewidth,
        height=\FigSevenPanelHeight,
        keepaspectratio
      ]{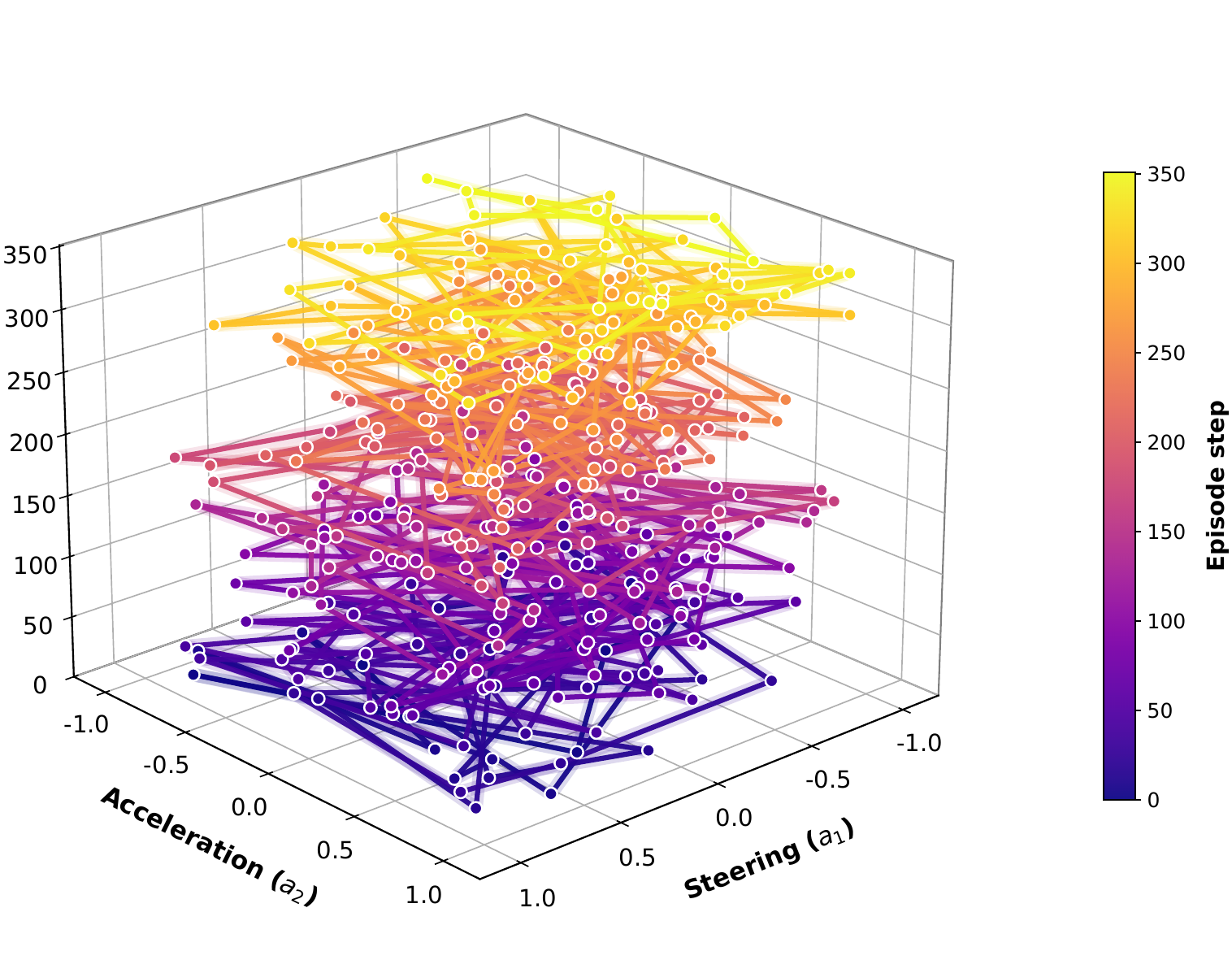}%
    }%
  \end{minipage}%
}%
\hspace{0.006\textwidth}%
\subfigure[Hybrid-Driving action distribution]{%
  \begin{minipage}[b]{0.238\textwidth}
    \centering
    \parbox[c][\FigSevenPanelHeight][c]{\linewidth}{%
      \centering
      \includegraphics[
        trim=0 30 0 45,
        clip,
        width=\linewidth,
        height=\FigSevenPanelHeight,
        keepaspectratio
      ]{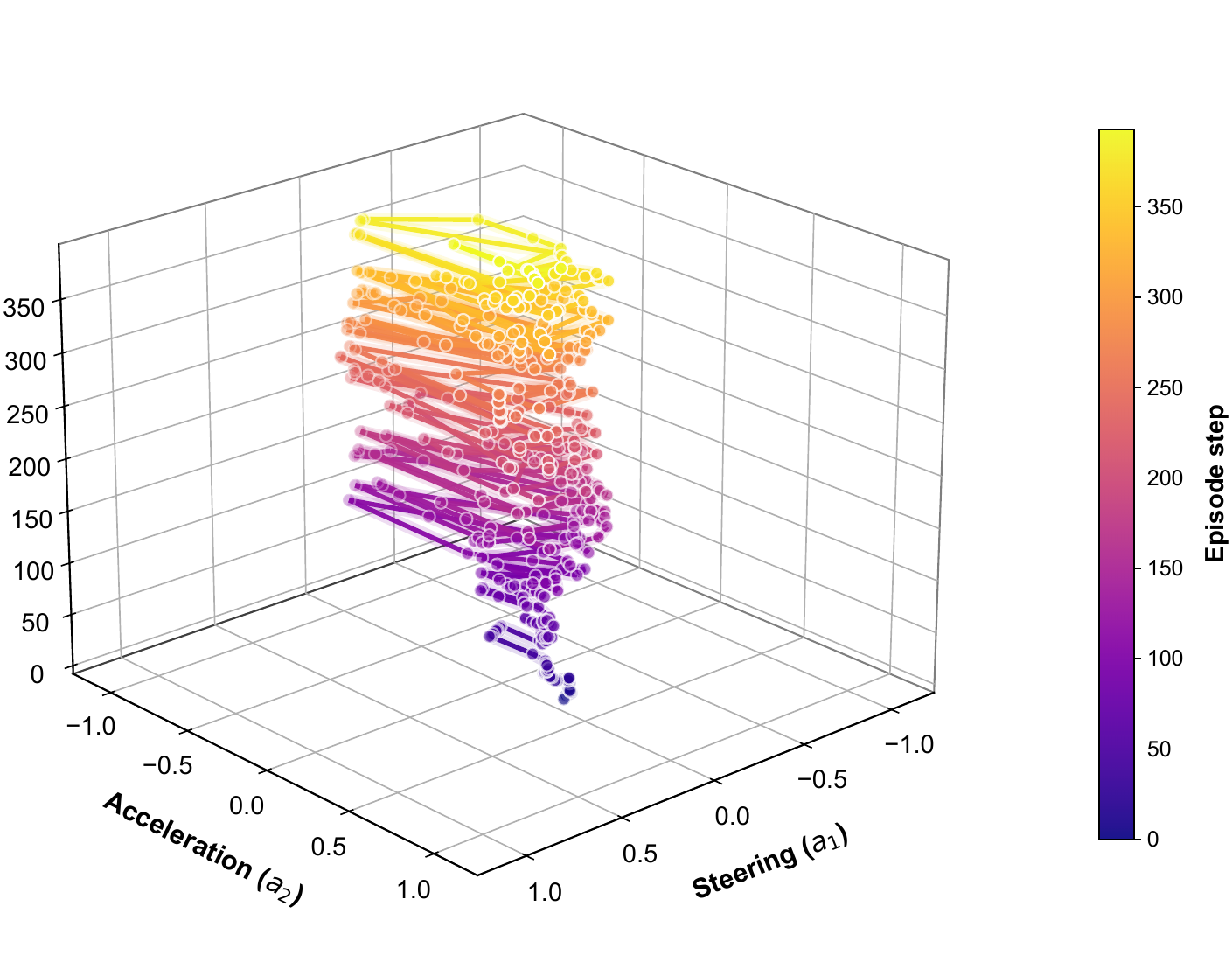}%
    }%
  \end{minipage}%
}%
\hspace{0.006\textwidth}%
\subfigure[SVPG action distribution]{%
  \begin{minipage}[b]{0.238\textwidth}
    \centering
    \parbox[c][\FigSevenPanelHeight][c]{\linewidth}{%
      \centering
      \includegraphics[
        trim=0 30 0 45,
        clip,
        width=\linewidth,
        height=\FigSevenPanelHeight,
        keepaspectratio
      ]{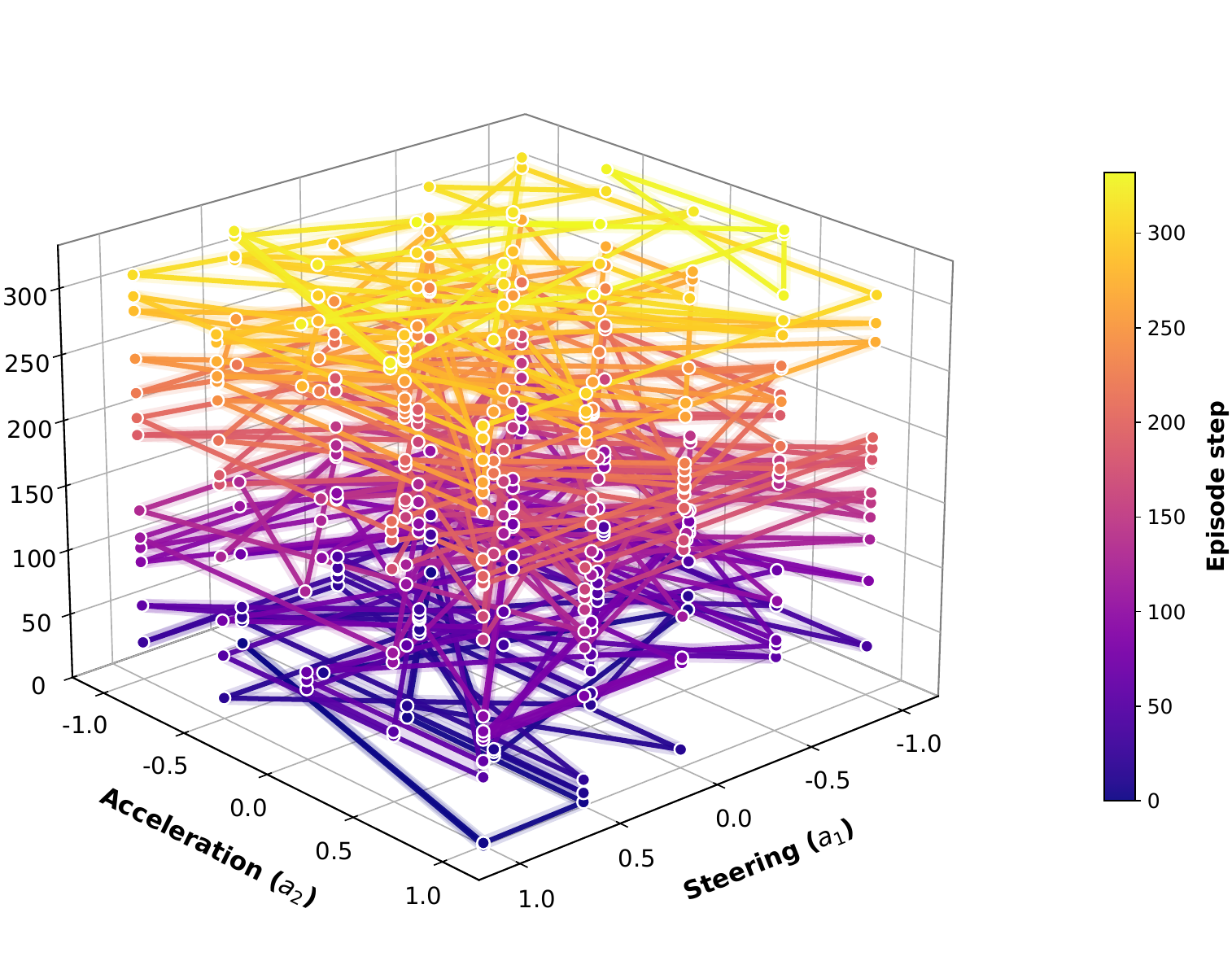}%
    }%
  \end{minipage}%
}%
\hspace{0.006\textwidth}%
\subfigure[Brain-SAD action distribution]{%
  \begin{minipage}[b]{0.238\textwidth}
    \centering
    \parbox[c][\FigSevenPanelHeight][c]{\linewidth}{%
      \centering
      \includegraphics[
        trim=0 30 0 45,
        clip,
        width=\linewidth,
        height=\FigSevenPanelHeight,
        keepaspectratio
      ]{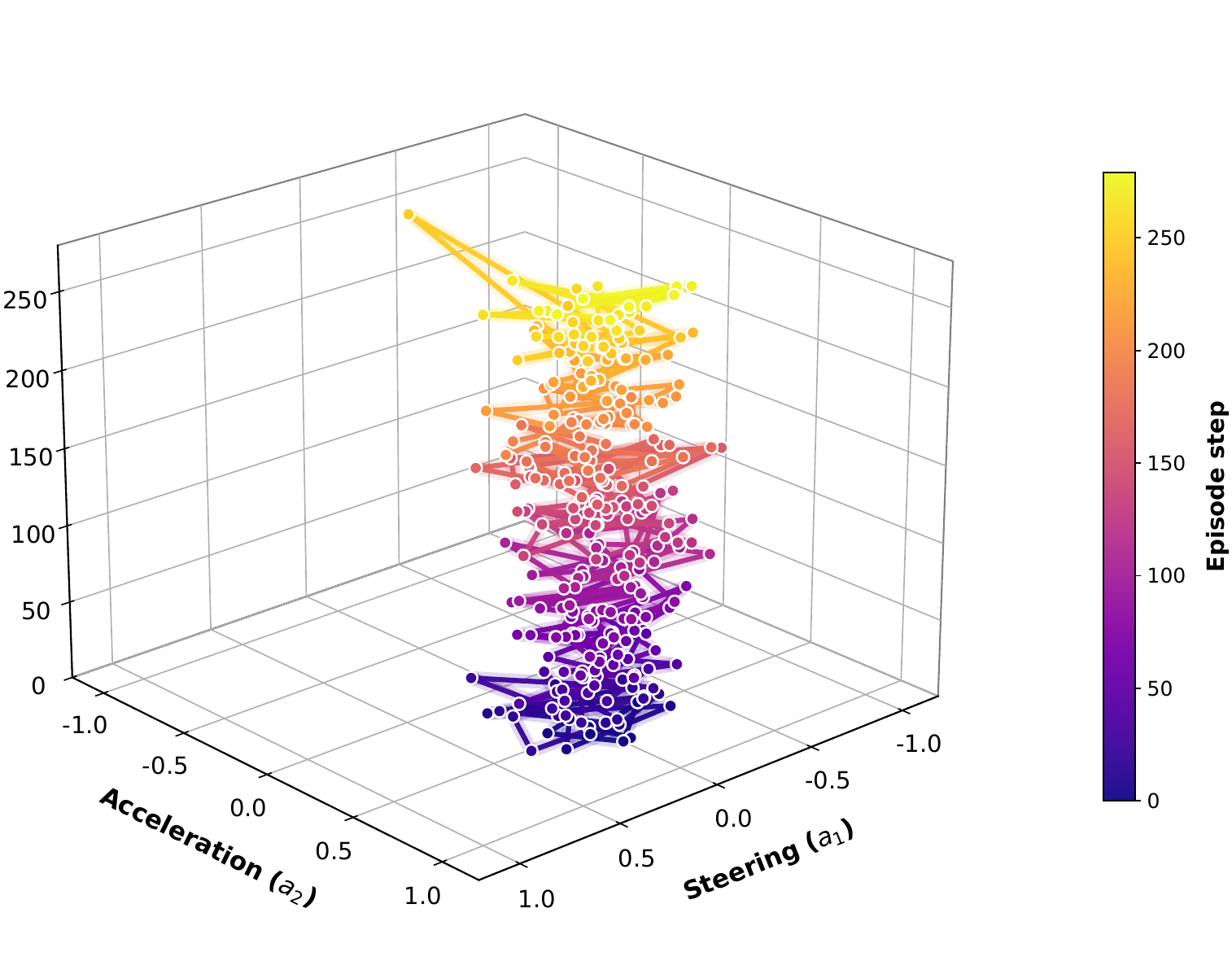}%
    }%
  \end{minipage}%
}%

\endgroup
\vspace{-0.3cm}

\caption{The training traces and action distributions on representative methods selected from \textbf{Table 4}, simulated in \textbf{Scene C}. (\textbf{NOTE}: In (\textit{a})--(\textit{d}), we present training traces on methods with TOP 2 \textbf{Reward} reported in \textbf{Table 4}; Here, the \textbf{Reward} are the average cumulative reward on the \textbf{\textit{best}} training round (150 episodes, 3 rounds in all); In (\textit{e})--(\textit{h}), we present action distributions across the \textbf{\textit{best episode}} (selected in the above best training round) on methods with TOP 1 \textbf{Reward} reported in \textbf{Table 4}, X-axis for acceleration positive as speeding, Y-axis for steering angle, Z-axis for temporal steps in the best episode.)}
\label{Fig:7}
\end{figure*}

\FloatBarrier

\subsection{Stress Test by Continuous Intersections}
\label{sec:continuous_intersections}

Based on all the mentioned above, in Fig. 8, we further construct 6 continuous intersections as Stress Test on Brain-SAD to ascertain whether Brain-SAD could still present corresponding reliability in the long-term driving task. In Fig. 8 (\textit{left}), the interaction complexity of each intersection has been disturbed (i.e., the density of traffic flow like the number of ahead or aside interactions), making the overall interaction complexity fluctuated as shown in Fig. 8 (\textit{right})\footnotemark. In this way, we select three representative brain-inspired action control methods according to the performance in Table 3 \& Table 4 for further discussion, including \textbf{SVPG} \cite{25}, spiking RL policy with synaptic plasticity; \textbf{FNI-RL} \cite{26}, static state-level fear-cost constrained RL policy; \textbf{EFT-RL} \cite{28}, goal-to-subgoal RL policy as human-problem solving.

\setcounter{bsStressFigureBase}{\value{figure}}
\setcounter{figure}{\value{bsStressFigureBase}}
\addtocounter{figure}{1}
%
\begin{table*}[!b]
\centering

\begingroup
\setlength{\abovecaptionskip}{0pt}
\setlength{\belowcaptionskip}{4pt}
{\captionof{table}{The performance of Brain-SAD on \textbf{continuous Stress Test across 6 intersections} in Fig. \ref{Fig:8} (\textbf{NOTE:} the \textbf{\textit{bold}} font represents the most optimal performance in each column, with the colored as the sub-optimal in each column. \textbf{ALL} the performance are averaged by \textbf{three} rounds).}
\label{tab:continuous_stress}}
\centering
\renewcommand{\arraystretch}{1.04}
\resizebox{\textwidth}{!}{%
\begin{tabular}{clccccccc}
\hline
\textbf{NO} & \textbf{Method} & \textbf{TCT $\downarrow$(s)} & \textbf{RECT $\downarrow$(s)} & \textbf{SR$\uparrow$(\%)} & \textbf{Reward$\uparrow$} & \textbf{SLC $\uparrow$(\%)} & \textbf{Comfortable $\uparrow$(\%)} & \textbf{Median-TTC$\uparrow$(s)} \\
\hline
1 & SVPG & \cellcolor{bsPink}157.00 $\pm$ 1.70 & 0.6868 $\pm$ 0.0145 & 88.89 $\pm$ 2.34 & 89.49 $\pm$ 0.52 & 82.33 $\pm$ 1.32 & 82.26 $\pm$ 0.15 & 1.3278 $\pm$ 0.0020 \\
2 & FNI-RL & 120.90 $\pm$ 17.60 & \cellcolor{bsPink}0.6507 $\pm$ 0.0027 & \cellcolor{bsPink}92.22 $\pm$ 0.38 & \cellcolor{bsPink}91.14 $\pm$ 4.08 & \cellcolor{bsPink}84.43 $\pm$ 0.50 & \cellcolor{bsPink}83.43 $\pm$ 0.15 & 1.2488 $\pm$ 0.1132 \\
3 & EFT-RL & 238.70 $\pm$ 21.70 & 0.7868 $\pm$ 0.0444 & 80.44 $\pm$ 11.16 & 86.38 $\pm$ 2.92 & 77.28 $\pm$ 2.43 & 79.53 $\pm$ 1.43 & \cellcolor{bsPink}\textbf{1.5418 $\pm$ 0.0899} \\
4 & \cellcolor{bsYellow}\textbf{Brain-SAD} & \cellcolor{bsYellow}\textbf{99.90 $\pm$ 0.70} & \cellcolor{bsYellow}\textbf{0.5702 $\pm$ 0.0109} & \cellcolor{bsYellow}\textbf{94.00 $\pm$ 0.67} & \cellcolor{bsYellow}\textbf{97.37 $\pm$ 0.17} & \cellcolor{bsYellow}\textbf{85.88 $\pm$ 0.02} & \cellcolor{bsYellow}\textbf{86.73 $\pm$ 0.05} & \cellcolor{bsYellow}1.4344 $\pm$ 0.0061 \\
\hline
\end{tabular}
}
\endgroup

\vspace{0.03cm}

\setlength{\abovecaptionskip}{0.235cm}
\setlength{\belowcaptionskip}{0pt}
\centering
%
\begingroup

\makeatletter
\def\@captype{figure}
\makeatother
\subfiguretopcapfalse
\setcounter{subfigure}{0}
\newlength{\FigNinePanelHeight}
\setlength{\FigNinePanelHeight}{0.164\textwidth}
\renewcommand*{\subcapsize}{\scriptsize}
\setlength{\subfigcapskip}{1.5pt}

\subfigure[Task Completion Time]{%
  \begin{minipage}[b]{0.238\textwidth}
    \centering
    \parbox[c][\FigNinePanelHeight][c]{\linewidth}{%
      \centering
      \includegraphics[
        width=\linewidth,
        height=\FigNinePanelHeight,
        keepaspectratio
      ]{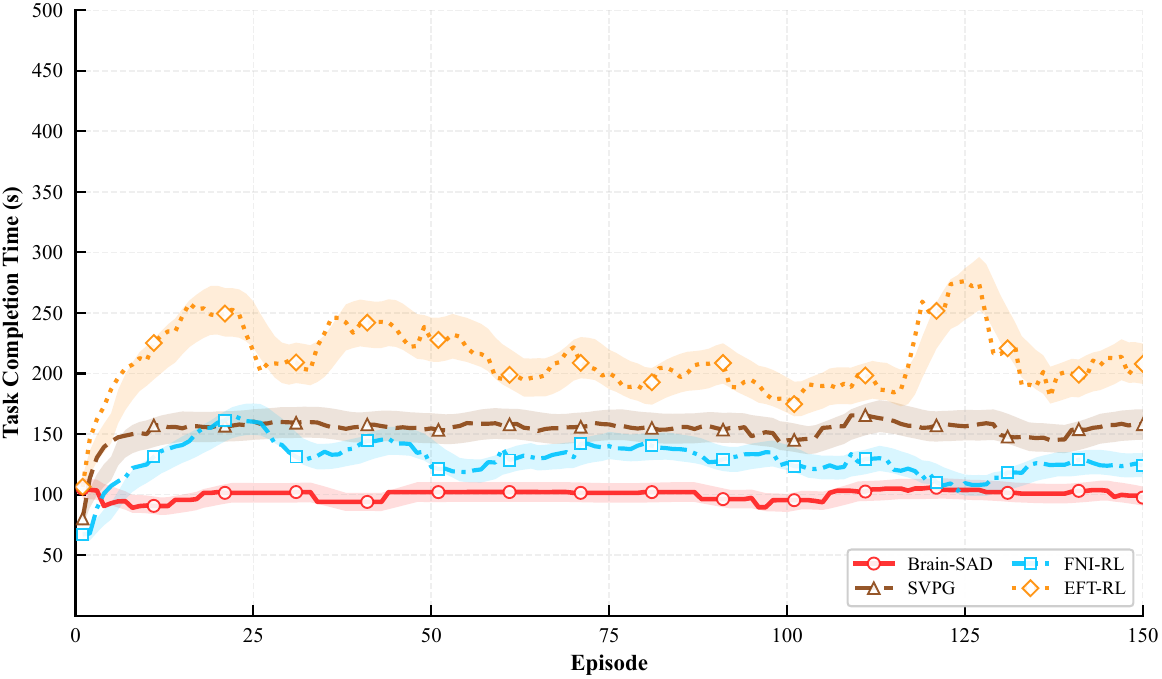}%
    }%
  \end{minipage}%
}%
\hspace{0.006\textwidth}%
\subfigure[Success Rate]{%
  \begin{minipage}[b]{0.238\textwidth}
    \centering
    \parbox[c][\FigNinePanelHeight][c]{\linewidth}{%
      \centering
      \includegraphics[
        trim=0 0 413.7455 22,
        clip,
        width=\linewidth,
        height=\FigNinePanelHeight,
        keepaspectratio
      ]{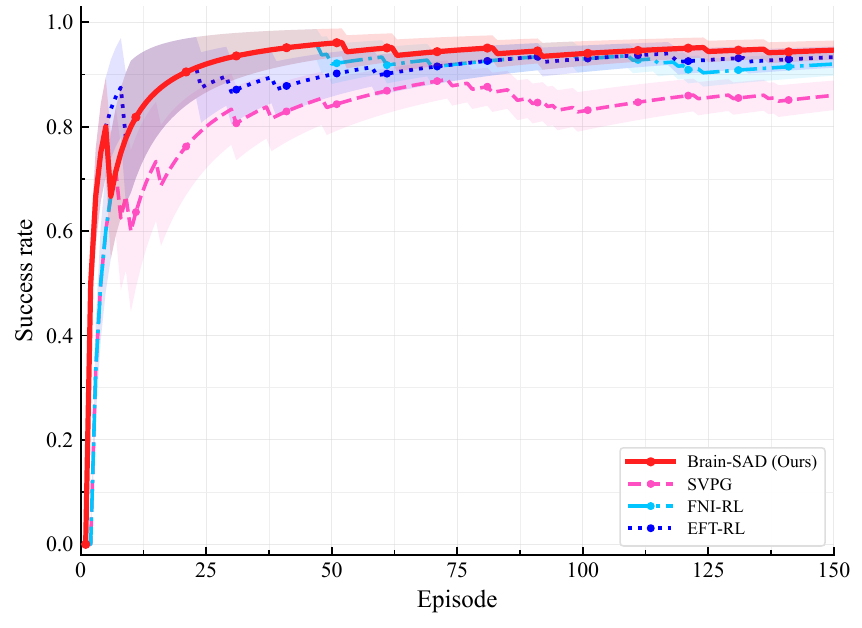}%
    }%
  \end{minipage}%
}%
\hspace{0.006\textwidth}%
\subfigure[Collision Rate]{%
  \begin{minipage}[b]{0.238\textwidth}
    \centering
    \parbox[c][\FigNinePanelHeight][c]{\linewidth}{%
      \centering
      \includegraphics[
        trim=413.7455 0 0 22,
        clip,
        width=\linewidth,
        height=\FigNinePanelHeight,
        keepaspectratio
      ]{Figs/Fig9-2.pdf}%
    }%
  \end{minipage}%
}%
\hspace{0.006\textwidth}%
\subfigure[Cumulative Average Reward]{%
  \begin{minipage}[b]{0.238\textwidth}
    \centering
    \parbox[c][\FigNinePanelHeight][c]{\linewidth}{%
      \centering
      \includegraphics[
        width=\linewidth,
        height=\FigNinePanelHeight,
        keepaspectratio
      ]{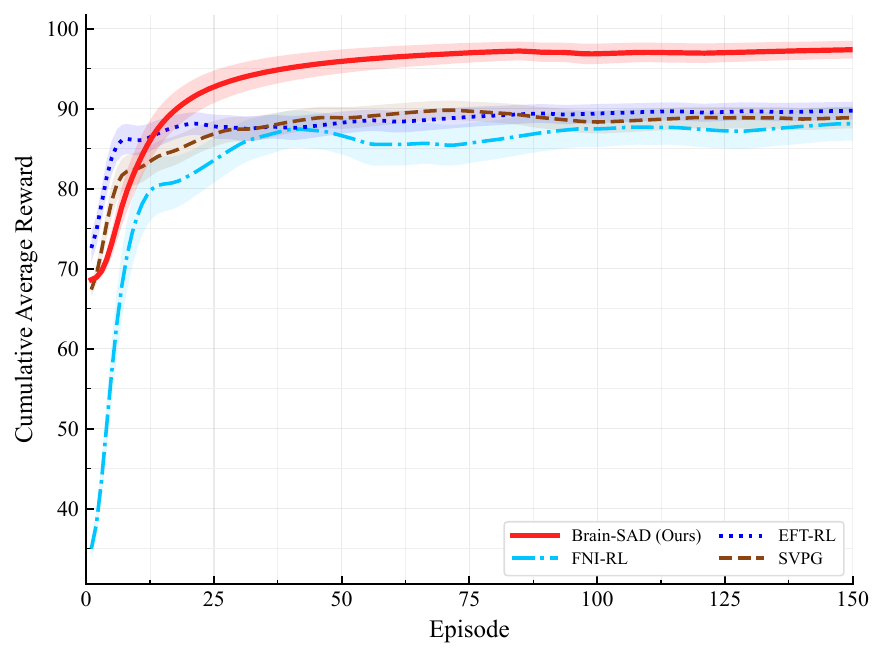}%
    }%
  \end{minipage}%
}%

\par\vspace{-0.3cm}

\subfigure[SVPG action distribution]{%
  \begin{minipage}[b]{0.238\textwidth}
    \centering
    \parbox[c][\FigNinePanelHeight][c]{\linewidth}{%
      \centering
      \includegraphics[
        trim=0 30 0 45,
        clip,
        width=\linewidth,
        height=\FigNinePanelHeight,
        keepaspectratio
      ]{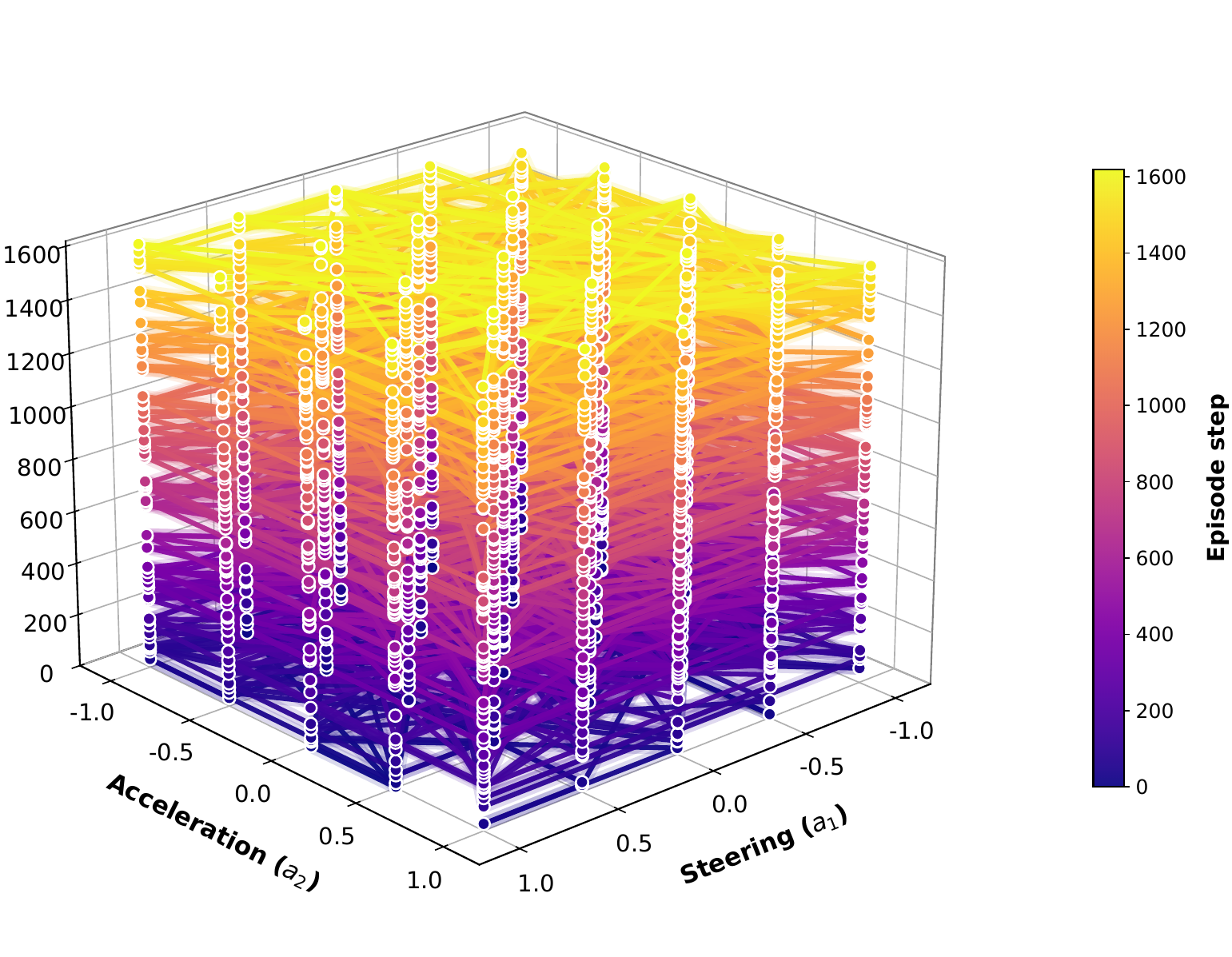}%
    }%
  \end{minipage}%
}%
\hspace{0.006\textwidth}%
\subfigure[FNI-RL action distribution]{%
  \begin{minipage}[b]{0.238\textwidth}
    \centering
    \parbox[c][\FigNinePanelHeight][c]{\linewidth}{%
      \centering
      \includegraphics[
        trim=0 30 0 45,
        clip,
        width=\linewidth,
        height=\FigNinePanelHeight,
        keepaspectratio
      ]{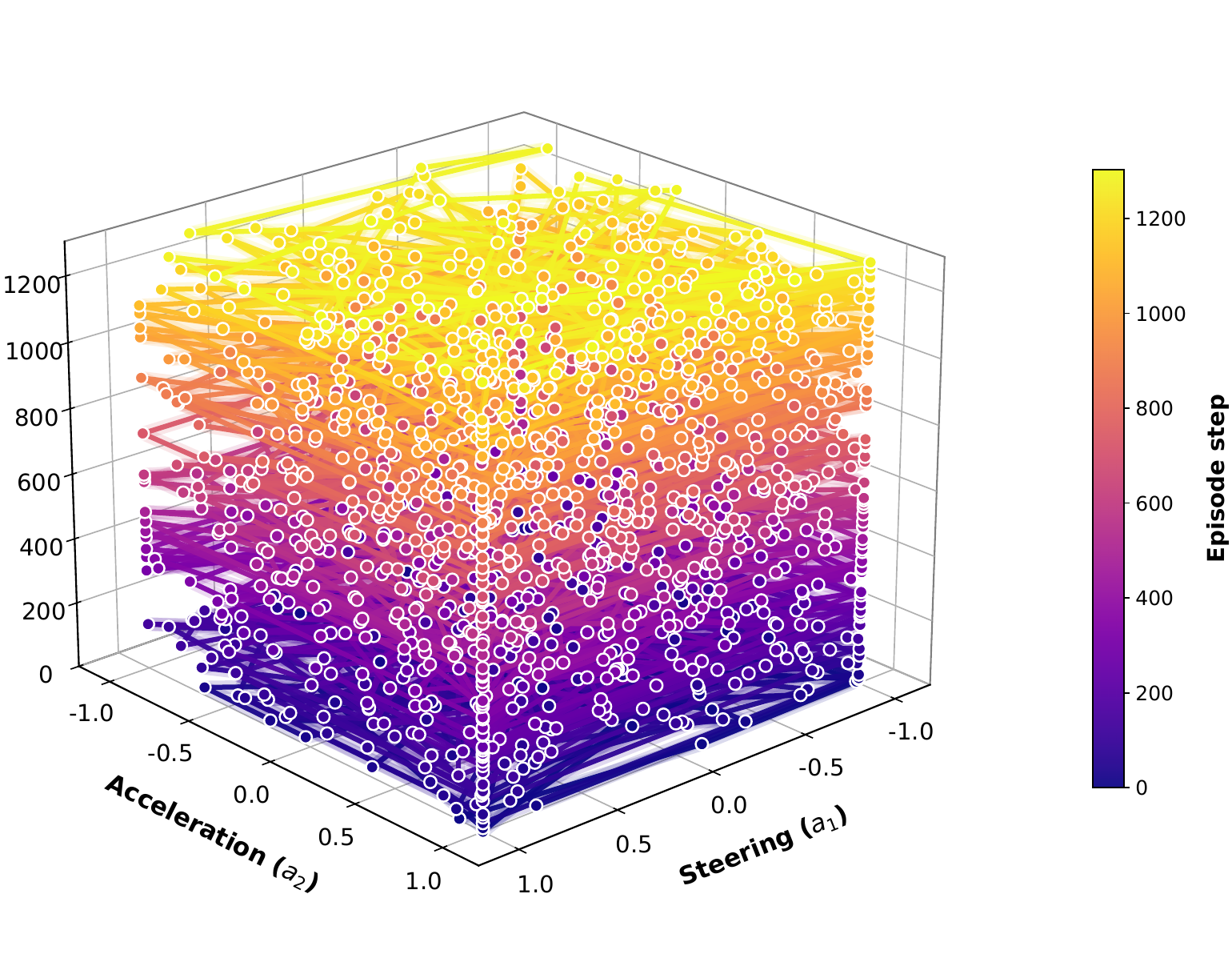}%
    }%
  \end{minipage}%
}%
\hspace{0.006\textwidth}%
\subfigure[EFT-RL action distribution]{%
  \begin{minipage}[b]{0.238\textwidth}
    \centering
    \parbox[c][\FigNinePanelHeight][c]{\linewidth}{%
      \centering
      \includegraphics[
        trim=0 30 0 45,
        clip,
        width=\linewidth,
        height=\FigNinePanelHeight,
        keepaspectratio
      ]{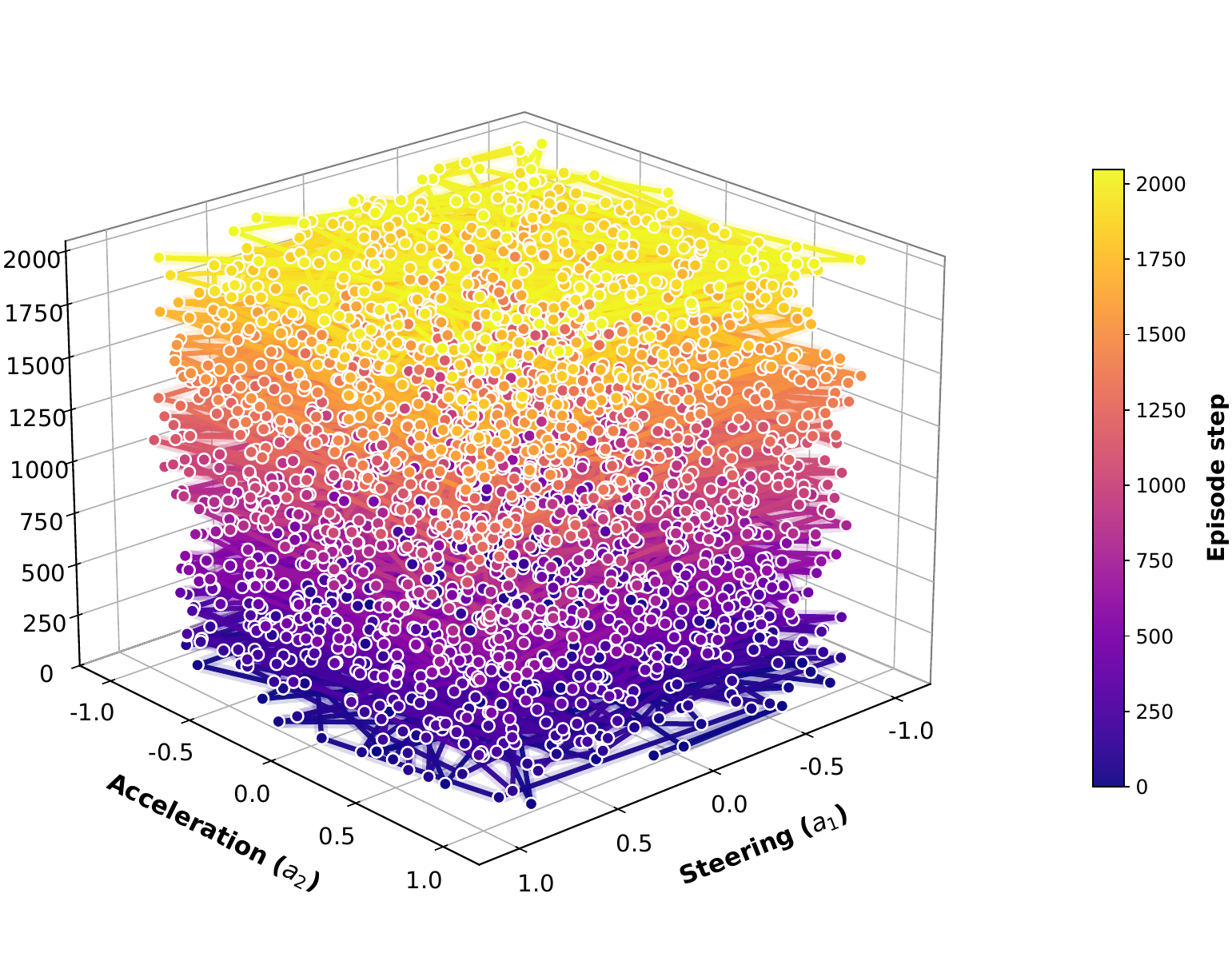}%
    }%
  \end{minipage}%
}%
\hspace{0.006\textwidth}%
\subfigure[Brain-SAD action distribution]{%
  \begin{minipage}[b]{0.238\textwidth}
    \centering
    \parbox[c][\FigNinePanelHeight][c]{\linewidth}{%
      \centering
      \includegraphics[
        trim=0 30 0 45,
        clip,
        width=\linewidth,
        height=\FigNinePanelHeight,
        keepaspectratio
      ]{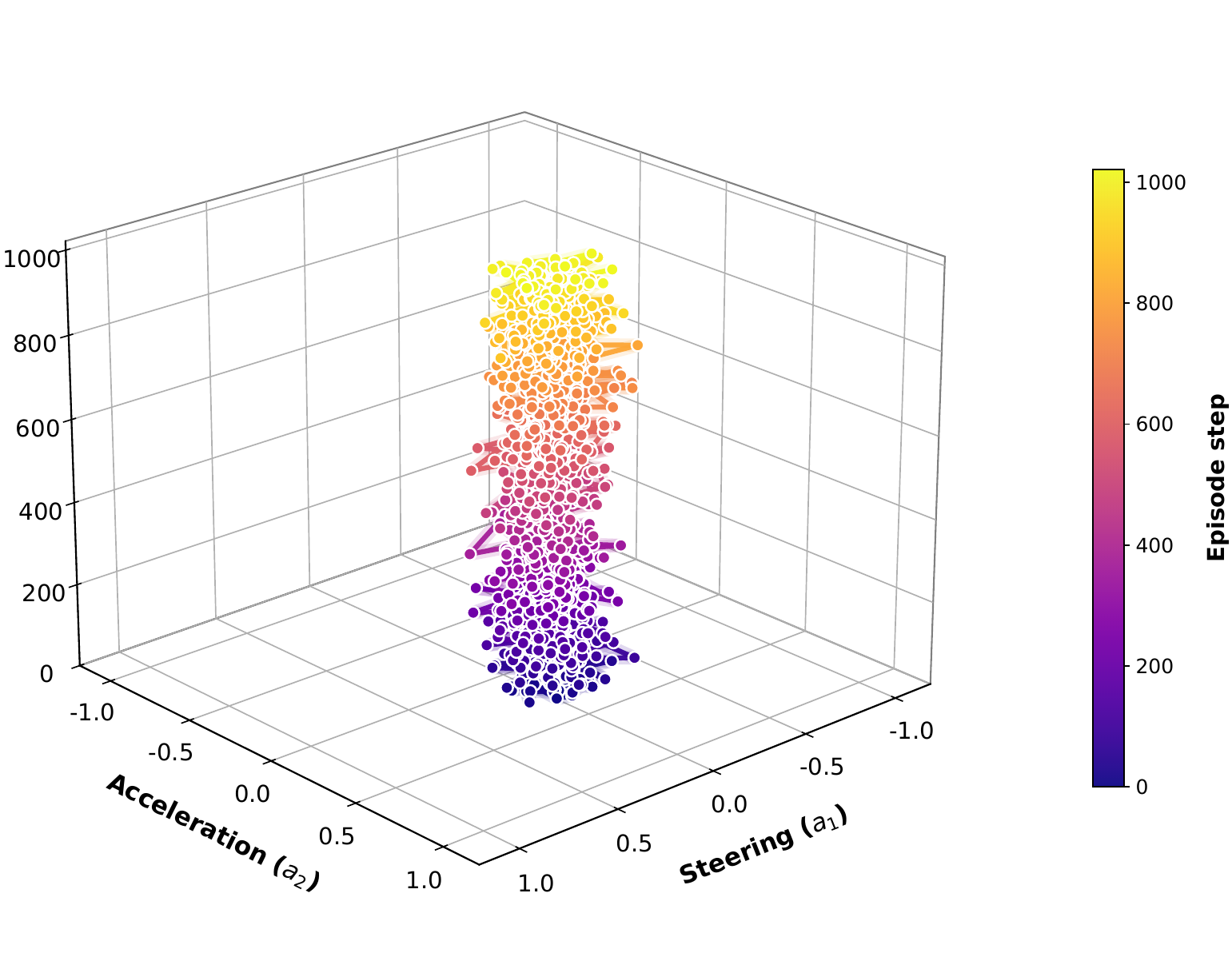}%
    }%
  \end{minipage}%
}%

\endgroup

\captionof{figure}{The training traces and action distributions on SVPG, EFT-RL, FNI-RL and Brain-SAD, simulated in 6 continuous intersections shown in \textbf{Fig. 8}. (\textbf{NOTE}: In (\textit{a})--(\textit{d}), we present training traces on methods in \textbf{Table 5}; Here, the \textbf{Reward} are the average cumulative reward on the \textbf{\textit{best}} training round (150 episodes, 3 rounds in all); In (\textit{e})--(\textit{h}), we present action distributions across the \textbf{\textit{best episode}} (selected in the above best training round), X-axis for acceleration positive as speeding, Y-axis for steering angle, Z-axis for temporal steps in the best episode.)}
\label{Fig:9}
\end{table*}

\setcounter{figure}{\value{bsStressFigureBase}}

\vspace{0.10cm}
\noindent\begin{minipage}{\columnwidth}
\centering
\setlength{\abovecaptionskip}{0.2cm}
\setlength{\belowcaptionskip}{0pt}

\noindent\makebox[\columnwidth][c]{%
  \begin{minipage}[c]{0.488\columnwidth}
    \centering
    \includegraphics[
      width=\linewidth,
      keepaspectratio
    ]{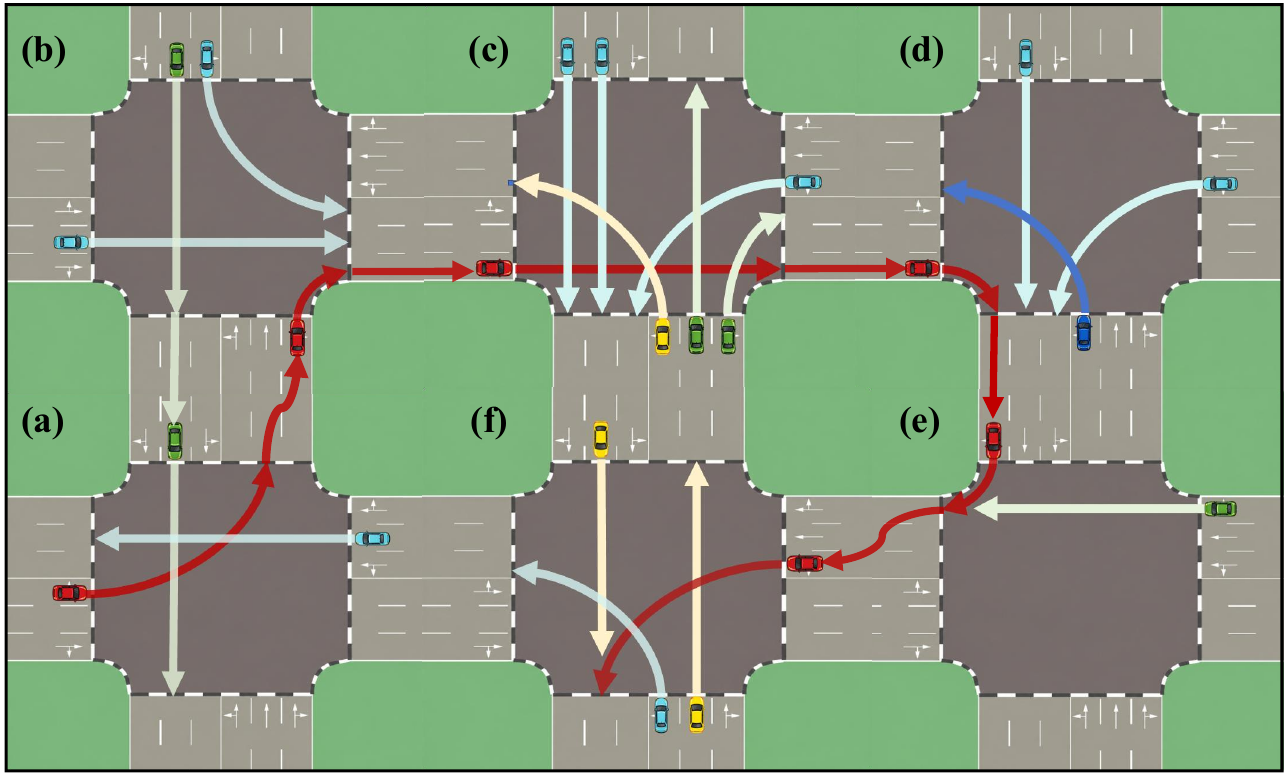}%
  \end{minipage}%
  \hfill
  \begin{minipage}[c]{0.488\columnwidth}
    \centering
    \includegraphics[
      width=\linewidth,
      keepaspectratio
    ]{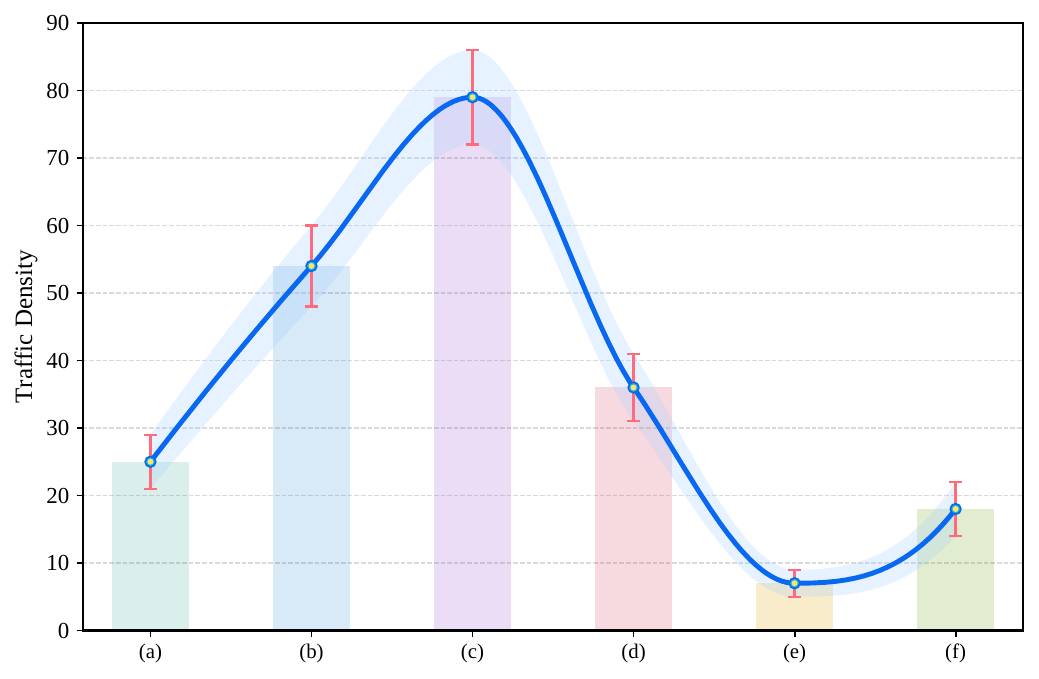}%
  \end{minipage}%
}
\vspace{-0.3cm}
{\footnotesize
\captionof{figure}{The scenario configuration with fluctuated interaction complexity
across 6 continuous intersections. (\textbf{NOTE}: the distance between
adjacent intersections are not fixed and ignored in Fig. 8)}
\label{Fig:8}
}
\vspace{-0.1cm}

\end{minipage}
\par
\footnotetext{See detailed configuration on Fig. 8 in Appendix \textbf{D}.}

\setcounter{figure}{\value{bsStressFigureBase}}
\addtocounter{figure}{2}
\vspace{0.20cm}
First, as shown in Table 5, our proposed Brain-SAD can still maintain \textbf{\textit{faster}} and \textbf{\textit{better}} even in Stress Test, achieving the \textbf{\textit{highest}} \textbf{SR} and \textbf{Reward} with the \textbf{\textit{shortest}} \textbf{TCT} and \textbf{RECT} time across 6 continuous intersections shown in Fig. 8. We also open up the training traces on \textbf{TCT}, \textbf{SR}, Collision Rate (\textbf{CR}) and \textbf{Reward} across the \textbf{\textit{best}} training round (with the highest average cumulative reward) in Fig. 9 (\textit{a})--(\textit{d}), where Brain-SAD has behaved better than SVPG, FNI-RL and EFT-RL with more obvious convergence at the most optimal level. Such the observation proves that by imposing \textbf{\textit{dynamic fear constraint}} coupled with action-impact (or \textbf{Value Estimation}) (see eq (5--8)), our proposed Brain-SAD has the reliability to resist the varying environmental disturbance, where the policy action distribution of Brain-SAD can be \textbf{\textit{online}} optimized by the most optimal posterior-derived action weights (see eq (9--11)), adaptive to the evolving intersections in time.

Second, our proposed Brain-SAD has achieved the most optimal \textbf{SLC}, \textbf{Comfortable} in Table 5. That is, across the 6 continuous intersections with fluctuated complexity, Brain-SAD has been adapted to continuous scene-evolving with \textit{higher} Speed Compliance (\textbf{SLC}), \textit{higher} stability on adjacent speeding to be \textbf{Comfortable}. For \textbf{Median-TTC} (Time-to-Collision, reflected by relative distance to neighbors) in Table 5, although our Brain-SAD has achieved the sub-optimal level than that of EFT-RL, yet the \textbf{TCT} on task completion and \textbf{RECT} on collision avoidance of EFT-RL is \textbf{\textit{much longer}} than Brain-SAD. Such the phenomenon can be further explained by the action distribution shown in Fig. 9 (\textit{e})--(\textit{h}), where we scatter and fully-connect all the (steer, accelerate) actions across the best episode selected in Fig. 9 (\textit{a})--(\textit{d}). It can be detected in Fig. 9 (\textit{h}) that, even facing continuous scene-evolving with fluctuated interaction complexity, Brain-SAD has still converged to fixed pattern with acceleration crowding at (-0.5-braking, 0.5-speeding) scope. However, since EFT-RL makes its own actions by predicting others to avoid collision, more fierce action shaking has occurred (see Fig. 9 (\textit{g})), presenting the characteristic as \textbf{\textit{slower}} in task completion, \textbf{\textit{better}} kept in relative distance (or higher Median-TTC in Table 5).

Consequently, based on the fierce action distribution shaking of SVPG, FNI-RL, EFT-RL (see Fig. 9 (\textit{e})--(\textit{g})), it can be drawn that the compared brain-inspired policy mainly learns \textit{how to avoid collision}, while according to the converged action distribution of Brain-SAD in Fig. 9 (\textit{h}), our Brain-SAD can learn-then-handle collision with constant speeding/braking in confidence, thus leading to the closer distance to neighbors. Such the conclusion can also be observed in Fig. 10, where the interaction trajectories with action intensity across 6 intersections have been compared between FNI-RL and Brain-SAD. It can be detected that compared with FNI-RL, Brain-SAD has passed each intersection with smoother traces in lower action intensity.

\begin{figure*}[!t]
\setlength{\abovecaptionskip}{0.18cm}
\setlength{\belowcaptionskip}{0pt}
\centering

%
\begingroup

\newlength{\FigTenPanelWidth}
\newlength{\FigTenGraphicWidth}
\newlength{\FigTenGraphicHeight}
\newlength{\FigTenGap}
\newlength{\FigTenLabelHeight}

\setlength{\FigTenPanelWidth}{0.159\textwidth}
\setlength{\FigTenGraphicWidth}{0.158\textwidth}
\setlength{\FigTenGraphicHeight}{0.151\textwidth}
\setlength{\FigTenGap}{0.0085\textwidth}
\setlength{\FigTenLabelHeight}{1.62em}

\newcommand{\FigTenCell}[3]{%
  \begin{minipage}[t]{\FigTenPanelWidth}%
    \centering
    \parbox[c][\FigTenGraphicHeight][c]{\linewidth}{%
      \centering
      \includegraphics[
        width=\FigTenGraphicWidth,
        height=\FigTenGraphicHeight
      ]{#1}%
    }%
    \par
    \vspace{1.0pt}%
    \parbox[c][\FigTenLabelHeight][c]{\linewidth}{%
      \centering
      \fontsize{7.25pt}{8.0pt}\selectfont
      \strut\textit{(#2)}~#3\strut
    }%
  \end{minipage}%
}

\noindent\makebox[\textwidth][c]{%
\FigTenCell{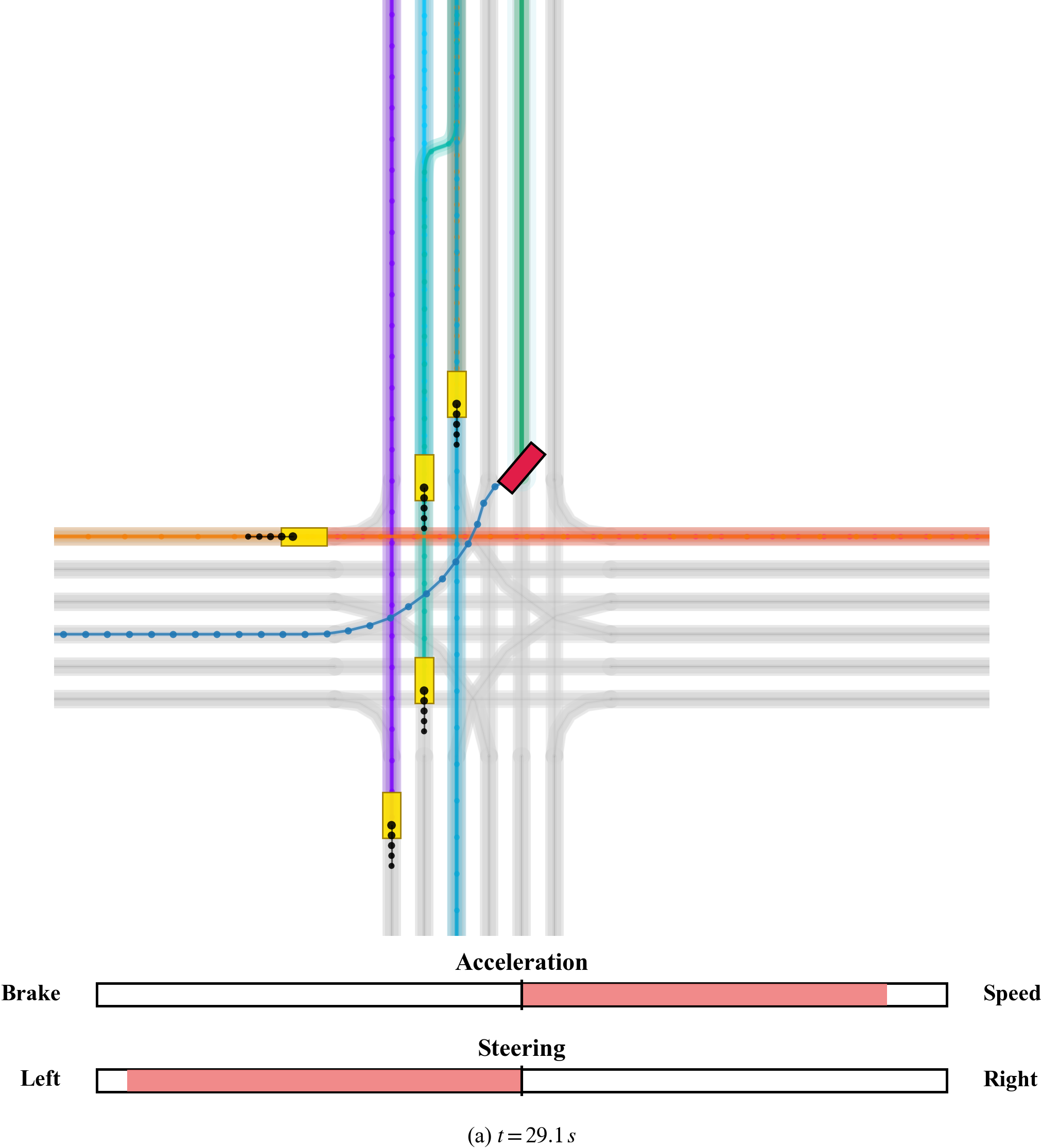}{a}{FNI-RL, 29.1s}%
\hspace{\FigTenGap}%
\FigTenCell{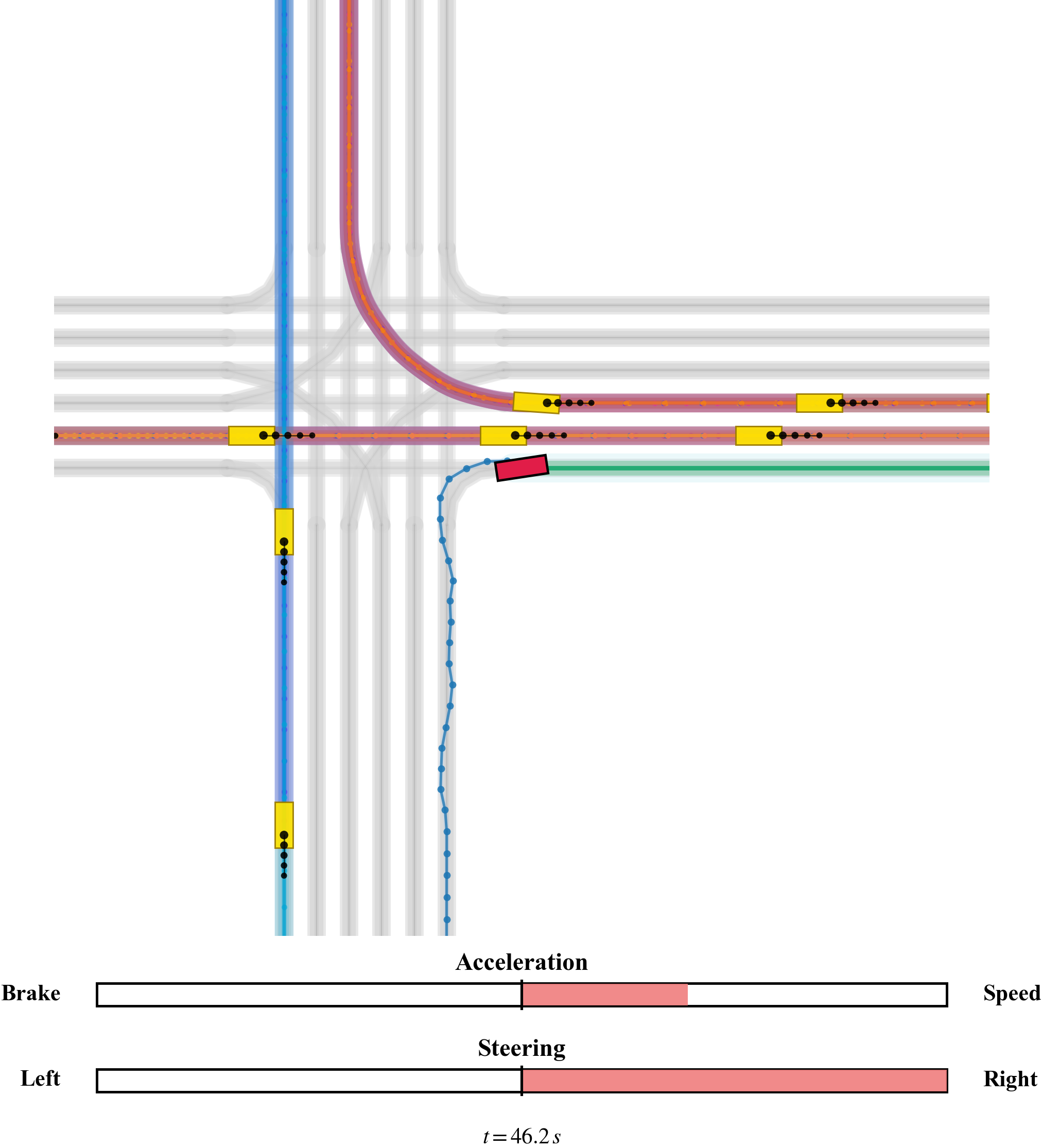}{b}{FNI-RL, 46.2s}%
\hspace{\FigTenGap}%
\FigTenCell{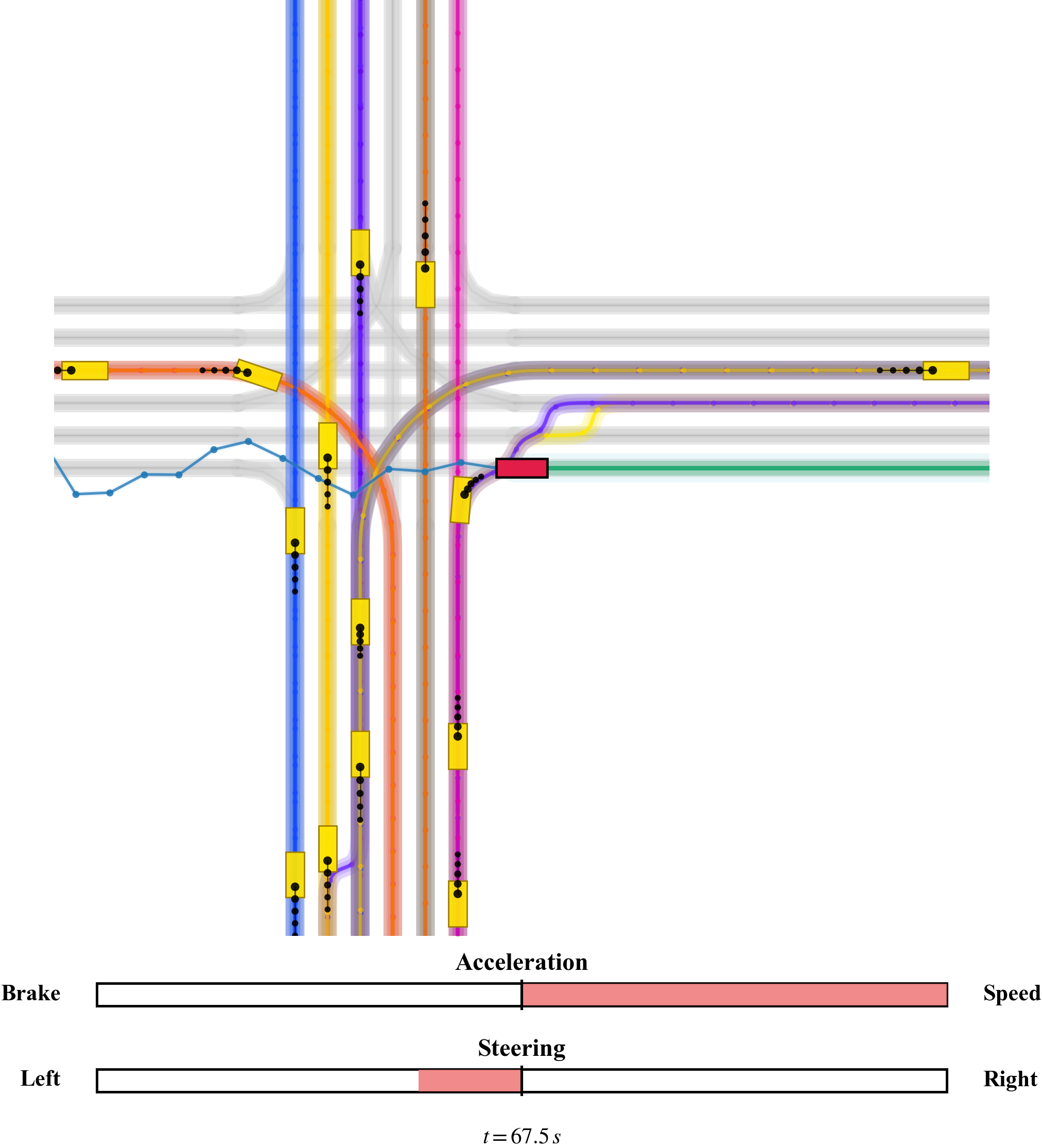}{c}{FNI-RL, 67.5s}%
\hspace{\FigTenGap}%
\FigTenCell{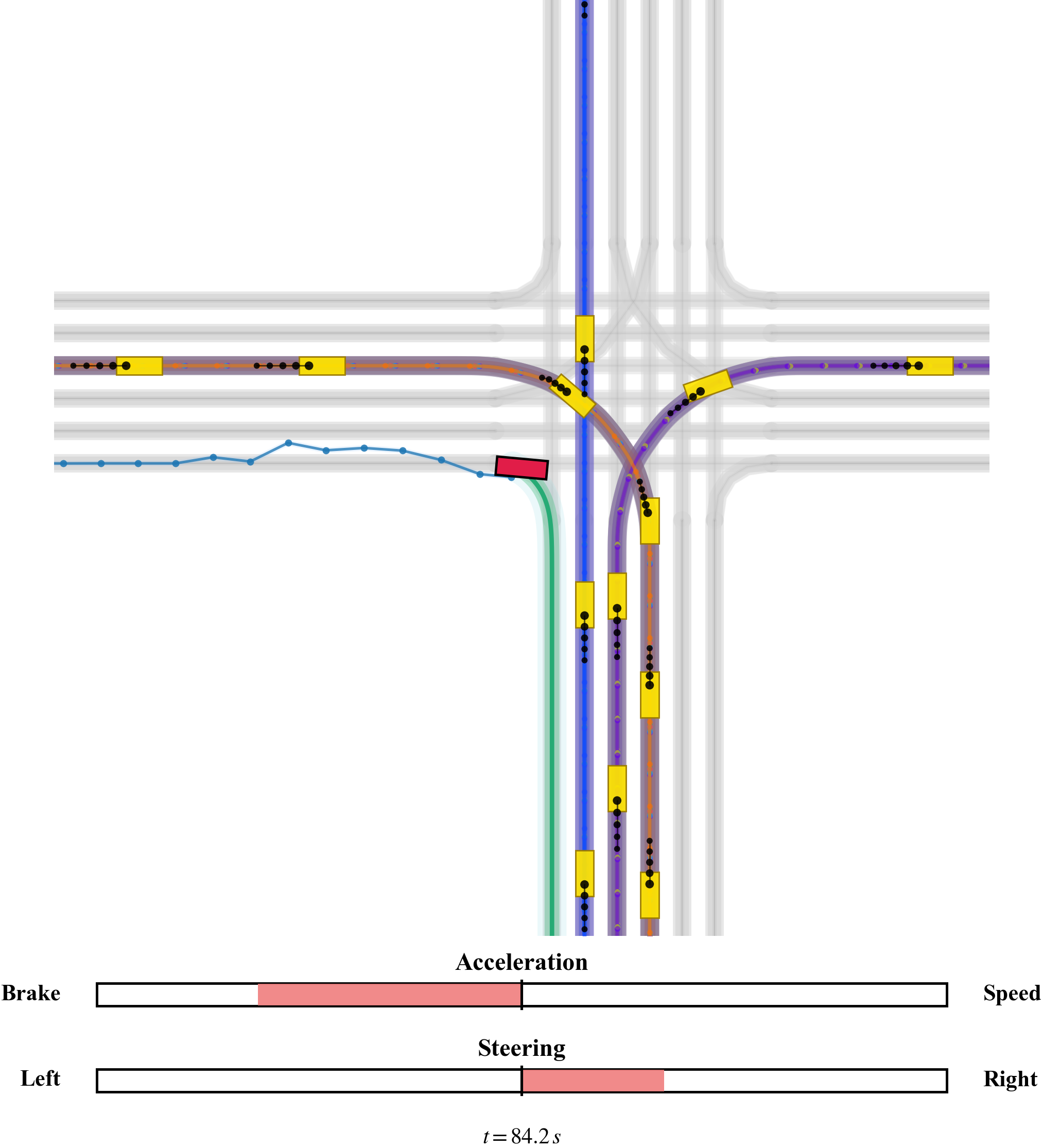}{d}{FNI-RL, 84.2s}%
\hspace{\FigTenGap}%
\FigTenCell{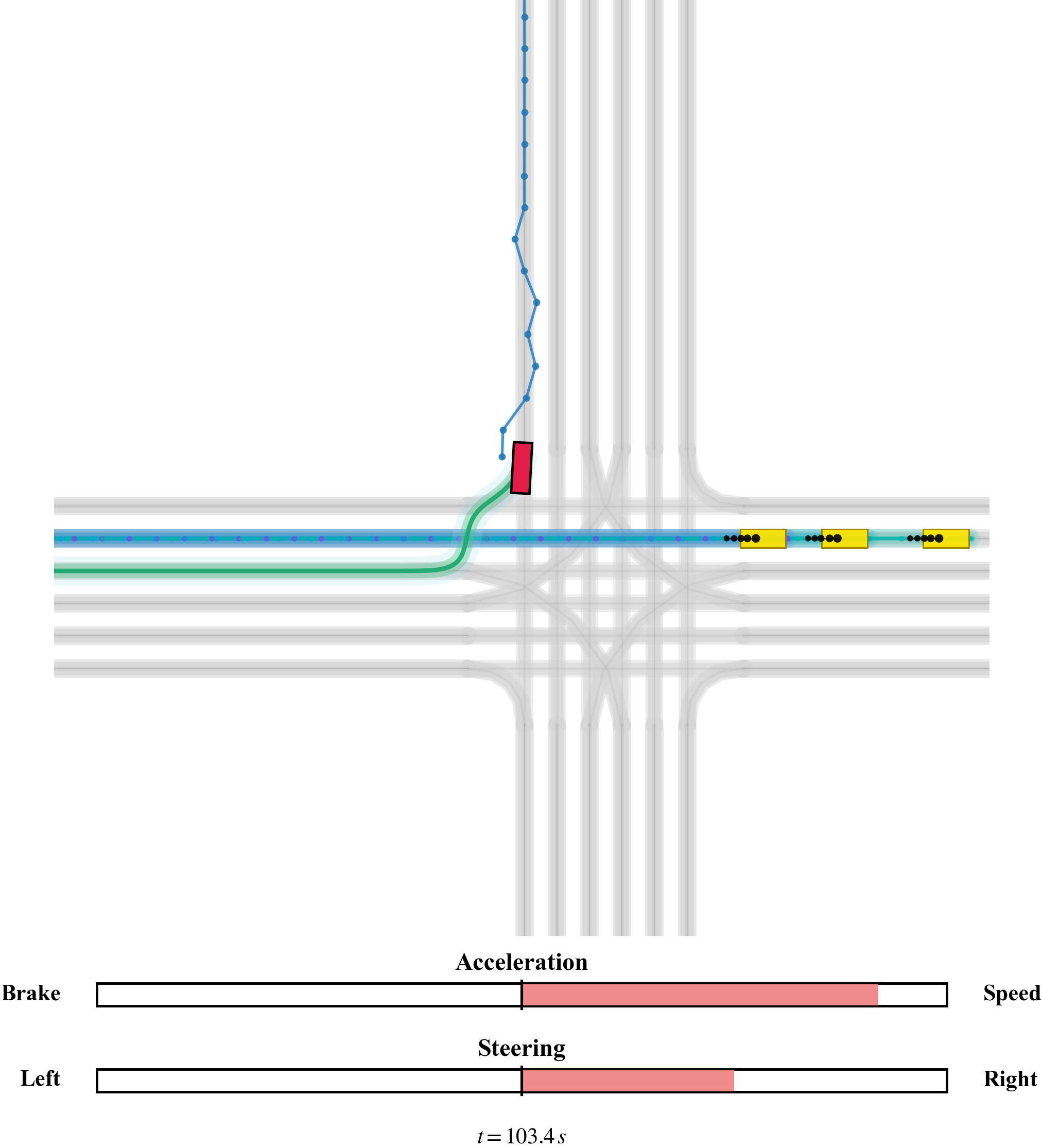}{e}{FNI-RL, 103.4s}%
\hspace{\FigTenGap}%
\FigTenCell{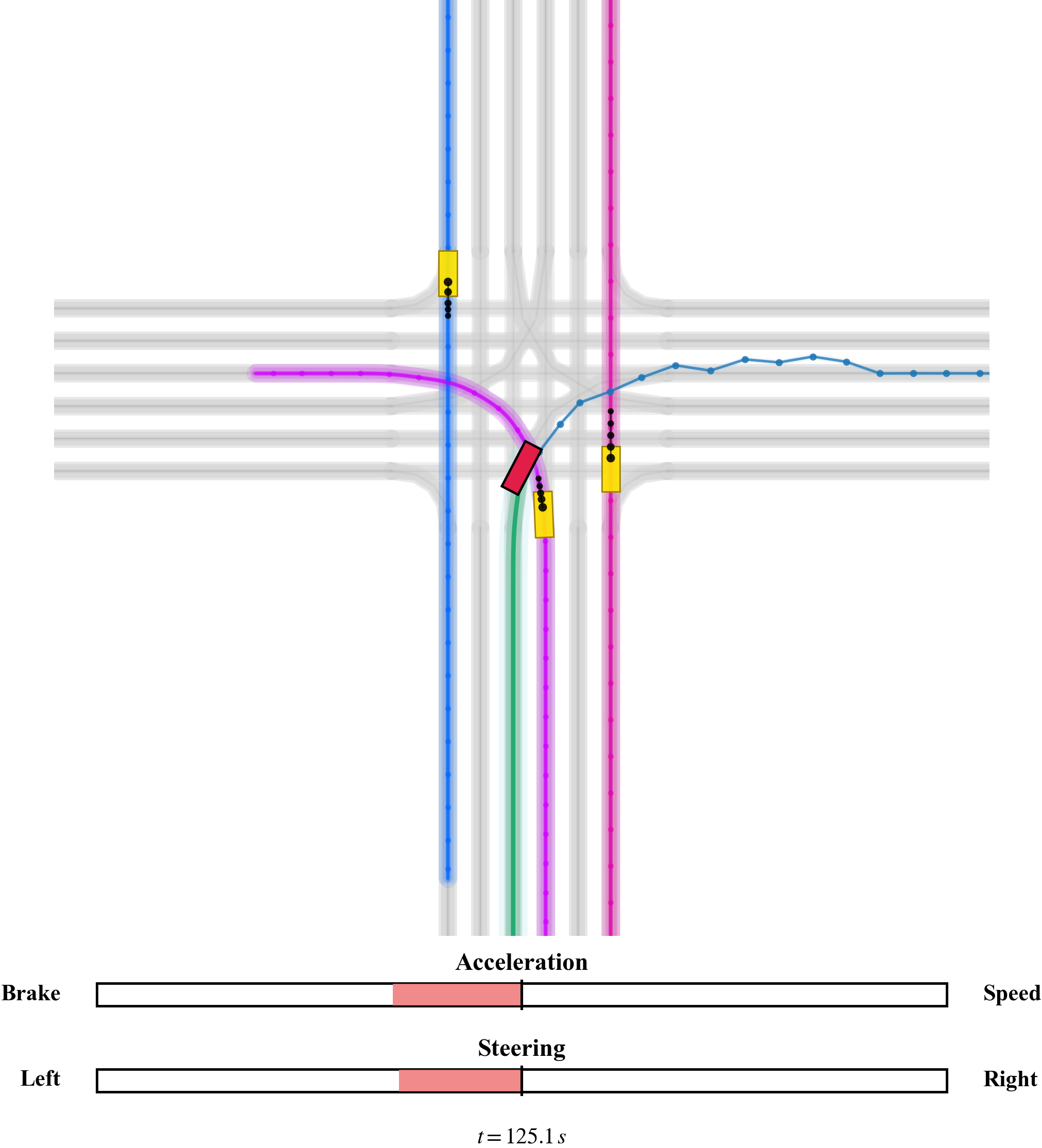}{f}{FNI-RL, 125.1s}%
}

\vspace{0.0cm}

\noindent\makebox[\textwidth][c]{%
\FigTenCell{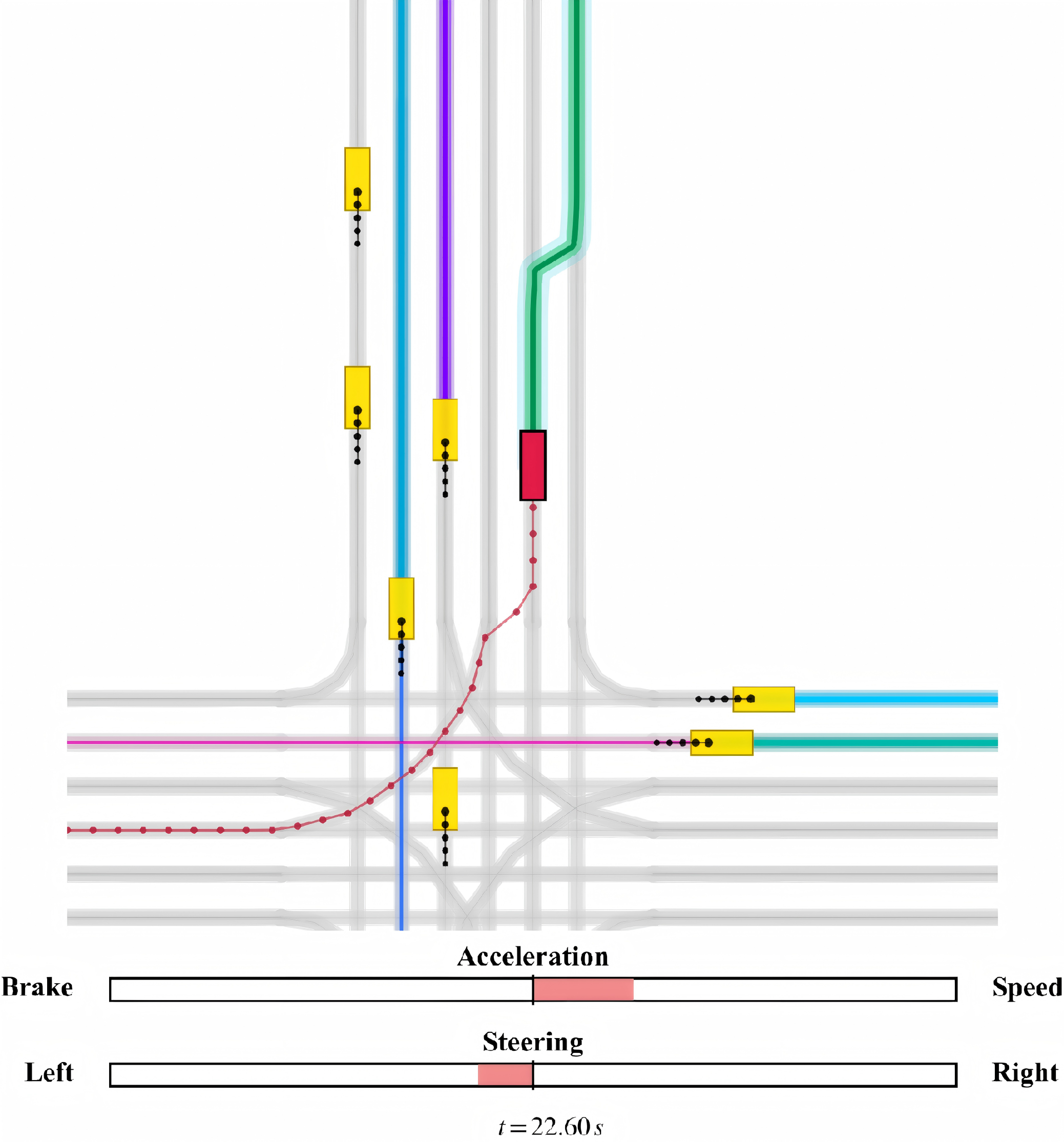}{g}{Brain-SAD, 22.6s}%
\hspace{\FigTenGap}%
\FigTenCell{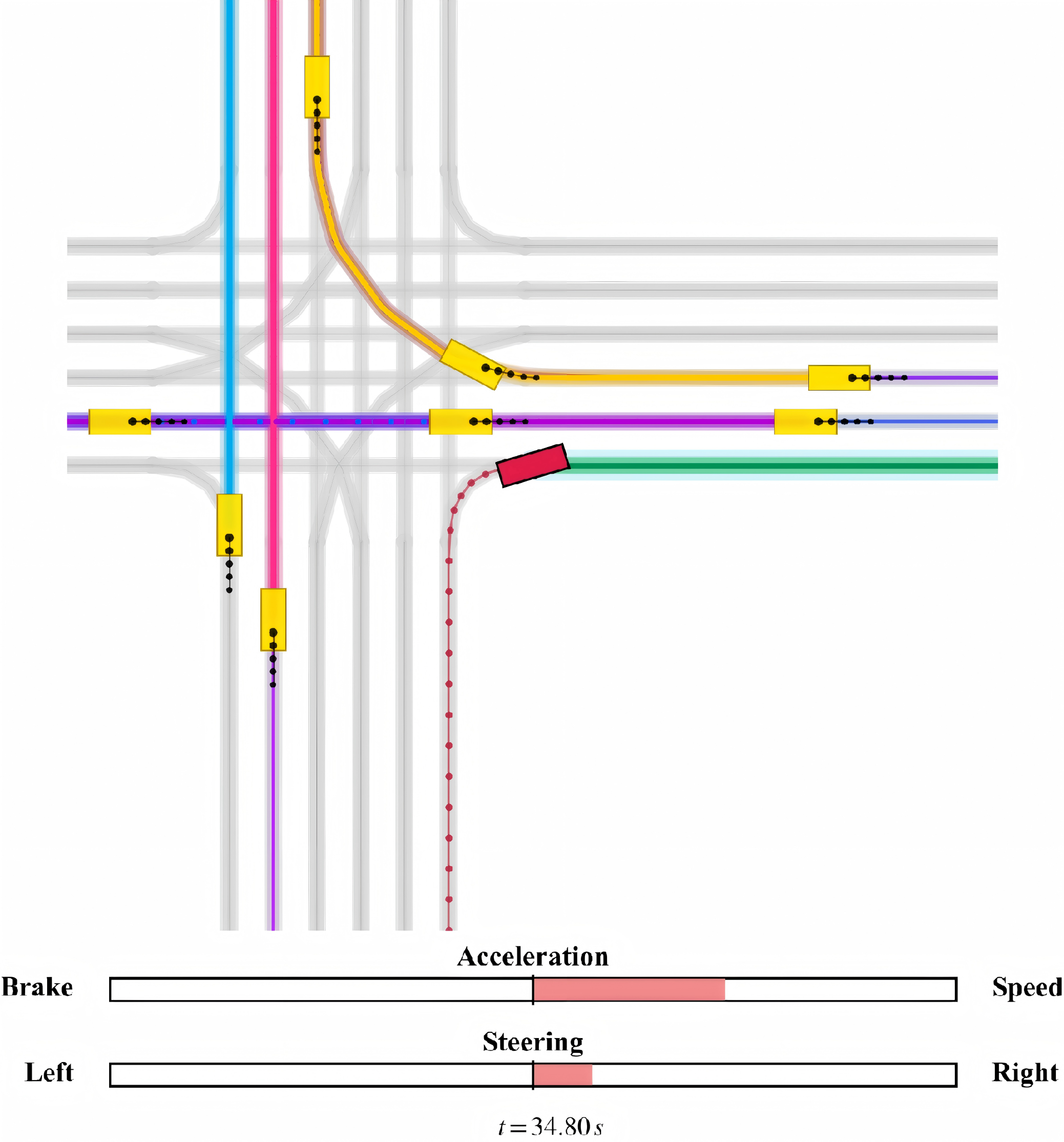}{h}{Brain-SAD, 34.8s}%
\hspace{\FigTenGap}%
\FigTenCell{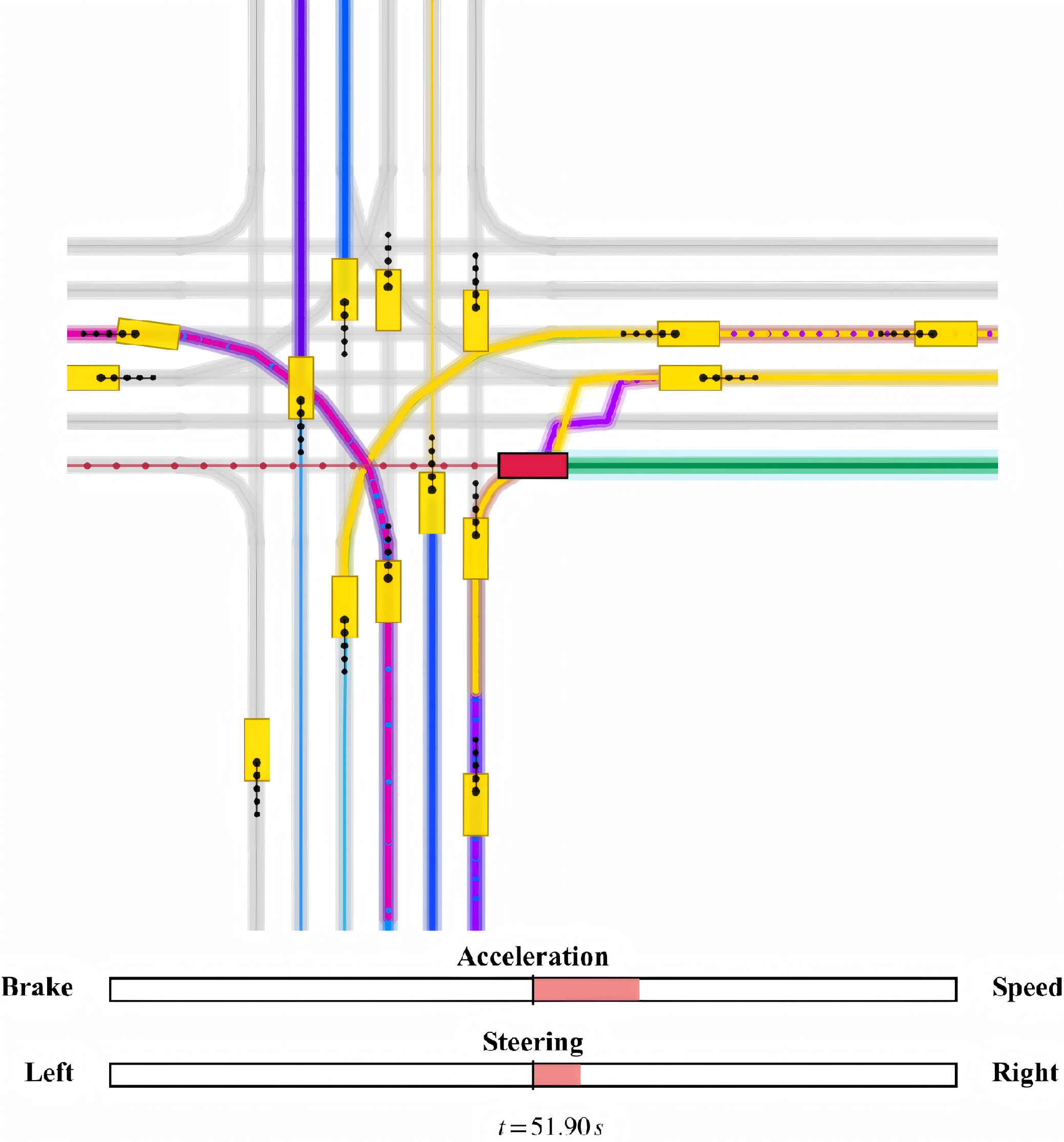}{i}{Brain-SAD, 51.9s}%
\hspace{\FigTenGap}%
\FigTenCell{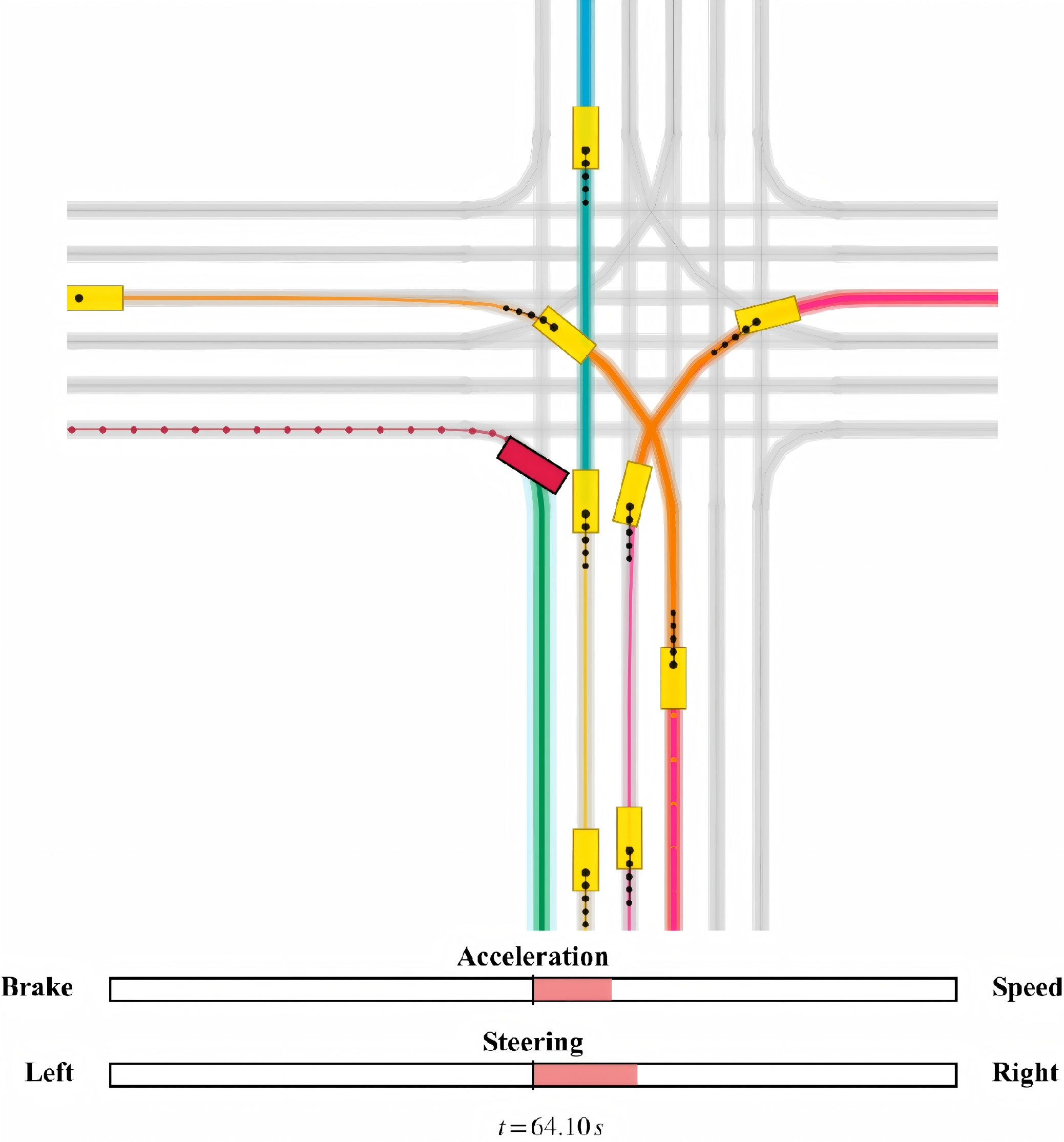}{j}{Brain-SAD, 64.1s}%
\hspace{\FigTenGap}%
\FigTenCell{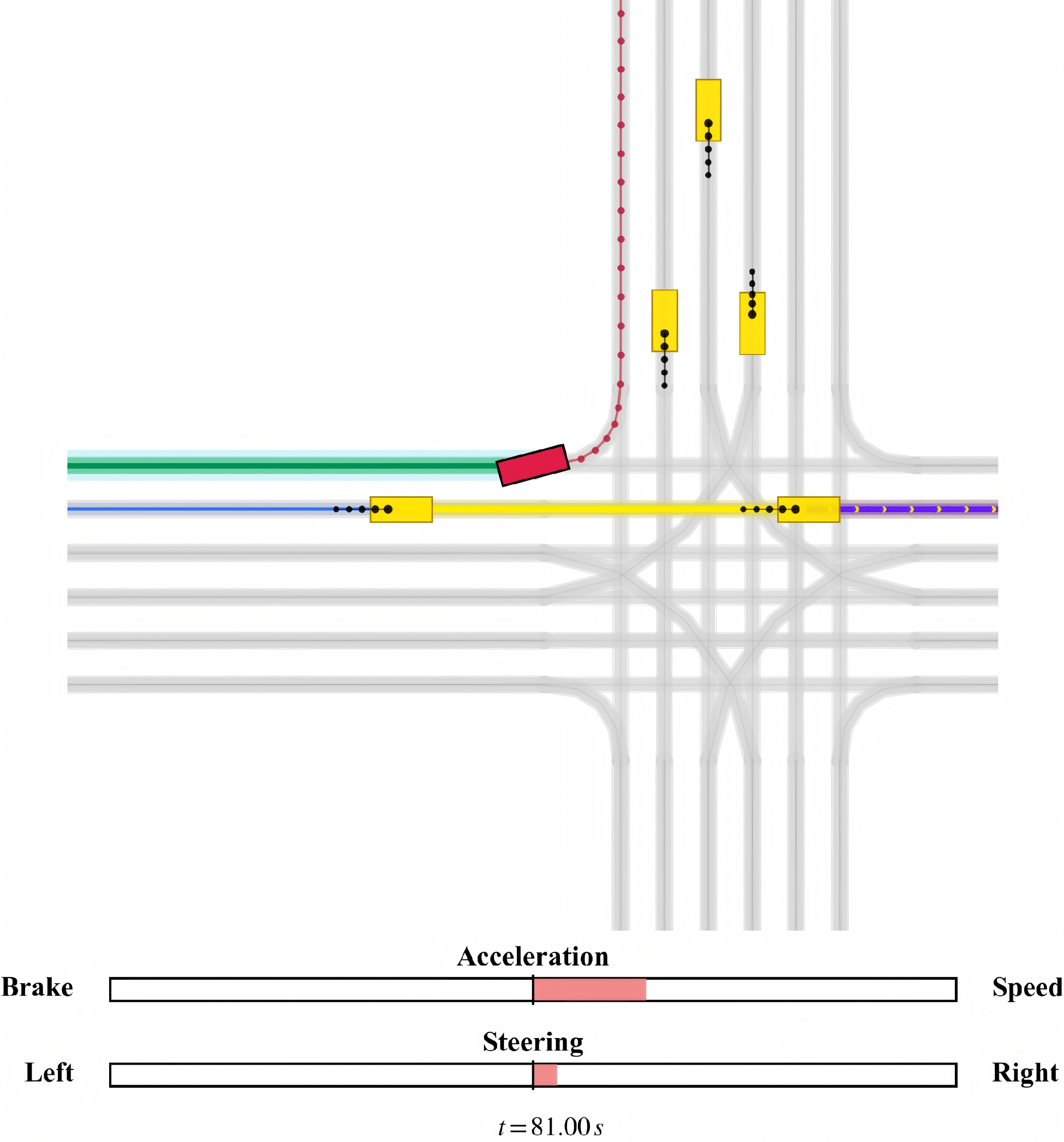}{k}{Brain-SAD, 81.0s}%
\hspace{\FigTenGap}%
\FigTenCell{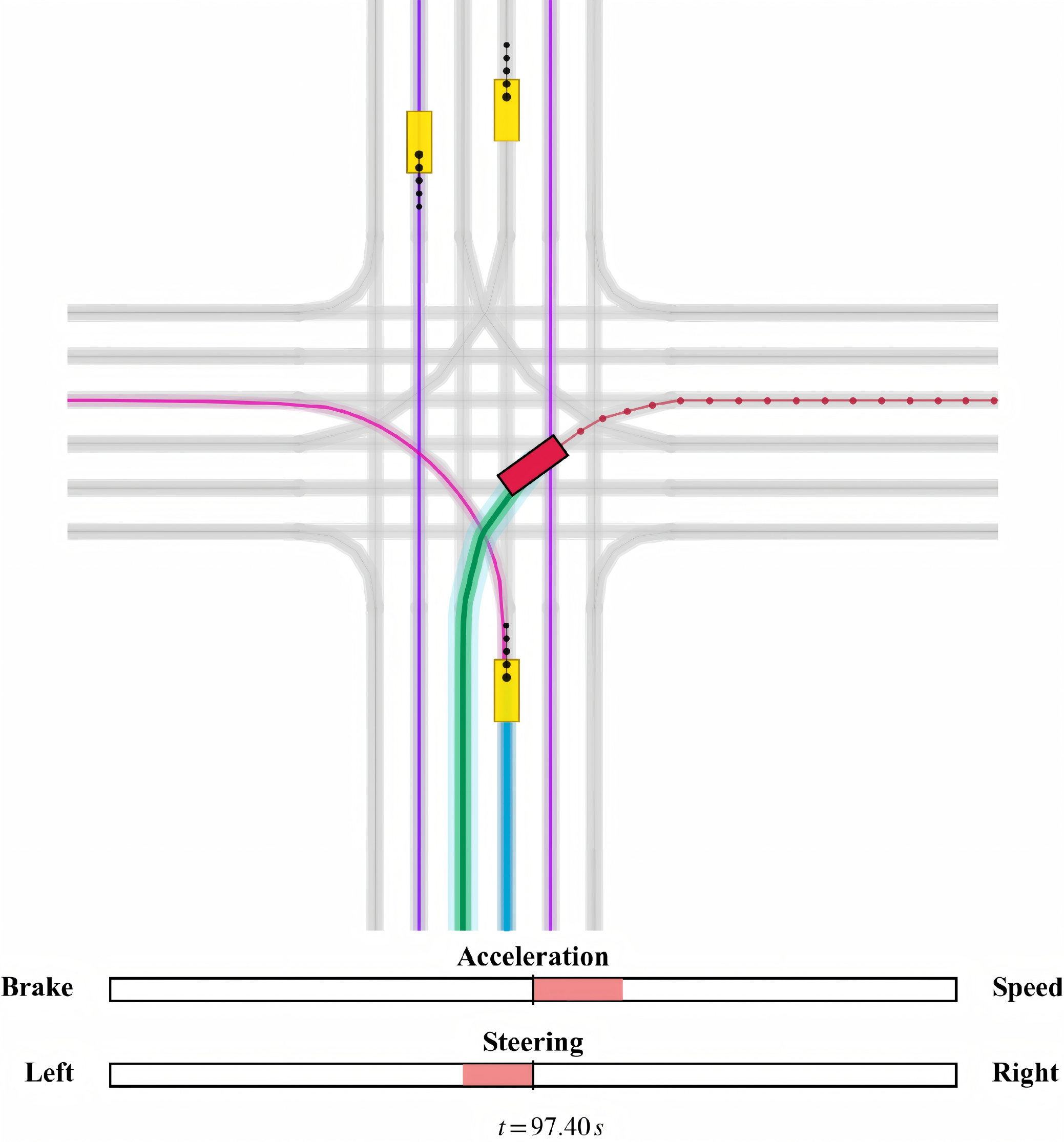}{l}{Brain-SAD, 97.4s}%
}

\endgroup
\vspace{-0.2cm}
\caption{The driving trace of the ego-vehicle when passing 6 continuous intersections controlled by FNI-RL (\textit{a}--\textit{f}) and Brain-SAD (\textit{g}--\textit{l}) in the same best episode used in Fig. 9 (\textit{e})--(\textit{h}).}
\label{Fig:10}
\vspace{-0.05cm}
\end{figure*}

Finally, in Fig. 11, we fetch the \textbf{\textit{best episode}} of Brain-SAD, SVPG, FNI-RL and EFT-RL, based on which we further compare their adaptability or reliability against the varying interaction complexity across the 6 continuous intersections in Fig. 8 (\textit{right}). In Fig. 11 (\textit{a}), we compare the perceptional fear-reaction of Brain-SAD and FNI-RL, where the purple line represents the static state-projected fear-signal of FNI-RL, with the brown line as dynamic fear signal of Brain-SAD computed by eq (3). The background in light \textbf{\textit{blue}} and \textbf{\textit{pink}} respectively means \textbf{\textit{long-}} or \textbf{\textit{short-term}} policy selected at each step. It can be observed in Fig. 11 (\textit{a}) that Brain-SAD has better environmental adaptation, or better dynamic fear perception than FNI-RL, where the fear-signal of Brain-SAD gradually decrease when entering into long-term policy, presenting more coordinated tendency to the interaction complexity (\textit{easy}-\textit{hard}-\textit{easy}) across 6 intersections configured in Fig. 8 (\textit{right}).

In this way, when turning to the policy selection in Fig. 11 (\textit{b}), our Brain-SAD has transited from short-term policy (in \textbf{\textit{pink}} background) to long-term policy (in \textbf{\textit{blue}} background), gradually learning how to handle collision or emergency as the regular evolving scenario via long-term policy. That is, in Fig. 11 (\textit{b}), the policy selection signal \textit{$g_{PFC}$} grows into 1.0 entering long-term policy (see policy selection loss $L_{\Phi}^{\text{selection}}$ in eq (17)) with distinct opposite tendency between \textit{Calmness} $Signal_{5-HT}^{t}$ and \textit{Nervousness} $Signal_{NE}^{t}$ constituting \textit{$g_{PFC}$} (see eq (4)). Consequently, in Fig. 11 (\textit{c}), when observing the cumulative reward at each step of Brain-SAD, SVPG, FNI-RL and EFT-RL, our Brain-SAD has obtained \textbf{\textit{steadier}} reward curve through \textbf{\textit{less}} steps at \textbf{\textit{higher}} level, revealing \textbf{\textit{better}} environmentally coordinated policy guidance effect by the dynamic fear-perception (see Fig. 11 (\textit{a}) \& (\textit{c})).

\begin{figure*}[!t]
\setlength{\abovecaptionskip}{0.00cm}
\setlength{\belowcaptionskip}{-0.1cm}
\centering

\begingroup
\newlength{\FigElevenImageHeight}
\setlength{\FigElevenImageHeight}{0.148\textwidth}

\subfigure[Perceptional fear-signal comparison]{%
  \begin{minipage}[b]{0.322\textwidth}
    \centering
    \includegraphics[
      width=\linewidth,
      height=\FigElevenImageHeight
    ]{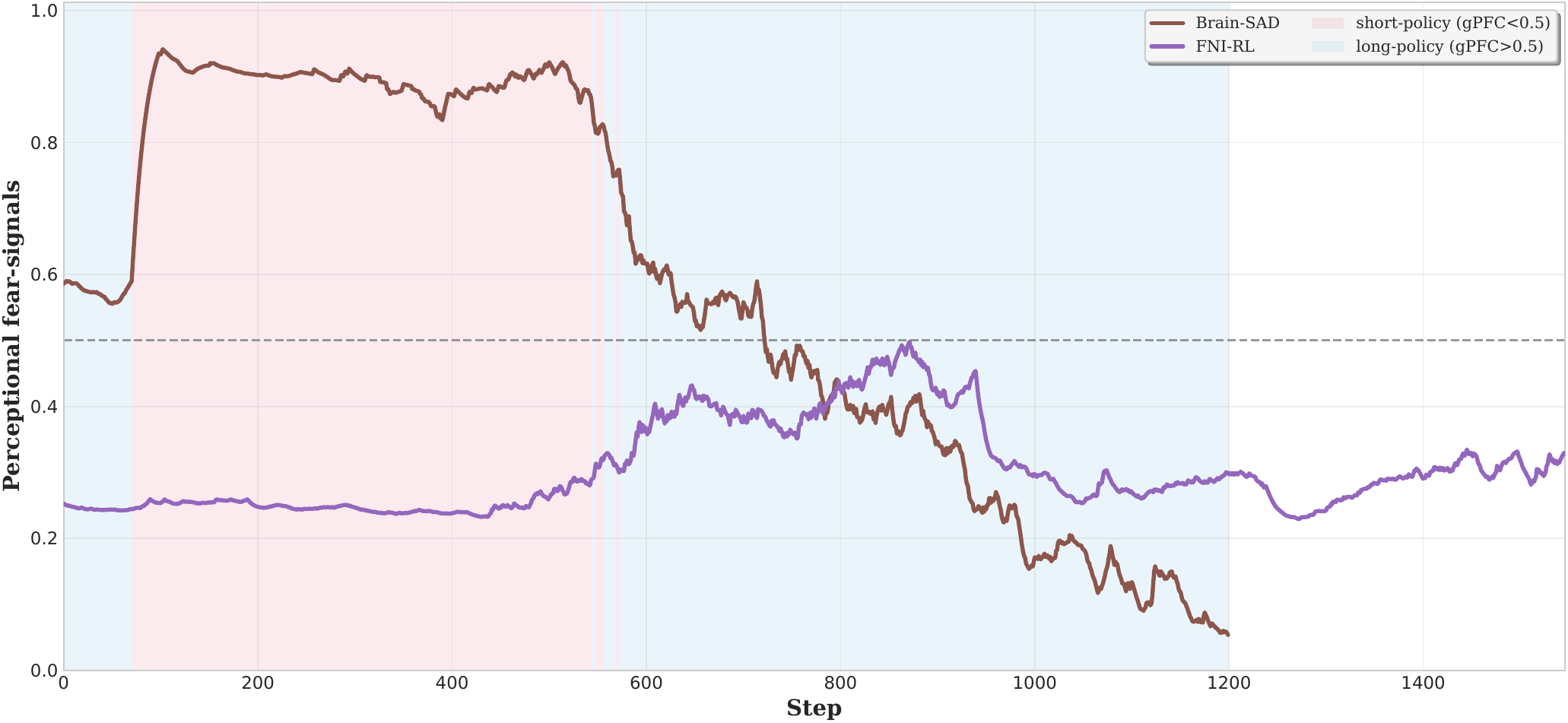}%
  \end{minipage}%
}%
\hspace{0.012\textwidth}%
\subfigure[Policy selection in Brain-SAD]{%
  \begin{minipage}[b]{0.322\textwidth}
    \centering
    \includegraphics[
      width=\linewidth,
      height=\FigElevenImageHeight
    ]{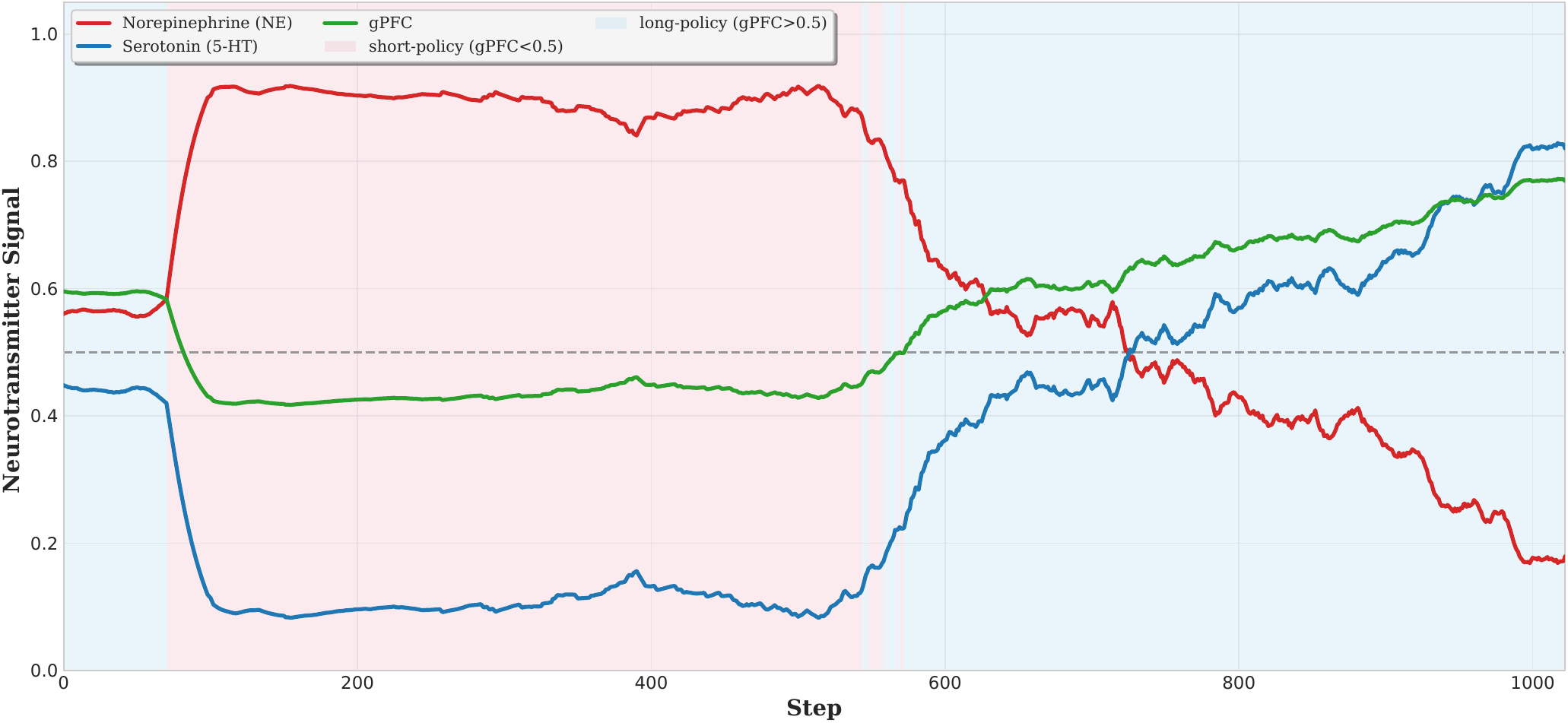}%
  \end{minipage}%
}%
\hspace{0.012\textwidth}%
\subfigure[Policy adaptability comparison]{%
  \begin{minipage}[b]{0.322\textwidth}
    \centering
    \includegraphics[
      width=\linewidth,
      height=\FigElevenImageHeight
    ]{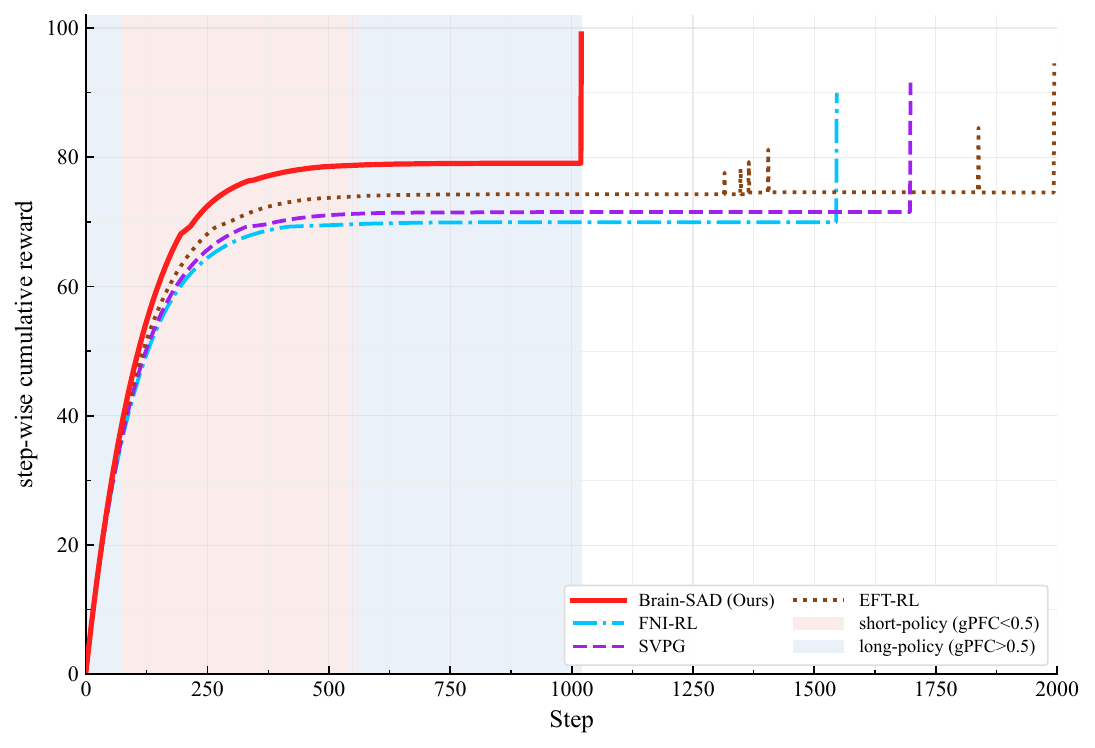}%
  \end{minipage}%
}%

\endgroup
\caption{Environmental adaptation comparison among Brain-SAD, SVPG, FNI-RL and EFT-RL. (\textbf{NOTE:} Simulated in the 6 continuous intersections in Fig. 8, we compare: \textbf{(\textit{a})}. Perceptional fear-signals between Brain-SAD and FNI-RL; \textbf{(\textit{b})}. Policy selection on Brain-SAD based on the fined-grained signals on \textit{calmness} and \textit{nervousness}; \textbf{(\textit{c})}. Policy adaptability by step-wise cumulative reward across continuous intersections trained in the \textbf{\textit{best}} episode).}
\label{Fig:11}
\vspace{-0.05cm}
\end{figure*}

\section{Conclusion \& Future Work}
\label{sec:conclusion_future_work}

In this paper, we propose Brain-SAD, a dynamic fear constrained RL framework with dual-path policy selection for Safe AD. By a series of experiments, we have discovered that Brain-SAD first has the advantage on environmentally coordinated dynamic fear perception, by which to adopt long-term or short-term policy more properly matched with the current interaction scenario. Second, the above dynamic fear perception has been converted into dynamic fear constraint coupled with action impact or value estimation, further imposed onto \textbf{\textit{online}} policy optimization, ensuring Brain-SAD can learn how to handle collision with stronger policy reliability against constant environmental disturbance. Finally, the dynamic fear constraint has also enabled Brain-SAD to complete continuous interactions in shorter time and higher success with more stable action distribution. In the future, we will devote effort to explore more fine-grained way to construct dynamic fear constraint, for instance, from value estimation via LM inference in more complex environment.

\FloatBarrier
\bibliographystyle{unsrt}
\bibliography{reference}
\vspace{0.2cm}

%

\raggedbottom

\vspace{-0.08cm}


\newlength{\AuthorPhotoWidth}
\setlength{\AuthorPhotoWidth}{0.76in}

\newlength{\AuthorPhotoHeight}
\setlength{\AuthorPhotoHeight}{0.82in}

\newlength{\AuthorPhotoBoxWidth}
\setlength{\AuthorPhotoBoxWidth}{\AuthorPhotoWidth}

\newlength{\AuthorBioGap}
\setlength{\AuthorBioGap}{0.07in}

\newlength{\AuthorBioTextWidth}
\setlength{\AuthorBioTextWidth}{%
    \dimexpr
    \columnwidth
    -\AuthorPhotoBoxWidth
    -\AuthorBioGap
    \relax
}

\newsavebox{\AuthorPhotoCoverBox}
\newlength{\AuthorScaledHeight}
\newlength{\AuthorCropX}
\newlength{\AuthorCropY}

\newcommand{\centercropauthorphoto}[1]{%
    \sbox{\AuthorPhotoCoverBox}{%
        \includegraphics[width=\AuthorPhotoWidth]{#1}%
    }%
    \setlength{\AuthorScaledHeight}{%
        \dimexpr\ht\AuthorPhotoCoverBox+\dp\AuthorPhotoCoverBox\relax
    }%
    \ifdim\AuthorScaledHeight<\AuthorPhotoHeight
        \sbox{\AuthorPhotoCoverBox}{%
            \includegraphics[height=\AuthorPhotoHeight]{#1}%
        }%
    \fi
    \setlength{\AuthorCropX}{\wd\AuthorPhotoCoverBox}%
    \addtolength{\AuthorCropX}{-\AuthorPhotoWidth}%
    \setlength{\AuthorCropX}{0.5\AuthorCropX}%
    \setlength{\AuthorCropY}{%
        \dimexpr\ht\AuthorPhotoCoverBox+\dp\AuthorPhotoCoverBox-\AuthorPhotoHeight\relax
    }%
    \setlength{\AuthorCropY}{0.5\AuthorCropY}%
    \clipbox{%
        \AuthorCropX\space
        \AuthorCropY\space
        \AuthorCropX\space
        \AuthorCropY
    }{\usebox{\AuthorPhotoCoverBox}}%
}

\vfill

\end{document}